\documentclass[preprint,12pt,sort]{elsarticle}

\journal{Information Fusion}

\usepackage{natbib}
\usepackage{graphicx} 
\usepackage{amsmath,amsfonts,amssymb,mathrsfs}
\usepackage[scr=rsfso]{mathalfa}
\usepackage[table,xcdraw]{xcolor}
\usepackage{bbm}
\usepackage{verbatim}
\usepackage{pgfplots}
\pgfplotsset{compat=1.18}	
\usetikzlibrary{pgfplots.statistics} 
\usetikzlibrary{matrix}
\usepgfplotslibrary{groupplots}
\usepackage{tabularx}
\usepackage{multirow}
\usepackage{url}

\usepackage{booktabs}
\usepackage{makecell}

\usetikzlibrary{positioning, calc, chains, patterns}
\usetikzlibrary{decorations, decorations.pathreplacing, decorations.pathmorphing, calligraphy}
\usetikzlibrary{arrows.meta}
\usetikzlibrary{
    shapes,
    shapes.geometric,
    shapes.symbols,
    shapes.arrows,
    shapes.multipart,
    shapes.callouts,
    shapes.misc}
    
\tikzset{>=Stealth[round]}
\tikzset{
    ncbar angle/.initial=90,
    ncbar/.style={
        to path=(\tikztostart)
        -- ($(\tikztostart)!#1!\pgfkeysvalueof{/tikz/ncbar angle}:(\tikztotarget)$)
        -- ($(\tikztotarget)!($(\tikztostart)!#1!\pgfkeysvalueof{/tikz/ncbar angle}:(\tikztotarget)$)!\pgfkeysvalueof{/tikz/ncbar angle}:(\tikztostart)$)
        -- (\tikztotarget)
    },
    ncbar/.default=0.5cm,
}

\tikzset{square left brace/.style={ncbar=0.2cm}}
\tikzset{square right brace/.style={ncbar=-0.2cm}}

\usepackage{hyperref}
\hypersetup{
  colorlinks=false,
  linkbordercolor=white,
  urlbordercolor=white,
  pdfborder={0 0 0}
}

\newcolumntype{C}[1]{>{\centering\arraybackslash}m{#1}}
\newcolumntype{Z}{>{\centering\arraybackslash}X}
\newcommand{\Fusion}[1]{\ensuremath{\mathscr{F}_{#1}}}
\newcommand{\Nlayers}[2][]{\ensuremath{#2^{#1}}}
\def\attention{\ensuremath{\mathbb{A}}}
\def\concat{\ensuremath{\oplus}}
\def\tensorOP{\ensuremath{\boxplus}}
\def\calibr{\ensuremath{\mathbb{S}}}
\def\kShare{\ensuremath{\rightleftharpoons}}
\def\out{\ensuremath{_{\to}}}

\newcommand{\Sync}[1]{#1\ensuremath{^\ddagger}}

\usepackage{color}

\graphicspath{{./img/}}

\begin{document}

\begin{frontmatter}



\title{A Systematic Review of Intermediate Fusion in Multimodal Deep Learning for Biomedical Applications}

\author[aff1]{Valerio Guarrasi}
\ead{valerio.guarrasi@unicampus.it}

\author[aff2]{Fatih Aksu\texorpdfstring{\fnref{contrib}}{}}
\ead{fatih.aksu@st.hunimed.eu}

\author[aff1]{Camillo Maria Caruso\texorpdfstring{\fnref{contrib}}{}}
\ead{camillomaria.caruso@unicampus.it}

\author[aff3]{Francesco Di Feola\texorpdfstring{\fnref{contrib}}{}}
\ead{francesco.feola@umu.se}

\author[aff1]{Aurora Rofena\texorpdfstring{\fnref{contrib}}{}}
\ead{aurora.rofena@unicampus.it}

\author[aff1]{Filippo Ruffini\texorpdfstring{\fnref{contrib}}{}}
\ead{filippo.ruffini@unicampus.it}

\author[aff1,aff3]{Paolo Soda\texorpdfstring{\corref{cor}}{}}
\ead{p.soda@unicampus.it, paolo.soda@umu.se}

\cortext[cor]{Correspondence: paolo.soda@umu.se, p.soda@unicampus.it}
\fntext[contrib]{These authors contributed equally to this work.}

\affiliation[aff1]{organization={Research Unit of Computer Systems and Bioinformatics, Department of Engineering, Università Campus Bio-Medico di Roma},
            city={Rome},
            state={Italy},
            country={Europe}}

\affiliation[aff2]{organization={Department of Biomedical Sciences, Humanitas University},
            city={Milan},
            state={Italy},
            country={Europe}}

\affiliation[aff3]{organization={Department of Diagnostics and Intervention, Radiation Physics, Biomedical Engineering, Umeå University},
city={Umeå},
state={Sweden},
country={Europe}}

\begin{abstract}
Deep learning has revolutionized biomedical research by providing sophisticated methods to handle complex, high-dimensional data.
Multimodal deep learning (MDL) further enhances this capability by integrating diverse data types such as imaging, textual data, and genetic information, leading to more robust and accurate predictive models.
In MDL, differently from early and late fusion methods, intermediate fusion stands out for its ability to effectively combine modality-specific features during the learning process.
This systematic review aims to comprehensively analyze and formalize current intermediate fusion methods in biomedical applications.
We investigate the techniques employed, the challenges faced, and potential future directions for advancing intermediate fusion methods.
Additionally, we introduce a structured notation to enhance the understanding and application of these methods beyond the biomedical domain.
Our findings are intended to support researchers, healthcare professionals, and the broader deep learning community in developing more sophisticated and insightful multimodal models. 
Through this review, we aim to provide a foundational framework for future research and practical applications in the dynamic field of MDL.
\end{abstract}




\end{frontmatter}

\section{Introduction} \label{sec:Introduction}


Deep learning has transformed the landscape of computational approaches in biomedical research, offering unique capabilities in handling complex, high-dimensional data~\cite{bib:lecun2015deep}.
Multimodal deep learning (MDL), which integrates multiple data types, has emerged as an innovative approach, leveraging the power of deep learning algorithms to interpret and integrate diverse data types, enhancing the robustness and accuracy of predictive models~\cite{bib:ramachandram2017deep}.
Unlike traditional unimodal methods, MDL utilizes a variety of data sources such as imaging, textual data, and genetic information, to name a few, providing a more comprehensive understanding of complex biomedical phenomena.
This approach is particularly relevant in a field where data diversity and volume are rapidly expanding, offering unprecedented opportunities for advancements in diagnosis, treatment, and patient care.

In the domain of multimodal learning, the integration of data from multiple sources can be achieved through various fusion techniques, i.e., early, late and intermediate fusion.
Early fusion combines features at the data level, potentially losing unique modality-specific characteristics.
Formally, let $x_1, x_2, \ldots, x_n$ represent feature vectors from $n$ different modalities.
The fusion function $\mathscr{F}$ combines these features into a single vector: 
\begin{equation}
    x = \mathscr{F} ( x_1, x_2, \ldots, x_n )
\end{equation}
which is then fed into a learning model $f$, resulting in the output $y$:
\begin{equation}
    y = f ( x )
\end{equation}
Late fusion, on the other hand, occurs at the decision level, often missing opportunities for deeper interaction between modalities.
Formally, the feature vectors $x_1, x_2, \ldots, x_n$ are fed into independent models $f_1, f_2, \ldots, f_n$, producing separate predictions $y_1, y_2, \ldots, y_n$.
The fusion function $\mathscr{F}$ then combines these predictions:
\begin{equation}
    y =  \mathscr{F}( y_1, y_2, \ldots, y_n )
\end{equation}
Intermediate fusion, striking a balance, integrates data at the feature extraction stage, allowing for a more effective combination of modality-specific features.
Formally, the feature vectors $ x_1, x_2, \ldots, x_n $ are processed by separate components of the model $ f_1, f_2, \ldots, f_n $, yielding intermediate representations $ h_1, h_2, \ldots, h_n $.
These intermediate representations are fused using $\mathscr{F}$
\begin{equation}
    h = \mathscr{F} ( h_1, h_2, \ldots, h_n )
\end{equation}
and then a final component $f$ processes $h$ to produce the output
\begin{equation}
    y = f ( h )
\end{equation}
Further details on each component can be found in Section~\ref{sec:definitions}.
Among the various fusion techniques employed in MDL, intermediate fusion stands out for its ability to effectively integrate information at essential stages of the learning process, potentially leading to more accurate and robust models.
This method is especially beneficial in biomedical applications where different data types, like imaging and genomic information, need to interact closely to produce meaningful insights.
The main feature of intermediate fusion lies in its ability to preserve and utilize the distinct qualities of each data type, enhancing the model's capability to handle the complexities inherent in biomedical data.

The utility of deep learning in the context of intermediate fusion in multimodal settings, particularly within biomedical research, is marked by its ability to process and fuse distinct data modalities at a more abstract level~\cite{bib:lahat2015multimodal}.
In intermediate fusion, deep learning models are particularly capable at entangling the complex, nonlinear relationships that often exist between different modalities in biomedical data.
This understanding is critical for accurately interpreting the multimodal nature of such data, where each modality carries its own information.
By leveraging deep learning's ability in feature extraction and representation learning, intermediate fusion methods can effectively bridge the gap between these diverse data sources, leading the way for more insightful biomedical analysis.
This approach not only may enhance the accuracy of predictive models but also may unveil patterns and interactions that might be missed by more conventional methods.

The application of MDL in the biomedical domain faces unique challenges, notably due to the heterogeneity and high dimensionality of the data~\cite{bib:sui2012review}.
Biomedical datasets often comprise varied data modalities, each with distinct characteristics and scales, making it difficult to achieve effective integration and analysis.
Intermediate fusion, applied within the deep learning framework, addresses these challenges efficiently.
By fusing data at the feature level, it allows to effectively manage the heterogeneity by learning a shared representation space, where the complex interaction between unimodal and multimodal features is explored.
Moreover, intermediate fusion helps in mitigating the curse of dimensionality, a common issue in biomedical datasets~\cite{bib:viswanath2017dimensionality}, by facilitating dimensionality reduction and feature selection in a way that preserves the most informative aspects of each modality.

While deep learning and intermediate fusion offer significant advantages in handling multimodal biomedical data, they are not without limitations~\cite{bib:ching2018opportunities}.
One of the primary concerns is the black-box nature of deep learning models, which often leads to a lack of transparency and interpretability~\cite{bib:ahmad2018interpretable}. 
This is particularly problematic in biomedical settings where understanding the reasoning behind a model's decision can be as critical as the decision itself.
Additionally, these models require large amounts of data to perform optimally, which can be a challenge in situations where data is scarce or privacy concerns limit data availability~\cite{bib:vayena2018machine}.
Intermediate fusion, despite its benefits, also faces challenges in optimally balancing the contributions of different modalities, especially when the data from one modality is sparse or of lower quality~\cite{bib:ramachandram2017deep}.
Furthermore, the computational complexity of deep learning models, tangled with the complexity of intermediate fusion, can lead to significant resource demands, posing challenges in terms of computational cost and time~\cite{bib:strubell2019energy}.
Acknowledging these limitations is crucial for advancing the field, as it guides researchers in seeking improvements and developing more robust, transparent, and efficient methods.
But in the end, the application of intermediate fusion techniques is still evolving, particularly in high-stakes areas such as disease diagnosis, treatment planning, and outcome prediction, where the precise modeling of complex biological interactions can significantly impact the decisions made by clinicians and healthcare providers~\cite{bib:min2017deep}.

\subsection{Objective of the Review}

The primary objective of this systematic review is to comprehensively analyze and formalize the current intermediate fusion methods of MDL utilized in biomedical applications.
The field of intermediate fusion in MDL, particularly within biomedical contexts, is still in its early stages.
As such, there is a significant need for a structured analysis and for introducing a formal notation that not only categorizes and evaluates these methods, but also provides a framework that can be extended beyond the biomedical domain.

This systematic review is guided by a set of research questions, designed to shed light on the field of intermediate fusion in MDL, particularly within biomedical applications:
\begin{enumerate}
    \item \textit{Which are the intermediate fusion techniques currently employed in biomedical deep learning models?} This question aims to identify and categorize the various methods of intermediate fusion, highlighting their unique features and applications in the biomedical context.
    \item \textit{What are the existing challenges and limitations associated with these intermediate fusion methods?} This seeks to understand the drawbacks and areas where current methods fall short, providing insights into potential directions for future improvement.
    \item \textit{What are the future directions that should be pursued for intermediate fusion methods in biomedical applications?} This examination focuses on exploring potential directions for advancement in intermediate fusion, highlighting emerging trends, potential improvements, and unexplored areas that could significantly enhance the application and efficacy of these techniques in biomedical contexts.
    \item \textit{How can the structured notation proposed in this review enhance the understanding and application of intermediate fusion techniques beyond just biomedical applications?} This question explores the broader implications of the review's findings, considering how the proposed notation and categorization can be applied to other domains within deep learning.
\end{enumerate}
Answering these questions will not only provide a thorough understanding of the current state of intermediate fusion in MDL for biomedical applications but will also offer a foundation for future research, extending the insights gained to a wider range of applications.

The findings and discussions presented in this review are particularly relevant to a diverse group of professionals and researchers:
\begin{itemize}
    \item \textbf{Researchers}: This review holds significant value for researchers at the intersection of deep learning and biomedical applications. It offers comprehensive insights into the current state of intermediate fusion methods, their performance outcomes, and potential research gaps. Researchers can use this information to guide future investigations, develop more sophisticated models, and explore new paths in MDL.
    \item \textbf{Healthcare Professionals and Practitioners}: Healthcare professionals, including radiologists, pathologists, and clinicians who frequently engage with multimodal imaging or diagnostic techniques, will find this review informative. It provides an overview of the latest advancements in MDL and its potential impact on clinical practice. Understanding these fusion methods can assist healthcare practitioners in evaluating the reliability and applicability of deep learning models, ultimately enhancing patient care and diagnostic accuracy.
    \item \textbf{Deep Learning Community}: Beyond the biomedical domain, this review is also helpful for professionals and researchers working with MDL across various fields. The introduction of a structured notation and analysis of intermediate fusion methods is extendable to other applications, offering a versatile framework that can be adapted to different modalities and datasets. This broader applicability makes the review a crucial resource for anyone interested in the practical and theoretical aspects of MDL.
\end{itemize}

This survey is accompanied by two supplementary documents that contain all the key information discussed in Section~\ref{sec:main_body} for each individual paper, so that in the main manuscript we can avoid the use of excessive citations and notations. Specifically:
\begin{itemize}
    \item \textbf{Supplementary Material A}: includes a table listing all the fields discussed in Sections~\ref{sec:modalities}, \ref{sec:unimodal_module}, \ref{sec:fusion_module}, \ref{sec:multimodal_module}, \ref{sec:target}, \ref{sec:learning}. The table features each analyzed paper as a row and the discussed fields as columns. Readers can refer to this table for specific information about each paper in any of the analyzed fields.
    \item \textbf{Supplementary Material B}: provides a detailed analysis of the fusion mechanisms used in each article. For every article, it presents the corresponding mathematical notation, graphical representation, and categorization, all of which are defined in Section~\ref{sec:fusion_module}. While the main text explains this methodology using selected examples, readers can refer to this material for the complete formalization developed for each individual paper.
\end{itemize}
Both supplementary documents have been uploaded to our GitHub repository \url{https://github.com/cosbidev/Intermediate-Multimodal-Fusion-Bio}, ensuring easy accessibility for further exploration.

\subsection{Comparison with Previous Surveys}

This review distinguishes itself within the landscape of multimodal learning literature by focusing specifically on intermediate fusion methods in the biomedical domain.
While existing surveys, out of the biomedical filed, such as~\cite{bib:guo2019deep, bib:jabeen2023review, bib:gao2020survey, bib:bayoudh2022survey, bib:ramachandram2017deep, bib:liang2022foundations, bib:ngiam2011multimodal, bib:zhang2020multimodal, bib:xu2023multimodal, bib:baltruvsaitis2018multimodal, bib:summaira2021recent, bib:bengio2013representation, bib:zhu2022vision+}, have broadly addressed multimodal machine learning across various fields, they have not deeply explored intermediate fusion techniques.
Similarly, reviews in the biomedical sphere, as~\cite{bib:acosta2022multimodal, bib:stahlschmidt2022multimodal, bib:zhang2020advances}, have touched upon the integration of multimodal machine learning but have not specifically honed in on intermediate fusion methods.
In the context of biomedical applications, some previous surveys have been limited in their categorization, focusing primarily on application domains rather than detailed methodological distinctions~\cite{bib:heiliger2023beyond, bib:lipkova2022artificial, bib:behrad2022overview, bib:zhou2019review}.
Additionally, certain reviews have concentrated solely on a limited set of modalities, often considering only two modalities~\cite{bib:xu2019deep, bib:huang2020fusion, bib:antonelli2019integrating}.
Our survey, therefore, fills a crucial gap by not only concentrating on intermediate fusion within biomedical applications but also by introducing a structured notation system with potential applicability across diverse domains in deep learning.

\section{Methodology}

\subsection{Inclusion/exclusion criteria} \label{sec:inclusion_exclusion}

Our systematic review adheres to specific inclusion and exclusion criteria to ensure a focused and relevant analysis of intermediate fusion in MDL for biomedical applications:
\begin{itemize}
    \item \textbf{Date}: We include studies published up to the present day to capture the most recent advancements in the field.
    \item \textbf{Language}: Only articles published in English are considered to ensure accessibility and standardization in analysis.
    \item \textbf{Study Design}: We exclude review articles, non-peer-reviewed sources such as opinion pieces, editorials, letters, and conference abstracts, as they do not provide original research or detailed methodology.
    \item \textbf{Target Application}: The focus is on articles that explore at least one biomedical application. Articles centered on non-biomedical applications or unrelated fields are excluded.
    \item \textbf{Modalities}: We include studies utilizing multiple modalities, considering both homogenous (e.g., imaging) and heterogeneous types (e.g., imaging combined with tabular data). Studies focusing solely on a single modality or not involving the fusion of multiple modalities are excluded.
    \item \textbf{Fusion}: Articles employing deep learning architectures for fusion, particularly those using intermediate fusion in an end-to-end manner, are included. Articles primarily using early or late fusion methods or non-deep learning approaches are excluded.
\end{itemize}

\subsection{Search Strategy}

Our systematic review implemented a comprehensive search strategy to identify relevant literature focusing on intermediate fusion methods in MDL within biomedical applications.
The search strategy was meticulously crafted to encompass a wide range of pertinent studies while maintaining specificity to the review's objectives.

The literature search was conducted across multiple electronic databases, ensuring a thorough and diverse collection of articles.
The databases included are: PubMed, IEEE Xplore, Scopus and Google Scholar.

To capture the breadth of literature relevant to our review, we utilized a set of synonyms and keywords across three main categories:
\begin{itemize}
\item Multimodal Deep Learning:
\[
    A = \left\{ \begin{array}{l}
    \text{``Multimodal Deep Learning''}, \\
    \text{``Multimodal Fusion''}, \\
    \text{``Multimodal Representation Learning''}
    \end{array} \right\}
\]
\item Biomedical Context:
\[
    B = \left\{ \begin{array}{l}
    \text{``Biomedical''}, \\
    \text{``Medical''}, \\
    \text{``Healthcare''}, \\
    \text{``Clinical''}, \\
    \text{``Biomedicine''}, \\
    \text{``Medicine''}
    \end{array} \right\}
\]
\item Intermediate Fusion:
\[
    C = \left\{ \begin{array}{l}
    \text{``Joint fusion''}, \\
    \text{``Intermediate fusion''}, \\
    \text{``Mid-level fusion''}, \\
    \text{``Feature-level fusion''}
    \end{array} \right\}
\]
\end{itemize}

The search query was formulated using a combination of these synonyms, structured as follows:
\[ \text{Query} = (A_1 \text{ OR } A_2 \text{ OR } \ldots) \text{ AND } (B_1 \text{ OR } B_2 \text{ OR } \ldots) \text{ AND } (C_1 \text{ OR } C_2 \text{ OR } \ldots) \]
This structure allowed for a comprehensive yet focused search, ensuring that the results were pertinent to the scope of our review.

\begin{figure}[t]
\centering
\includegraphics[width=\textwidth]{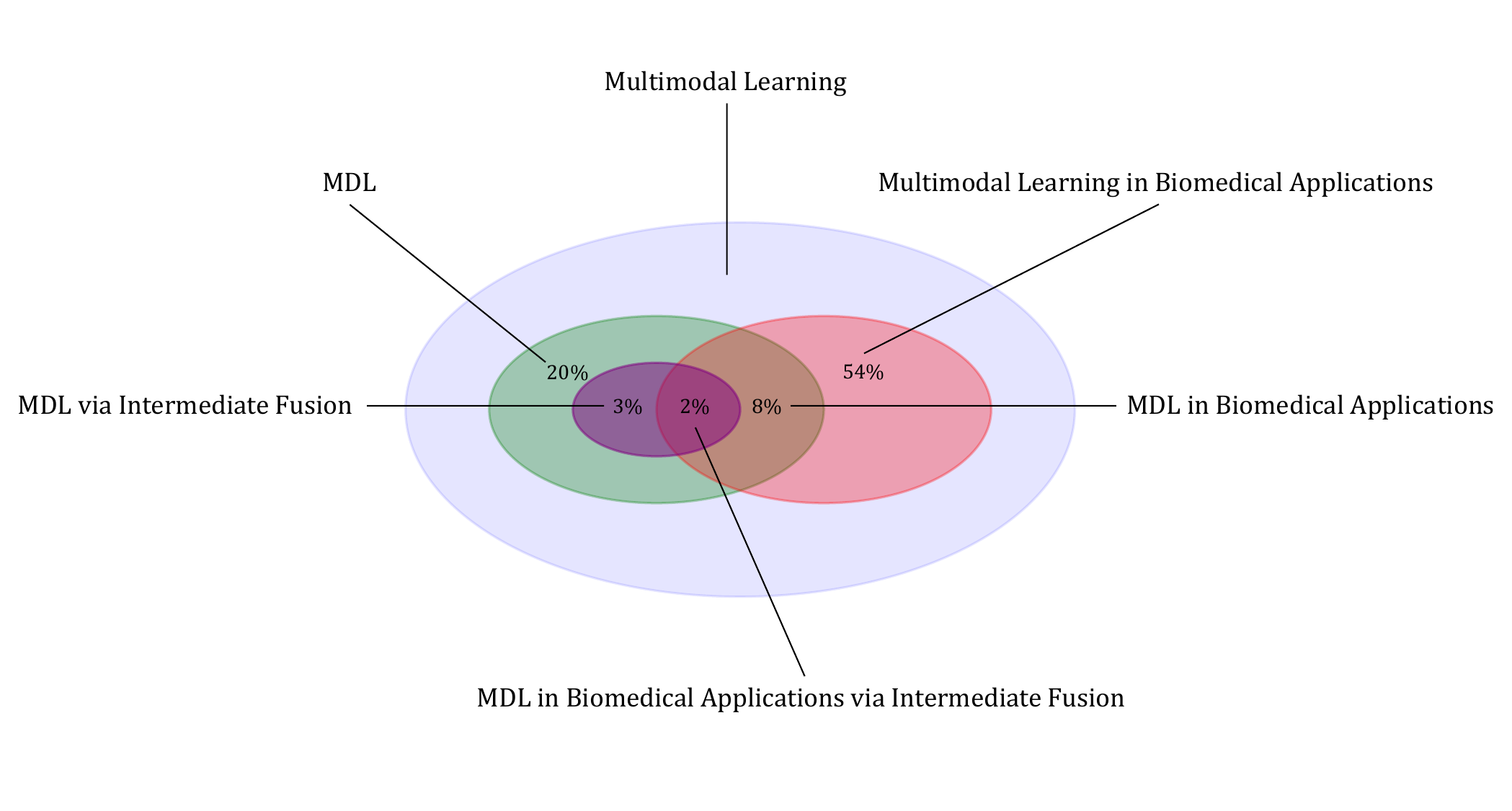}
\caption{Venn Diagram of Search Hits by Category. This diagram displays the percentage of search hits for each specified category, ranging from general fields to more specialized areas, i.e., ``Multimodal Learning'', ``Multimodal Learning in Biomedical Applications'', ``Multimodal
Deep Learning'', ``Multimodal Deep Learning in
Biomedical Applications'', ``Multimodal Deep Learning via Intermediate Fusion'', ``Multimodal Deep Learning in Biomedical Applications via Intermediate Fusion''.}
\label{fig:search_analysis}
\end{figure}

To understand the current state and trajectory of research in this area, we conducted a comprehensive search analysis by examining the publication numbers across various subdomains of multimodal learning, as shown in \figurename~\ref{fig:search_analysis}.
We can observe that the significant drop of publications from ``Multimodal Learning'' to ``Multimodal Deep Learning'' highlights that, while multimodal learning is a broad field, the incorporation of deep learning techniques is a more focused area.
This reflects the growing interest in leveraging deep learning's feature extraction and representation learning capabilities within multimodal settings.
The slight narrowing for ``Multimodal Learning in Biomedical Applications'' and for ``Multimodal Deep Learning in Biomedical Applications'' illustrates the increasing application of these methodologies in this field.
Furthermore, we notice from the progression of publications in ``Multimodal Deep Learning'' to  ``Multimodal Deep Learning via Intermediate Fusion'' suggests that intermediate fusion is an emerging and specialized area within the broader field of MDL. 
The fact that only $2\%$ publications are categorized under ``Multimodal Deep Learning in Biomedical Applications via Intermediate Fusion'' indicates that this area, despite its promising applications, remains relatively underdeveloped.
This gap underscores the need for further research and exploration. 
These results justify the necessity of your systematic review, which aims to lay the foundational groundwork by defining further research and development in this promising field.

\subsection{Study Selection}

The study selection process for this systematic review was meticulously designed to ensure the inclusion of the most relevant and high-quality studies focusing on intermediate fusion in MDL within biomedical applications.
This multi-stage process involved initial screening and full-text assessment.

\begin{enumerate}
    \item \textbf{Initial Screening}: Initially, we reviewed the titles and abstracts of articles identified from the search. Articles that clearly did not meet the inclusion criteria, described in~\ref{sec:inclusion_exclusion}, were excluded at this stage.
    \item \textbf{Full-text Assessment}: We retrieved the full texts of articles that seemed potentially relevant based on the initial screening. Each full-text article was assessed for eligibility, focusing on whether they adhered to the inclusion and exclusion criteria. Articles that did not meet the criteria, described in~\ref{sec:inclusion_exclusion}, after a full-text review, were excluded. To ensure reliability and minimize bias, this process involved two independent reviewers, among the authors.
\end{enumerate}

\begin{figure}[t]
\centering
\includegraphics[width=0.6\textwidth]{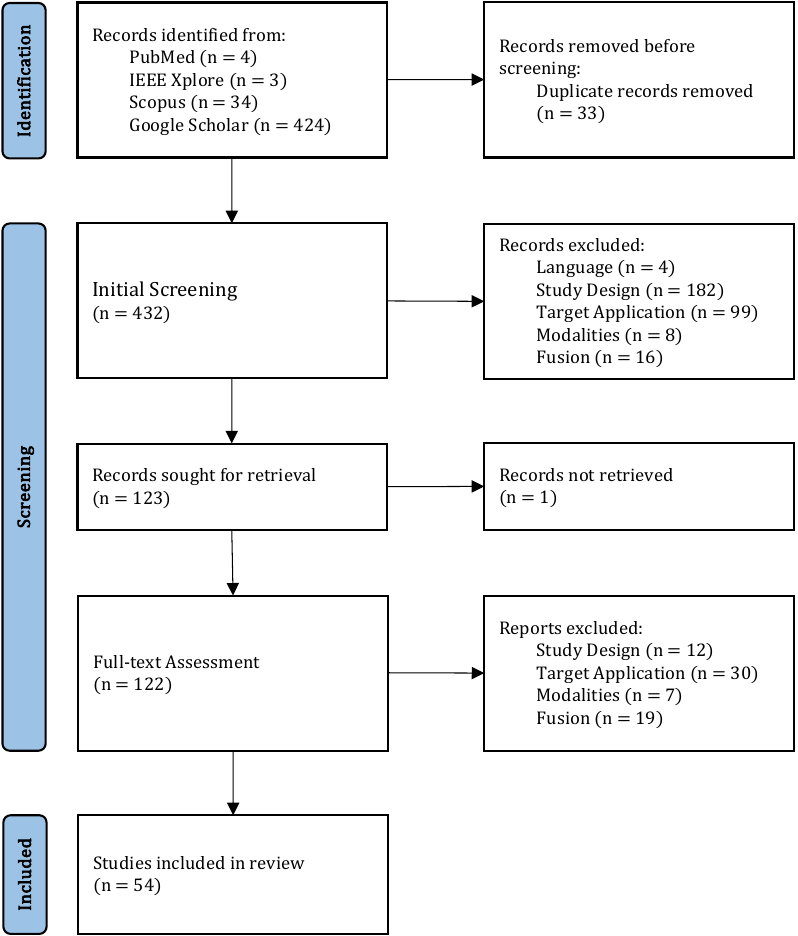}
\caption{PRISMA Flow Chart of Literature Selection Process. This flow chart outlines the systematic process of screening and selecting studies for inclusion in the review. It details the number of records identified, included, and excluded at each stage of the search and selection process, from initial database search through to the final included studies.}
\label{fig:prisma}
\end{figure}

In cases of discrepancies regarding the eligibility of an article, these were resolved through discussion.
We maintained a record of the reasons for excluding articles during the full-text assessment phase. To systematically track the number of articles at each stage of the selection process, a PRISMA flow diagram, shown in \figurename~\ref{fig:prisma}, was utilized.
Following the described steps, we resulted in 54 manuscripts related to MDL in biomedical applications via intermediate fusion.

\section{Intermediate Fusion in Biomedical Applications} \label{sec:main_body}

In this section, we present the various components that constitute a model employing intermediate fusion, as represented in \figurename~\ref{fig:joint_fusion}.
These components are presented in the order they appear in the model: beginning with the \textit{modalities} as inputs (Section~\ref{sec:modalities}), followed by the \textit{unimodal module} that extracts unimodal features (Section~\ref{sec:unimodal_module}), the \textit{fusion module} that combines these features (Section~\ref{sec:fusion_module}), the \textit{multimodal module} that processes the fused features further (Section~\ref{sec:multimodal_module}), and finally the \textit{target} as the output of the model (Section~\ref{sec:target}).
Then we also analyze the \textit{learning} strategies used to train the model(Section~\ref{sec:learning}).
Each of these components is detailed in the corresponding subsections, where we introduce their taxonomies and review their state-of-the-art applications in biomedicine.

\subsection{Definitions}\label{sec:definitions}

Intermediate fusion, as already introduced in Section~\ref{sec:Introduction}, within the domain of MDL, is an approach that involves extracting features from different modalities using specialized unimodal neural networks, and then merging these features into a fused multimodal representation.
This fused representation is subsequently fed into another neural network to yield the final outcome.
A distinctive feature of intermediate fusion is that, via the back-propagation of the loss, both the unimodal and multimodal modules are trained.
This allows for continuous improvement in capturing the intra- and inter-relationships inherent in multimodal data.
This methodology is particularly useful in complex fields like biomedicine.

\begin{figure}[t]
\centering
\includegraphics[width=\textwidth]{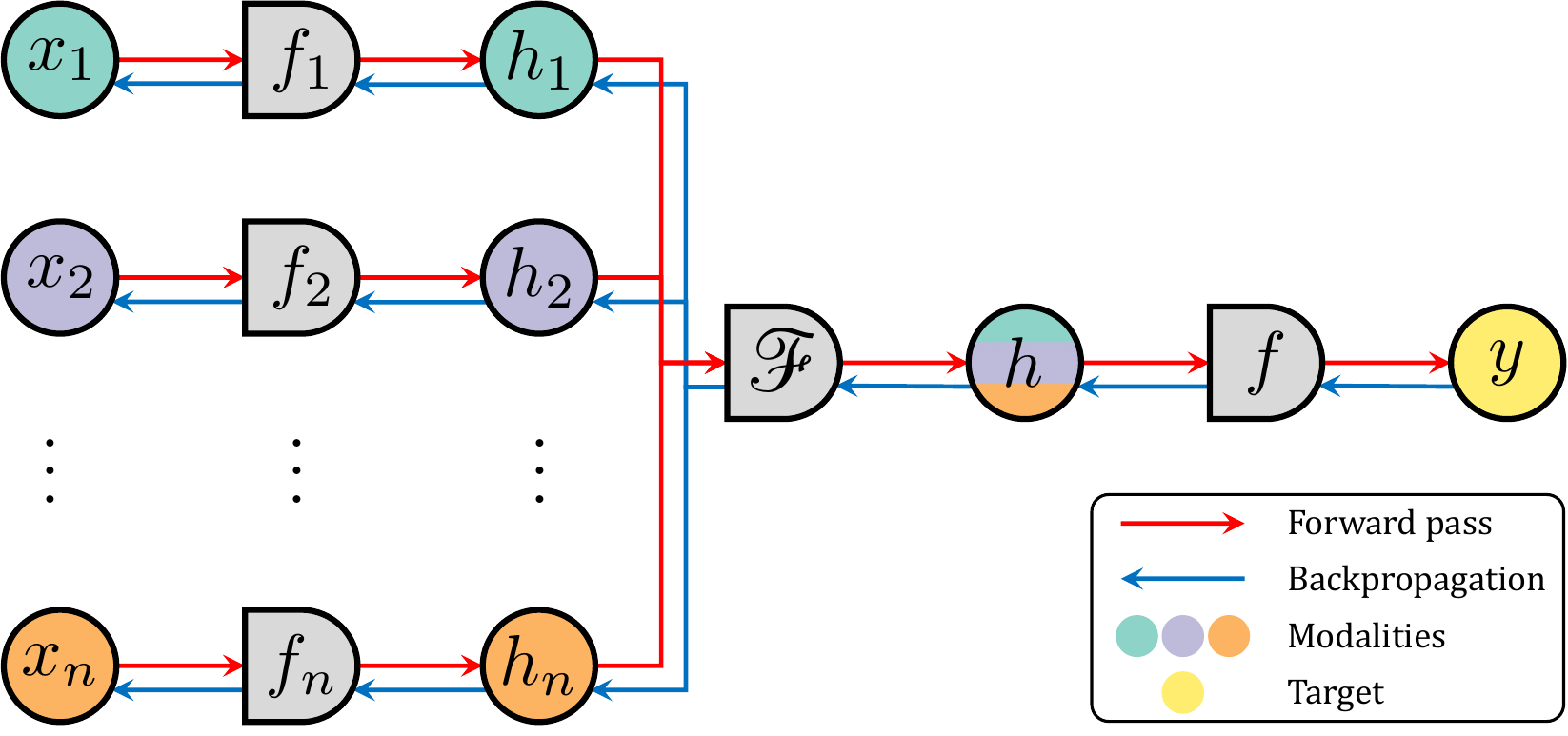}
\caption{Schematic Representation of Intermediate Fusion in MDL. This diagram illustrates the intermediate fusion approach within the framework of MDL. It begins with various data modalities \(x_1, x_2, \ldots, x_n\) each represented in distinct colors (green, purple, orange), progressing through specialized unimodal modules \(f_1, f_2, \ldots, f_n\) where individual features \(h_1, h_2, \ldots, h_n\) are extracted, maintaining the color coding of their respective modalities. These features are then integrated by the fusion module \(\mathscr{F}\), resulting in a multimodal feature representation \(h\) that is a blend of all the input colors, signifying the fusion of modal features. The subsequent multimodal module \(f\) processes these integrated features, aiming to achieve the specified target \(y\) shown in yellow. Red arrows indicate the forward pass through the network, i.e., the inference phase, while blue arrows represent the backpropagation, facilitating the training of both unimodal and multimodal components.}
\label{fig:joint_fusion}
\end{figure}

This method, as shown in \figurename~\ref{fig:joint_fusion}, can be broken down into several key components: modalities, unimodal modules, fusion module, multimodal module, and the target.
Each of these plays a critical role in the overall functionality and efficacy of the fusion process:

\begin{itemize}
    \item \textbf{Modalities}: Modalities, denoted as $x_1, x_2, \ldots, x_n$ refer to the different types of data or sources of information used in the model. In the context of biomedical applications, these could include imaging data, textual or tabular clinical notes, genetic information, or other relevant medical measurements.
    \item \textbf{Unimodal Module}: The unimodal modules, denoted as $f_1, f_2, \ldots, f_n$  refer to the parts of the network that processes each modality independently. Here, deep learning models are used to extract features, denoted as $h_1, h_2, \ldots, h_n$, specific to each modality.
    \item \textbf{Fusion Module}: The fusion module  $\mathscr{F}$ is the core of intermediate fusion, which takes the features extracted by unimodal modules and combines them into a unified multimodal representation, denoted as $h$. This fusion can be performed using various techniques like concatenation, averaging, or more complex operations that account for interactions between modalities.
    \item \textbf{Multimodal Module}: After fusion, the combined features are fed into the multimodal module $f$. This part of the network processes the fused representation to extract insights that are only apparent when considering all modalities together.
    \item \textbf{Target}: The target $y$ refers to the  output of the network. In biomedical applications, this could be a diagnosis, prognosis, or any other relevant medical outcome. The target guides the choice of the loss function, which in turn influences the back-propagation to iteratively improve the layers part of the unimodal, fusion and multimodal modules.
\end{itemize}

The subsequent sections will delve into the specifics of each component, analyzing their functions and significance in the context of intermediate fusion for biomedical applications.

\subsection{Modalities} \label{sec:modalities}
MDL with intermediate fusion represents a rapidly growing field in Artificial Intelligence (AI), where information from multiple modalities is integrated at intermediate stages of a deep neural network.
Each modality provides unique information, enriching the model's comprehension of context and thereby enhancing its performance.
The different types of data offer complementary perspectives on complex medical scenarios, and their fusion provides a deeper understanding.
For instance, medical image analysis can be integrated with textual data extracted from medical reports or laboratory data to obtain a more comprehensive view of a medical condition. 
Consequently, integrating multiple biomedical modalities strengthens predictive accuracy and fortifies models against input data fluctuations, amplifying their reliability and utility in practical applications.

\subsubsection{Type of Modalities} \label{sec:type_modalities}
From our analysis of the 54 articles included in our survey, we identified 6 macro-modalities:
\begin{itemize}
    \item \textbf{Imaging}: this modality encompasses various medical images that serve crucial roles in both diagnosis and ongoing monitoring of medical conditions. X-rays, Magnetic Resonance Imaging (MRI), Computed Tomography (CT), Positron Emission Tomography (PET), Ultrasounds (US) are included in this category.
    \item \textbf{Tabular}: under this modality fall tabular clinical data (spanning from demographic information to laboratory measurements), genomics data (which are used to analyze disease risk, genetic predisposition, or treatment response) and data related to drugs chemical features to analyze their functionality.
    \item \textbf{Text}: textual data comprises free-text reports and transcripts documenting question-answer dialogues between healthcare professionals and patients.
    \item \textbf{Time Series}: this category encompasses various types of signals.
    Specifically, we distinguish between electrical signals, clinical time series, and tactile signals. Electrical signals are derived from electrocardiograms (ECG), electroencephalograms (EEG), electroglottographs (EGG), electromyography (EMG), and electrodermal activity (EDA).
    Clinical time series typically consists of measurements recorded during patients' hospital stays, which may include vital signs, laboratory results, and other metrics.
    Tactile signals, utilized in a single article~\cite{bib:05_li2022aviper}, are recorded using a specialized tactile glove equipped with sensors.
    \item \textbf{Video}: video data can be related to imaging techniques over time, incorporating techniques like Temporal Enhanced Ultrasounds (TeUS) and Dynamic Contrast-Enhanced Ultrasonography (DCE-US), but it can also include diverse RGB videos capturing movements of hands or the entire body.
    \item \textbf{Audio}: it includes speech but also audio of coughing and breathing.
\end{itemize}

\begin{figure}[t]
\centering
\includegraphics[width=\textwidth]{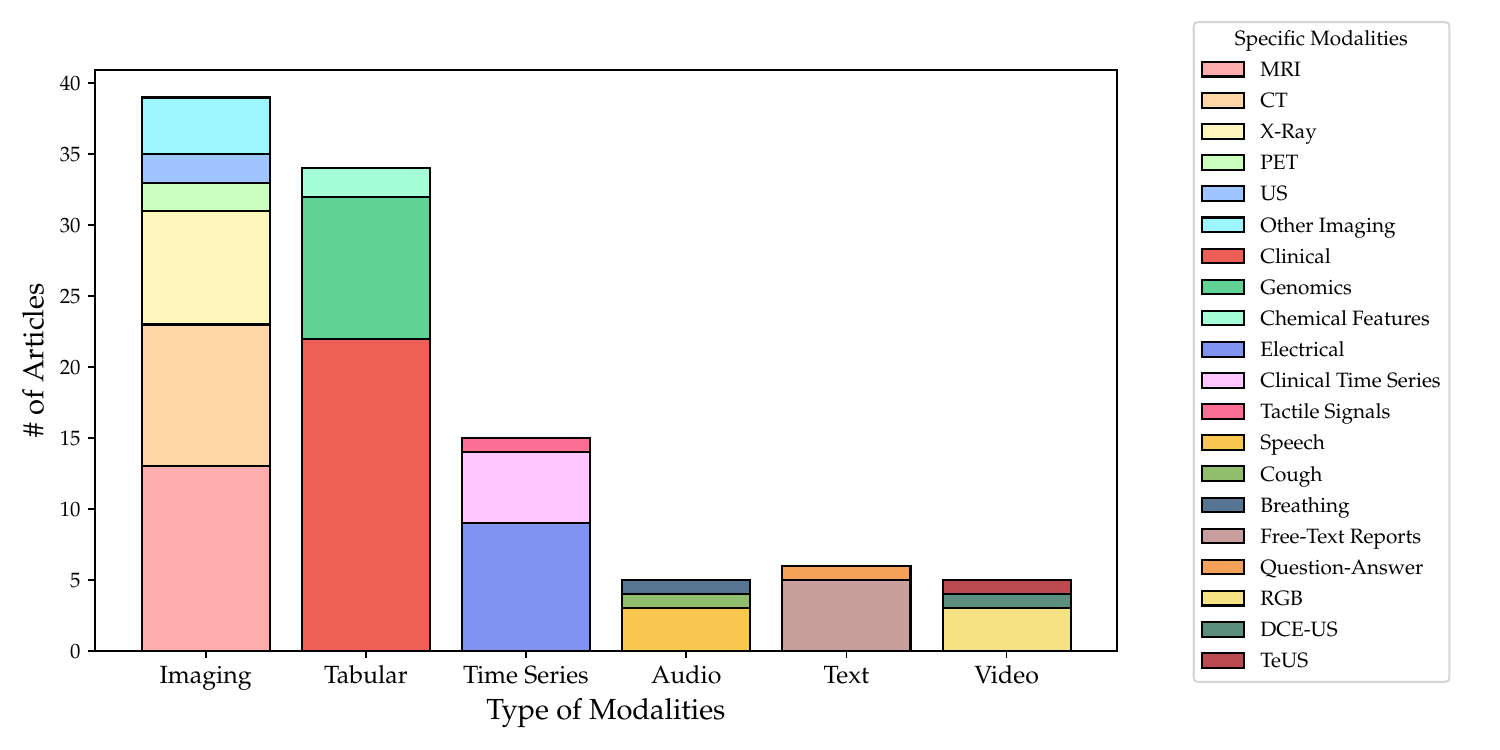}
\caption{Stacked bar plot showing the type of modalities used in the analyzed articles. The bar for each macro-modality is segmented, providing information on the specific modalities included.}
\label{fig:type_modalities}
\end{figure}

\figurename~\ref{fig:type_modalities} illustrates the distribution of the aforementioned modalities used in the analyzed articles.
It is observed that imaging and tabular data constitute the most frequently used modalities, representing 37\% and 35\% of the total modalities, respectively.
Among imaging modalities, MRI scans are the most prevalent, accounting for 33\% of cases, including variations such as functional MRI (fMRI) and structural MRI (sMRI).
Following MRI, CT scans account for 26\%, X-rays for 21\%, and PET scans and US both for 5\%.
Other types of imaging data, including histopathology images, microflow imaging, 2D fundus images, 3D optical coherence tomography scans, and skin lesion images, are less frequently used and collectively account for 10\% of the imaging modality.
Among the tabular modalities, the vast majority is represented by clinical data at 65\%, followed by genomics data at 29\% and chemical features at 6\%.
The prevalence of imaging and tabular modalities reflects the wide availability of diagnostic images and tabular data in the medical context, which we will also see in the publicly available datasets described in Section~\ref{sec:data_source}.
Imaging technologies are crucial for diagnosing and monitoring various conditions, as they are well-integrated into medical practice, with established protocols and trained professionals to interpret the results. 
Tabular data are structured information essential for a comprehensive assessment and management of patients and for personalized medicine.
Furthermore, it can be noted that time series represent 15\% of modalities, suggesting a growing interest in temporal data analysis in the medical field.
In particular, 60\% of time series are represented by electrical signals, 33\% by clinical time series, and the remaining 7\% by tactile signals.
On the other hand, text, audio, and video represent smaller percentages, 5\%, 4\%, and 4\%, respectively, indicating a lower availability of data in these forms or a lower applicability in the considered medical contexts.
Indeed, while textual data is prevalent and highly informative in healthcare, it often requires substantial preprocessing to be useful for deep learning due to its unstructured nature and variability in terminology.
Audio data is less frequently collected and stored compared to other data types.
Although the use of audio in medical diagnostics shows promise for specific applications, such as the automated diagnosis of respiratory conditions from breath sounds~\cite{bib:liang2022foundations}, it is not yet as universally applicable or validated as other modalities like imaging.
Video data requires significant storage and processing capabilities and is often specific to certain fields, such as surgery.
In many medical contexts, other data types provide sufficient information for diagnosis and treatment, thereby reducing the reliance on video.
Specifically, in text, audio, and video modalities, the most represented categories are free-text reports at 83\%, speech at 60\%, and RGB video at 60\%.

In the articles we have analyzed, various modalities are combined through intermediate fusion techniques, enabling deep learning models to achieve a more comprehensive understanding of medical contexts.
Section~\ref{sec:modalities_combination} will explore the different combinations of modalities utilized in these articles.

\subsubsection{Data Origin} \label{sec:data_origin}
Given the importance of modalities for MDL in biomedicine, we analyzed the origin of the datasets used in each of the 54 articles included in our survey.
We specifically distinguished whether the data was real or synthetic and this analysis reveals a significant trend: a substantial majority of the studies rely on real data, while only a marginal proportion, specifically 2 out of 54 articles~\cite{bib:42_shetty2023multimodal, bib:50_sun2023toward}, also incorporate synthetic data generated by generative adversarial networks for models' training.
This observation aligns with the expectation that real data are typically reliable and representative of reality, especially in the biomedical domain.
At the same time, it underscores that synthesizing data using generative adversarial networks still poses an open technical challenge that researchers may choose to address to bridge the gap in big data in the biomedical domain.

\subsubsection{Data Source} \label{sec:data_source}
The aforementioned real data sources mostly come from publicly available datasets.
Indeed, only about 30\% of the articles make use of private datasets.
The presence of a significant number of articles utilizing public datasets suggests that accessible and usable data resources exist for research in this field, enabling researchers to conduct studies and develop models without necessarily having to create or acquire private data.
Acquiring, annotating, and managing private datasets can be costly and time-consuming, and their use may sometimes be limited by privacy concerns.
In contrast, utilizing existing public datasets can be a convenient and practical solution for many researchers, further promoting comparability and reproducibility in research.

A total of 52 distinct public datasets are used in the analyzed articles, with \tablename~\ref{tab:dataset} presenting the ones used in at least two articles.
Detailed information on public datasets used in only one article can be found in Supplementary Material A.

The datasets listed in \tablename~\ref{tab:dataset} primarily include modalities within the macro-categories of imaging, e.g., CT scans, X-rays, and MRI, and tabular data, e.g., clinical and genomics data.
As we explored in Section~\ref{sec:type_modalities}, these are the most commonly utilized modalities in the analyzed articles, given their prevalence in the medical domain.
The table shows that the most commonly used datasets, e.g., COBRE and MIMIC-CXR, appear in only 4 articles. This clearly indicates the absence of a benchmark dataset utilized in the majority of studies, making it difficult to establish the overall effectiveness of the models.

\begin{table}[t]
    \centering
    \resizebox{\linewidth}{!}{
    \begin{tabular}{|l|l|l|l|l|}
    \hline
    \textbf{Dataset} & \textbf{\# of Articles}  & \textbf{Modalities} & \textbf{Used Modalities} & \textbf{Availability} \\ \hline
    COBRE~\cite{cobre} & 4 & \begin{tabular}[c]{@{}l@{}}fMRI, sMRI,\\ clinical data,\\ genomics data\end{tabular} & \begin{tabular}[c]{@{}l@{}}{\cite{bib:14_rahaman2023deep, bib:31_rahaman2021multi, bib:53_rahaman2022two}}: fMRI, sMRI, genomics data\\ {\cite{bib:04_liu2022attention}}: fMRI, sMRI, clinical data\end{tabular} &     {\href{https://fcon_1000.projects.nitrc.org/indi/retro/cobre.html}{link}}
    \\ \hline
    MIMIC-CXR~\cite{mimic_cxr} & 4 & \begin{tabular}[c]{@{}l@{}}chest X-ray,\\ free-text reports\end{tabular} & \begin{tabular}[c]{@{}l@{}}{\cite{bib:06_lobantsev2020comparative, bib:26_zeng2022miftp}}: chest X-ray, free-text reports\\ {\cite{bib:25_hayat2022medfuse}}: chest X-ray\\ {\cite{bib:46_kohankhaki2022radiopaths}}: chest X-ray, free-text reports \end{tabular} & {\href{https://physionet.org/content/mimic-cxr/2.0.0/}{link}}\\ \hline
    ADNI~\cite{adni} & 3 & \begin{tabular}[c]{@{}l@{}}MRI,\\ clinical data,\\ genomics data\end{tabular} & {\cite{bib:24_ostertag2023long, bib:43_ostertag2020predicting, bib:50_sun2023toward}}: MRI, clinical data & {\href{https://adni.loni.usc.edu/data-samples/access-data/}{link}} \\ \hline
    ASCERTAIN~\cite{ascertain} & 3 & \begin{tabular}[c]{@{}l@{}}electrical signals (EEG, ECG, EDA),\\ facial activity data\end{tabular} & {\cite{bib:13_radhika2021deep, bib:40_kuttala2023multimodal, bib:48_radhika2021stress}}: electrical signals (ECG, EDA) & {\href{https://ascertain-dataset.github.io/}{link}} \\ \hline
    CLAS~\cite{clas} & 3 & \begin{tabular}[c]{@{}l@{}}electrical signals (ECG, EDA),\\ photoplethysmography signals,\\ accelerometer signals\end{tabular} & {\cite{bib:13_radhika2021deep, bib:40_kuttala2023multimodal, bib:48_radhika2021stress}}: electrical signals (ECG, EDA) & {\href{https://www.wwwsensornetworkslab.com/clas}{link}} \\ \hline
    fBIRN~\cite{fbirn} & 3 & \begin{tabular}[c]{@{}l@{}}fMRI, sMRI,\\ genomics data\end{tabular} & {\cite{bib:14_rahaman2023deep, bib:31_rahaman2021multi, bib:53_rahaman2022two}}: fMRI, sMRI, genomics data & {\href{http://fbirnbdr.nbirn.net:8080/BDR}{link}} \\ \hline
    MRPC~\cite{mrpc} & 3 & \begin{tabular}[c]{@{}l@{}}fMRI, sMRI,\\ genomics data\end{tabular} & {\cite{bib:14_rahaman2023deep, bib:31_rahaman2021multi, bib:53_rahaman2022two}}: fMRI, sMRI, genomics data & on request  \\ \hline
    NSCLC-Radiogenomics~\cite{nslc_radiogenomic} & 3 & \begin{tabular}[c]{@{}l@{}}CT, PET/CT,\\ genomics data,\\ clinical data\end{tabular} & {\cite{bib:09_hou2023deep, bib:28_wang2021modeling, bib:38_subramanian2020multimodal}}: CT, genomics data & {\href{https://www.cancerimagingarchive.net/collection/nsclc-radiogenomics/}{link}} \\ \hline
    CPTAC-PDA~\cite{cptac-pda} & 2 & \begin{tabular}[c]{@{}l@{}}US, CT, MRI, PET,\\ histopathology images,\\ clinical data\end{tabular} & {\cite{bib:07_menegotto2021computer, bib:08_menegotto2020computer}}: CT, clinical data & {\href{https://www.cancerimagingarchive.net/collection/cptac-pda/}{link}} \\ \hline
    MIMIC III~\cite{mimic-iii} & 2 & \begin{tabular}[c]{@{}l@{}}clinical data,\\ clinical time series,\\ free-text reports\end{tabular} & \begin{tabular}[c]{@{}l@{}}{\cite{bib:12_niu2023deep}}: free-text reports, clinical time series\\ {\cite{bib:44_ma2023predicting}}: clinical time series, clinical data \end{tabular} & {\href{https://physionet.org/content/mimiciii/1.4/}{link}} \\ \hline
    MIMIC-IV~\cite{mimc-iv} & 2 & \begin{tabular}[c]{@{}l@{}}clinical data,\\ clinical time series\end{tabular} & \begin{tabular}[c]{@{}l@{}}{\cite{bib:25_hayat2022medfuse}}: clinical time series;\\ {\cite{bib:46_kohankhaki2022radiopaths}}: clinical data \end{tabular} & {\href{https://physionet.org/content/mimiciv/1.0/}{link}} \\ \hline
    TCGA-KIRP~\cite{tcga-kirp} & 2 & \begin{tabular}[c]{@{}l@{}}CT, MRI, PET,\\ clinical data\end{tabular} & {\cite{bib:07_menegotto2021computer, bib:08_menegotto2020computer}}: CT, clinical data & {\href{https://www.cancerimagingarchive.net/collection/tcga-kirp/}{link}} \\ \hline
    TCGA-LIHC~\cite{tcga-lihc} & 2 & \begin{tabular}[c]{@{}l@{}}CT, MRI, PET,\\ clinical data\end{tabular} & {\cite{bib:07_menegotto2021computer, bib:08_menegotto2020computer}}: CT, clinical data & {\href{https://www.cancerimagingarchive.net/collection/tcga-lihc/}{link}} \\ \hline
    TCGA-STAD~\cite{tcga-stad} & 2 & \begin{tabular}[c]{@{}l@{}}CT,\\ clinical data\end{tabular} & {\cite{bib:07_menegotto2021computer, bib:08_menegotto2020computer}}: CT, clinical data & {\href{https://www.cancerimagingarchive.net/collection/tcga-stad/}{link}} \\ \hline
    \end{tabular}}
    \caption{Publicly available datasets used at least in two articles on intermediate learning in biomedical research.}
    \label{tab:dataset}  
\end{table}

\subsubsection{Sample Size}
A crucial factor to consider when training MDL models with intermediate fusion techniques is the sample size of the dataset being used, i.e., the number of instances in a dataset.
Intermediate fusion involves combining different input modalities into a shared layer within a neural network to enhance the overall performance of the model.
This process is particularly complex and preferably requires large datasets so that the model has enough examples from each modality to learn the relationships between them effectively.
Additionally, since intermediate fusion utilizes large networks with numerous weights, it is unsuitable for datasets with limited samples.
The sample sizes of the datasets involved in the articles vary as follows: less than 100 samples for 7 datasets, 101 to 500 samples for 24 datasets, 501 to 1000 samples for 14 datasets, 1001 to 1500 samples for 6 datasets, and more than 1501 samples for 27 datasets.

These findings underscore the prevalence of moderate-sized datasets (with 101-1000 samples) in MDL research.
However, a notable proportion (27 datasets) surpass the threshold of 1501 samples.
These larger datasets hold promise for training robust models capable of capturing diverse variations, potentially yielding more generalizable outcomes.
Conversely, 7 datasets comprise fewer than 100 samples, indicating that dealing with small datasets is still a challenge in MDL.
Small datasets may lead to overfitting, where models perform well on the training data but fail to generalize to unseen data.
Researchers working with small datasets may need to employ techniques such as data augmentation or transfer learning to mitigate this challenge.

\subsubsection{Class Distribution}
In addition to the necessity of employing large datasets, if one intends to tackle a classification problem with intermediate fusion techniques, it is equally important to have balanced datasets.
Balancing the diverse classes within the dataset is essential to prevent the model from being biased towards a particular class at the expense of others.
However, real-world data, especially in the medical field, often exhibit imbalanced distributions.
Since intermediate fusion methods are particularly sensitive to class imbalance, we further analyze the articles that tackle classification tasks to determine whether they use balanced datasets.
To this end, we employed the Gini-index~\cite{gini}, denoted as G, computed as:

\begin{equation}
G = 1 - \sum_{i=1}^{|C|}P(C_i)^2
\end{equation}
where $C$ is the set of classes, $|C|$ is the number of classes, and $P(C_i)$ is the class prior probability.
As a consequence, datasets can be defined as:
\begin{itemize}
    \item \textbf{Balanced}: if $0.48 < G \leq 0.5$;
    \item \textbf{Imbalanced}: if $0.42 <  G \leq 0.48$;
    \item \textbf{Highly Imbalanced}: if $0 \leq G \leq 0.42$.
\end{itemize}
As an example, in a binary classification problem, these ranges of Gini-index correspond to having a balanced dataset when the class priors range from 50-50 to 41-59, an imbalanced dataset when the class priors range from 40-60 to 31-69, and a highly imbalanced dataset when the class priors range from 30-70 to 0-100.

\begin{figure}[t]
\centering
\includegraphics[width=\textwidth]{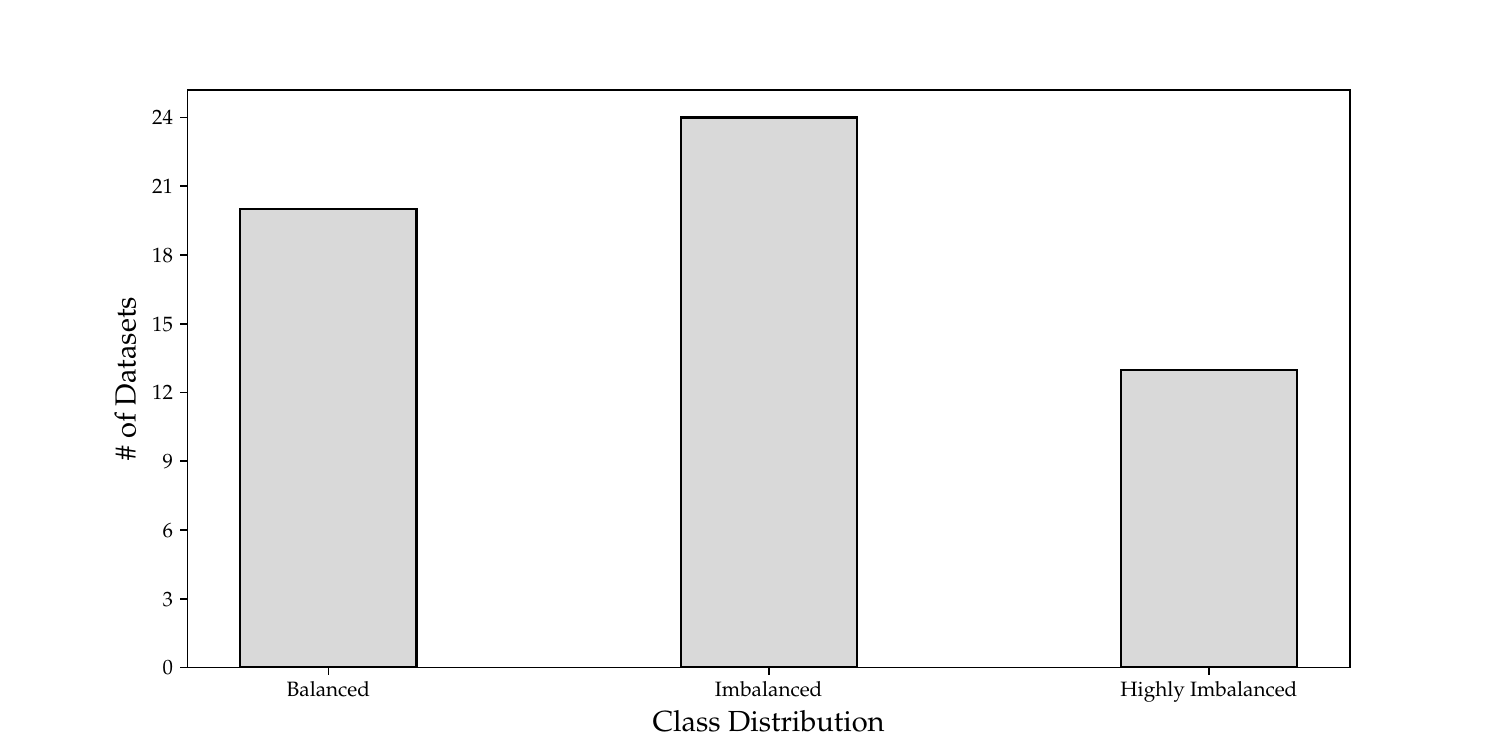}
\caption{Distribution of Balanced, Imbalanced, Highly Imbalanced datasets.}
\label{fig:class_distribution}
\end{figure}

\figurename~\ref{fig:class_distribution} graphically shows the distribution of balanced, imbalanced, and highly imbalanced datasets used in the reviewed articles.
It is worth noting that the combined number of imbalanced and highly imbalanced datasets, 24 and 13, respectively, exceeds the total number of balanced datasets, which stands at 20.
Only 37\% of articles that use imbalanced or highly imbalanced datasets address the issue of class imbalance through the use of specific techniques.
The preferred ones include the use of the Synthetic Minority Over-sampling Technique~\cite{smote} as employed in~\cite{bib:13_radhika2021deep, bib:40_kuttala2023multimodal, bib:48_radhika2021stress, bib:45_seo2021predicting}, or the use of specific loss functions such as the Label-Distribution-Smooth Margin loss~\cite{ldsm_loss} used by~\cite{bib:01_li2022bi}, the weighted version of the Binary Cross-Entropy Loss used by~\cite{bib:37_cahan2023multimodal}, and the Focal Loss~\cite{focal_loss} used by~\cite{bib:46_kohankhaki2022radiopaths, bib:54_cahan2022weakly}.
However, the presence of studies based on unbalanced datasets highlights the need to pay more attention to this phenomenon.
Without such caution, there is a risk of training models that are less robust, characterized by biases in their predictions and overfitting on dominant classes.

\subsubsection{Missing Modalities}
A common challenge in biomedical multimodal data is the presence of one or more missing modalities, an issue emerging when certain expected data modalities for a given sample are unavailable.
This can happen due to various factors, such as registration errors, technical limitations in capturing specific data types, or simply because certain modalities were not collected for all the patients in the dataset.
In the context of MDL, it is essential to handle the presence of missing modalities appropriately, as this can pose difficulties in the data processing for the model.
Despite the frequency of datasets with missing modalities in the biomedical field~\cite{bib:missing-modality-healthcare}, only 4 articles~\cite{bib:25_hayat2022medfuse, bib:46_kohankhaki2022radiopaths, bib:43_ostertag2020predicting, bib:44_ma2023predicting} deal with datasets where a modality is missing to a certain extent, developing models specifically designed to handle this situation. More details are available in Section~\ref{sec:missing}.
Conversely, 6 articles addressed the issue through imputation techniques for missing modalities in the datasets.
The imputation techniques employed include the use of neural networks and machine learning algorithms, as in~\cite{bib:07_menegotto2021computer, bib:08_menegotto2020computer}, or the replacement of missing values with mean or predefined values, as in~\cite{bib:12_niu2023deep, bib:45_seo2021predicting, bib:16_holste2021end, bib:32_bhattacharya2022multi}.
The remaining articles, however, do not address the issue of missing modalities: they use datasets without missing modalities or remove samples with missing modalities.

\subsubsection{Modalities Combination} \label{sec:modalities_combination}
\begin{figure}[t]
  \centering
  \resizebox{\textwidth}{!}{
  \begin{tikzpicture}
    \node (a) at (0,0) {(a)};
    \node (img1) [above left=.5cm and -2.6cm of a] {\includegraphics[height=5cm, trim={2cm 2cm 4cm 2cm}]{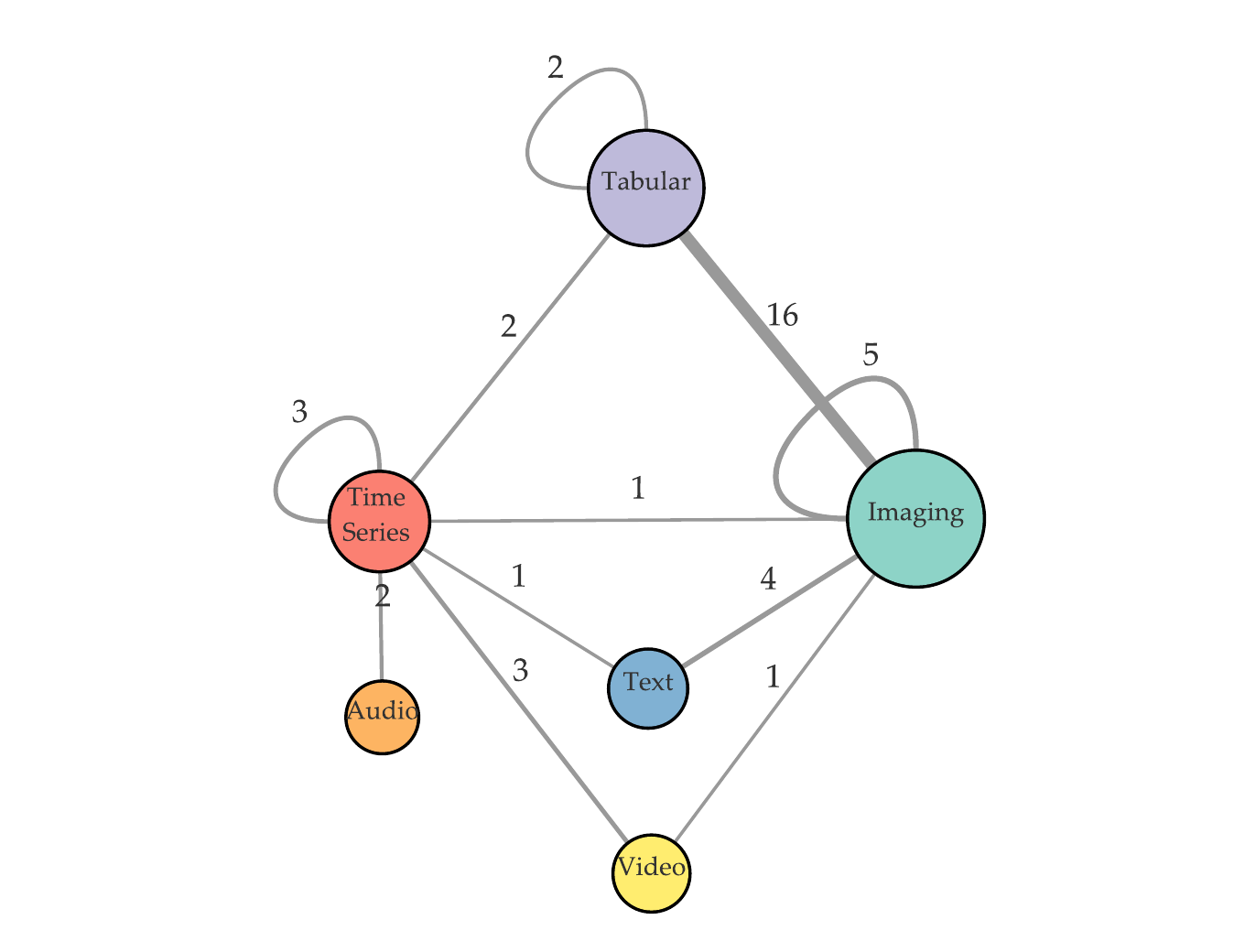}};

    \node (b) [right=5.5cm of a] {(b)};
    \node (img2) [above left=.5cm and -4cm of b]  {\includegraphics[height=5cm, trim={2cm 2cm 2cm 2cm}]{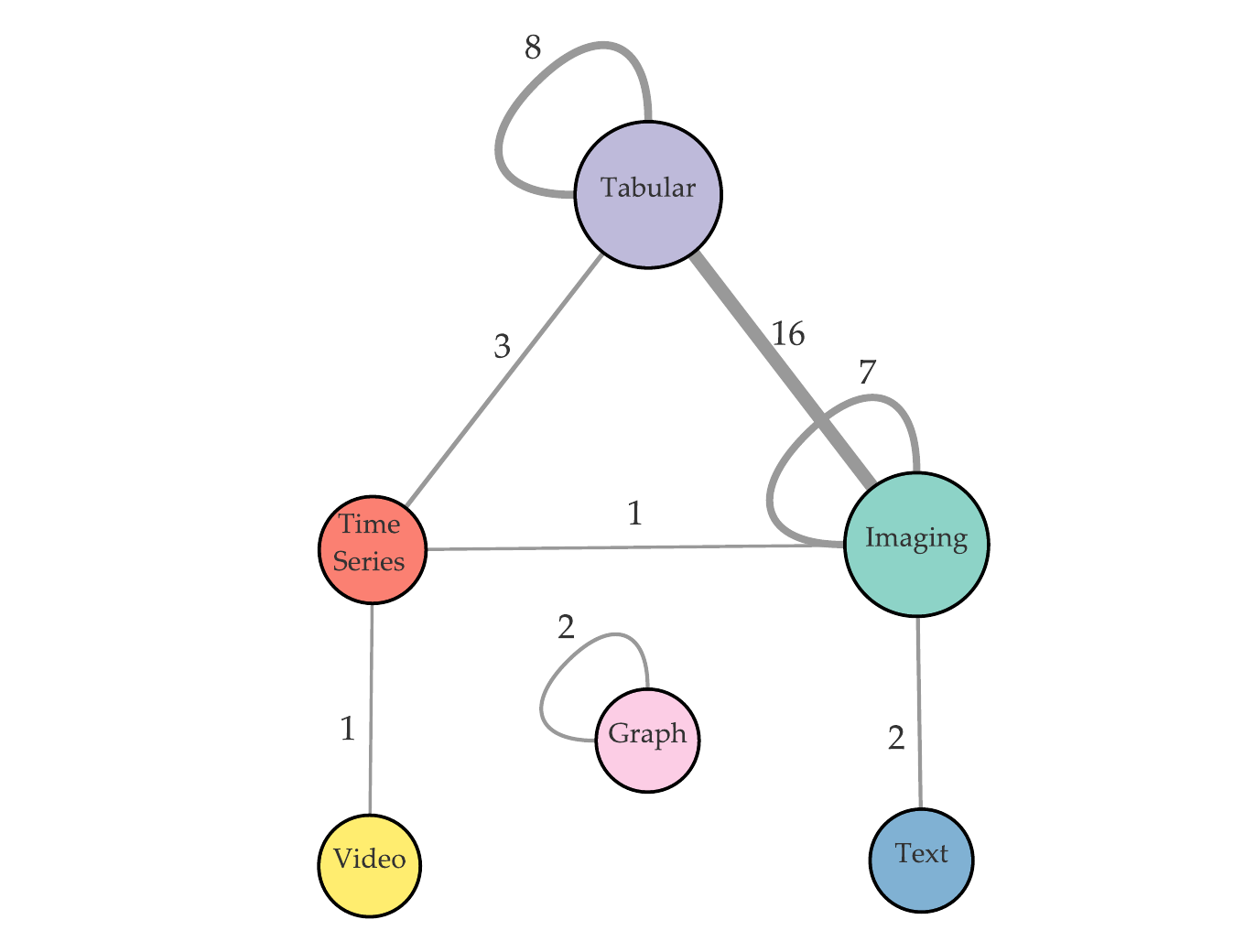}};
  \end{tikzpicture}}
  \caption{Macro-modalities combinations in bimodal articles before (a) and after (b) feature extraction. The numbers on the edges represent the frequency of articles in which two modalities, indicated as nodes, are employed together. The width of the edges is drawn proportional to the corresponding number of articles. The sizes of the nodes are also arranged according to the number of articles that the modalities are used for.}
  \label{fig:bimodal_modalities}
\end{figure}

We define the multimodal dimension characterizing the studies exploiting the number of modalities provided as input to deep learning models.
We note that a significant proportion of studies (74\% of articles) utilize bimodal datasets, indicating a prevalent trend in research.
Additionally, 20\% of the articles employ trimodal datasets, while a smaller fraction, only 6\%, engage with high-modal datasets, characterized by the incorporation of more than three modalities.
\figurename~\ref{fig:bimodal_modalities} (a) illustrates the macro-modalities combinations within the datasets used in the bimodal articles.
The numbers on the edges represent the frequency of articles in which two modalities, indicated as nodes, are employed together.
The width of the edges is drawn proportional to the corresponding number of articles.
The sizes of the nodes are also arranged according to the number of articles that the modalities are used for. 
We observe that the most prevalent combination is the imaging modality with the tabular modality, appearing in 16 articles, which accounts for 40\% of the bimodal articles.
Among these, the most common pairings involve CT scans and clinical data~\cite{bib:07_menegotto2021computer, bib:08_menegotto2020computer, bib:37_cahan2023multimodal, bib:54_cahan2022weakly, bib:47_sousa2023single}, followed by MRI scans with clinical data~\cite{bib:50_sun2023toward, bib:24_ostertag2023long, bib:43_ostertag2020predicting, bib:16_holste2021end}, and then CT scans with genomics data~\cite{bib:09_hou2023deep, bib:28_wang2021modeling, bib:38_subramanian2020multimodal}.
These findings align with expectations because, as outlined in Section~\ref{sec:type_modalities}, the imaging and the tabular are the most common macro-modalities and, within them, CT, MRI, clinical data, and genomics data are prevalent.
The integration of imaging and tabular data offers a more comprehensive view of patients, with images providing detailed visual information on anatomy and tissue structures, while tabular data may contain clinically relevant information not found in images.

\figurename~\ref{fig:bimodal_modalities} (a) shows that around 13\% of bimodal articles, i.e., 5 studies, utilize datasets combining two distinct types of imaging modalities.
They include the combination of different MRI variants~\cite{bib:19_he2021hierarchical, bib:23_pan2021liver}, CT and PET scans~\cite{bib:03_ahmad2023aatsn}.
It is reasonable to speculate that different imaging modalities can provide complementary information on anatomical and functional tissue characteristics; their combined use can improve diagnostic accuracy by more clearly identifying anomalies or areas of interest and, in a deep learning model, can enhance predictive capability.

Additionally, 10\% of the bimodal articles, i.e., 4 studies, utilize datasets combining the imaging and text modalities. Notably, X-rays paired with free-text reports is the most prevalent combination~\cite{bib:42_shetty2023multimodal, bib:06_lobantsev2020comparative, bib:26_zeng2022miftp}, providing direct visual information from X-rays alongside corresponding medical interpretations.

Moving on to the trimodal domain, the most recurring macro-modalities combination is the use of two imaging modalities and one tabular modality, encompassing 55\% of the trimodal articles.
Within this subset, studies predominantly combine different MRI variants with genomics~\cite{bib:14_rahaman2023deep, bib:31_rahaman2021multi, bib:53_rahaman2022two} or clinical data~\cite{bib:04_liu2022attention, bib:02_li2022dynamic}.
This trend is consistent with what happens in the bimodal domain, and the same observations apply.

Conversely, within the high-modal domain, no discernible trend emerges due to the limited number of articles falling into this category~\cite{bib:30_meng2022msmfn, bib:39_huang2020multimodal, bib:52_alsherbiny2021trustworthy}.

For more comprehensive insights into macro-modalities combinations and specific modalities combinations for each article, please refer to Supplementary Material A.

\subsubsection{Feature Extraction} \label{Features_Extraction}
As discussed in Section~\ref{sec:modalities_combination}, understanding how modalities are combined is fundamental. Equally crucial, however, is considering the application of feature extraction techniques to data before they are fed into the models.
Feature extraction is not part of the training process and should not be confused with the unimodal module, which will be described in Section~\ref{sec:unimodal_module}.
Feature extraction can play a pivotal role in preparing data for training deep learning models, involving the transformation of raw data into a set of representative features.
These features serve as the actual input for the unimodal models and may belong to a different category of modalities than the original data.
\figurename~\ref{fig:bimodal_modalities} (b) illustrates the combination of macro-modalities in bimodal articles following feature extraction.
Comparing it with \figurename~\ref{fig:bimodal_modalities} (a), with which \figurename~\ref{fig:bimodal_modalities} (b) shares the structure, it allows us to visualize the differences between before and after feature extraction.
The emergence of the graph modality following feature extraction is immediately evident.
In particular, there are 2 articles that, through the feature extraction process, transform the raw data, 
video and time series\cite{bib:17_cen2022exploring}, and audio and time series~\cite{bib:29_chen2022ms}, in the graph modality.
An interesting trend is also the transformation of modalities into the tabular type after the feature extraction process.
In fact, as noted from \figurename~\ref{fig:bimodal_modalities} (a), only 2 articles originally use datasets where both combined modalities are tabular.
However, as shown in \figurename~\ref{fig:bimodal_modalities} (b), the number of articles combining tabular data with other tabular data increases to 8 after feature extraction.
In particular, this type of transformation is adopted by the 3 articles that used datasets with the combination of the time series modalities~\cite{bib:13_radhika2021deep, bib:40_kuttala2023multimodal, bib:48_radhika2021stress}, by 2 articles that used an imaging and a tabular modality~\cite{bib:28_wang2021modeling, bib:38_subramanian2020multimodal}, and by one article that used a video and a time series combination~\cite{bib:10_njoku2022deep}.
Similarly, the text modality is converted into the tabular modality in 3 out of 5 articles~\cite{bib:06_lobantsev2020comparative, bib:12_niu2023deep, bib:42_shetty2023multimodal}, and this trend is also repeated in trimodal and high-modal articles.
In particular, as declared in Section~\ref{sec:modalities_combination}, in the trimodal field, the most frequently present combination of modalities in datasets is with two imaging modalities and one tabular modality but in 33\% of the cases~\cite{bib:14_rahaman2023deep, bib:31_rahaman2021multi}, after feature extraction, the data actually provided as input to the unimodal models all belong to the tabular modality.
In the high-modal field instead, having only 3 articles, it is not possible to define a trend of the most used combination of modalities but, in one case~\cite{bib:39_huang2020multimodal}, the original imaging modality used in combination with different tabular modalities is in turn converted to tabular before being provided as input to the unimodal model.
This trend shows increased confidence among researchers in working with tabular data.

\begin{figure}[t]
\centering
\includegraphics[width=\textwidth]{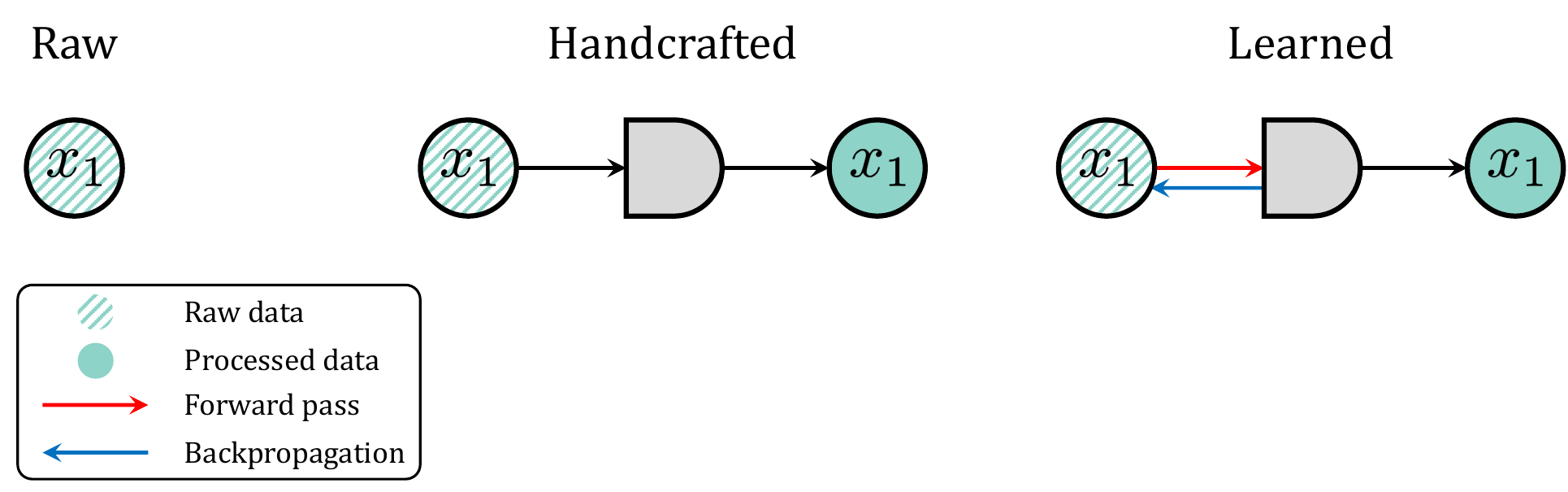}
\caption{Methods for extracting input features from various modalities. \textit{Raw}: No feature extraction techniques applied; the raw data is used directly. \textit{Handcrafted}: Features are manually engineered by experts to suit the problem domain, resulting in processed data used as input for the models. \textit{Learned}: Features are extracted by training a neural network or machine learning model, which processes the raw data through forward and backward passes, providing the processed input data for the models.}
\label{fig:feature_extraction_scheme}
\end{figure}

However, feature extraction isn't always employed, and in these cases, data is used in its raw form as input for the unimodal models.
Conversely, when feature extraction is applied, the input features for the unimodal models can be defined as handcrafted or learned, depending on the method of feature extraction utilized.
\figurename~\ref{fig:feature_extraction_scheme} shows the 3 ways in which input features can be provided to models.
In handcrafted feature extraction, features are typically engineered by human experts based on their understanding of the problem domain, which is crucial.
Conversely, in learned feature extraction, features are automatically learned from raw data using a neural network or other machine learning models.
In this case, instead of manually designing features, a model is trained with a forward and a backward pass to automatically discover relevant features from the data during the learning process.

\begin{figure}[t]
\centering
\includegraphics[width=\textwidth]{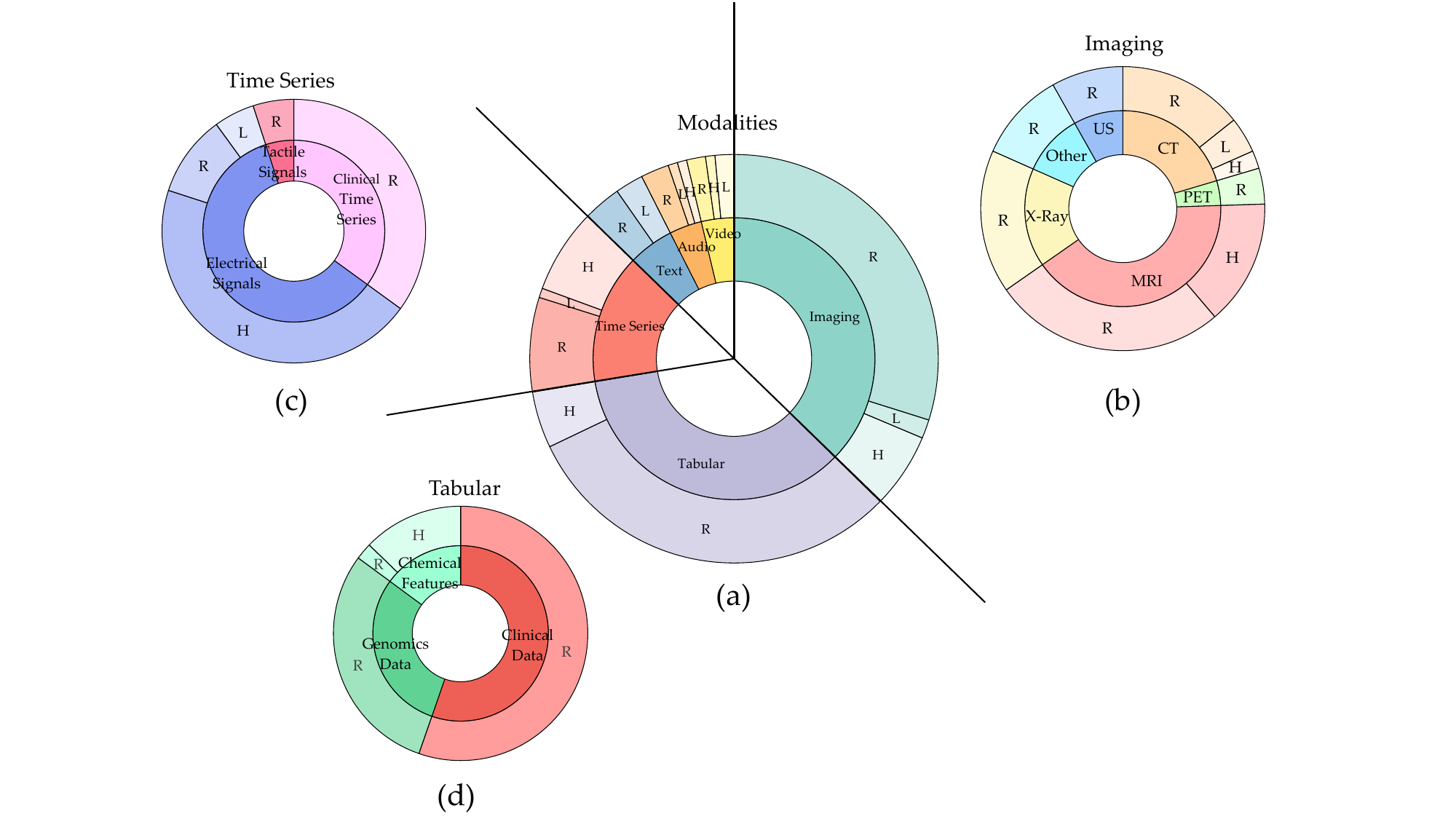}
\caption{Pie charts providing an overview of how feature extraction is performed for macro-modalities (a) and specific modalities belonging to imaging (b), time series (b), and tabular (c).
R indicates that data are used in their raw form, H indicates that handcrafted feature extraction is performed, and L indicates that learned feature extraction is performed.}
\label{fig:feature_extraction}
\end{figure}

The pie charts in \figurename~\ref{fig:feature_extraction} provide an overview of how feature extraction is performed for macro-modalities and for specific modalities belonging to the most represented macro-modalities, i.e., imaging, tabular, and time series.
\figurename~\ref{fig:feature_extraction} (a) illustrates the 6 macro-modalities $-$ imaging, tabular, time series, text, audio, and video $-$ within the datasets before feature extraction. 
Each modality is represented proportionally within the innermost circumference, while the outer circumference is segmented to reflect the portion of data used in its raw form (denoted with R), the portion subjected to handcrafted feature extraction (H), and the portion undergoing learned feature extraction (L).
Furthermore, for the 3 most represented macro-modalities $-$ imaging, time series, tabular $-$ \figurename~\ref{fig:feature_extraction} (b), (c), (d), respectively, show how feature extraction is applied to the specific modalities included in the macro ones.

Looking at \figurename~\ref{fig:feature_extraction}, the following emerges:
\begin{itemize}
    \item In the context of imaging macro-modality, raw data is utilized in 80\% of cases. In 16\% of cases, concerning CT~\cite{bib:28_wang2021modeling} and MRI~\cite{ bib:14_rahaman2023deep, bib:31_rahaman2021multi, bib:53_rahaman2022two, bib:04_liu2022attention}, handcrafted feature extraction is employed. Only in 4\% of cases, solely involving CT~\cite{bib:38_subramanian2020multimodal, bib:39_huang2020multimodal}, the features are learned.
    \item For the tabular macro-modality, in 87\% of cases, the used data are raw. In 13\% of cases, all related to chemical features~\cite{bib:52_alsherbiny2021trustworthy}, features extraction is handcrafted.
    \item For the time series macro-modality, raw data are used in 50\% of cases. Conversely, feature extraction is handcrafted in 45\% of cases~\cite{bib:13_radhika2021deep,bib:40_kuttala2023multimodal,bib:48_radhika2021stress, bib:17_cen2022exploring, bib:10_njoku2022deep, bib:27_mohammed2023mmhfnet} and features are learned in 5\% of cases~\cite{bib:29_chen2022ms}. The cases where data is not used in raw form exclusively involve electrical signals.
    \item For the text macro-modality, in 57\% of cases features are raw, while in 43\% of cases~\cite{bib:42_shetty2023multimodal,  bib:12_niu2023deep, bib:16_holste2021end} features are learned.
    \item For the video macro-modality, in 40\% of cases features are raw, in another 40\% of cases features are learned~\cite{bib:10_njoku2022deep, bib:20_sedghi2020improving} and in 20\% of cases~\cite{bib:17_cen2022exploring} features are handcrafted.
    \item For the audio macro-modality, in 60\% of cases features are raw, in 20\% of cases features are learned~\cite{bib:29_chen2022ms} and in another 20\% of cases features are handcrafted~\cite{bib:27_mohammed2023mmhfnet}.
\end{itemize}

Based on this evidence, we can state that the use of raw data is prevalent compared to the extraction of handcrafted or learned features.
This is particularly evident in imaging and tabular modalities, where 80\% and 87\% of cases, respectively, use raw data.
This suggests that deep learning models taking a single modality as input, described in Section~\ref{sec:unimodal_module}, can directly process this type of modality without the need for feature extraction.
For the remaining macro-modalities $-$ time series, text, audio, and video $-$ while the prevalence of raw data usage persists, the percentages are less skewed compared to imaging and tabular scenarios.
This underscores the fact that the distinct nature of the data may necessitate different processing approaches.
While imaging and tabular data lend themselves to more direct interpretation and benefit from raw data utilization, time series, text, audio, and video may require more extensive processing to extract relevant information.

After analyzing modalities, the following Section will provide a detailed examination of the deep learning models employed for processing unimodal data, prior to the intermediate fusion of the different modalities.

\subsection{Unimodal Module} \label{sec:unimodal_module}

The power of intermediate fusion lies in its ability to capture both intra- and inter-modal relationships inherent in multimodal data. 
In contrast to early fusion, each modality is processed through a dedicated branch denoted as $f_i$ in \figurename~\ref{fig:joint_fusion}, extracting modality-specific features, i.e., the marginal representations denoted as $h_i$ in the same figure. 
Extracting these marginal representations, facilitated by a global loss derived from fused features, allows for adaptability in relation to other modalities. 
This process empowers neural networks to derive representations that optimize the overarching goal.
The rest of this section reports on the analysis of the unimodal networks adopted in the literature, shedding light on the preferences and trends observed in the field.

\subsubsection{Architecture Types}
\label{sec:unimodal_architecture}

\begin{figure}[t]
\centering
\includegraphics[width=\textwidth]{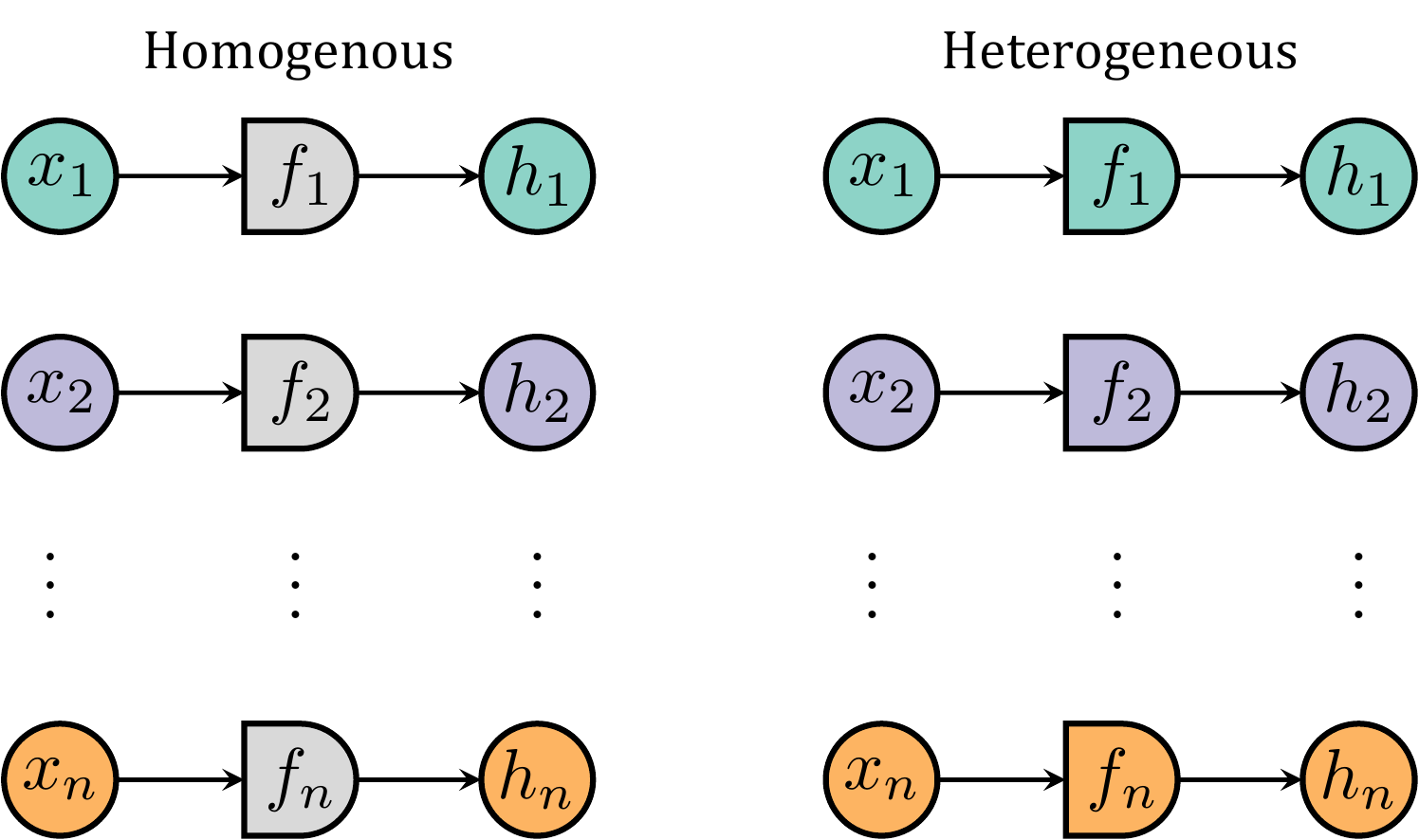}
\caption{Schematic representation of homogeneity: $f_1, f_2, ..., f_n$ represent unimodal modules and the colors represent the type of architecture/modality. Unimodal modules are homogenous when all the modules have the same type of architecture (as in the figure left), and heterogenous if there are at least two different types of architectures, modality-specific (as in the figure right).}
\label{fig:unimodal_homogeneity}
\end{figure}
The analysis of the literature suggests categorizing the reviewed papers into two classes: homogeneous and heterogeneous.
We say that a multimodal model is homogeneous when all constituent unimodal networks share a uniform type of architecture; conversely, we say it is heterogeneous when at least one network diverges from the others, as schematically depicted in \figurename~\ref{fig:unimodal_homogeneity}.
In our inspection of the 54 articles within the scope of this survey, we discerned that 24 articles exhibit homogeneity, signifying uniformity across their unimodal networks, whereas 30 articles manifest heterogeneity, indicating the presence of diverse network types. 
Therefore, it's challenging to identify a trend in uniformity, as this classification appears intricately linked to the specific characteristics of input modalities. 
With the same type of modality, unimodal modules tend to have the same type of architecture and vice versa.

We identified six distinct types of architectures employed as unimodal modules in the literature: 
\begin{itemize}
\item \textbf{Convolutional Neural Network (CNN)}: Specialized for processing data with a grid-like topology, such as images, using convolutional layers to capture spatial hierarchies.
\item \textbf{Fully Connected Neural Network (FCNN)}: Composed of layers in which each neuron is connected to every neuron in the previous and following layers, suitable for learning non-spatial hierarchies such as those in tabular data.
\item \textbf{Recurrent Neural Network (RNN)}: Designed for sequential data, with the ability to retain information from previous inputs by looping the output back into the network.
\item \textbf{Graph Neural Network (GNN)}: Operating on graph structures, they leverage node and edge information to capture complex relationships and interdependencies.
\item \textbf{Autoencoder (AE)}: They aim to compress input into a lower-dimensional representation and then reconstruct it, used for tasks like denoising or dimensionality reduction.
\item \textbf{Transformer (TF)}: They utilize self-attention mechanisms to weigh the significance of different parts of the input data differently, excelling in parallel processing and handling sequential data without the need for recurrence.
\end{itemize}

Among the 54 articles investigated, 39 incorporated CNNs for at least one modality, 24 utilized FCNNs, and 10 employed RNNs. 
TFs, GNNs, and AEs were selected in 4, 3, and 2 articles, respectively. 
Furthermore, 5 articles opted to employ raw data without employing any unimodal network for at least one modality in their methodologies. 
These findings match with the fact that a large fraction of works use images and clinical tabular data, as shown in Section~\ref{sec:type_modalities} 

\subsubsection{Modality Distributions}
\label{sec:unimodal_modalities}

\begin{figure}[t]
\centering
\includegraphics[width=\textwidth]{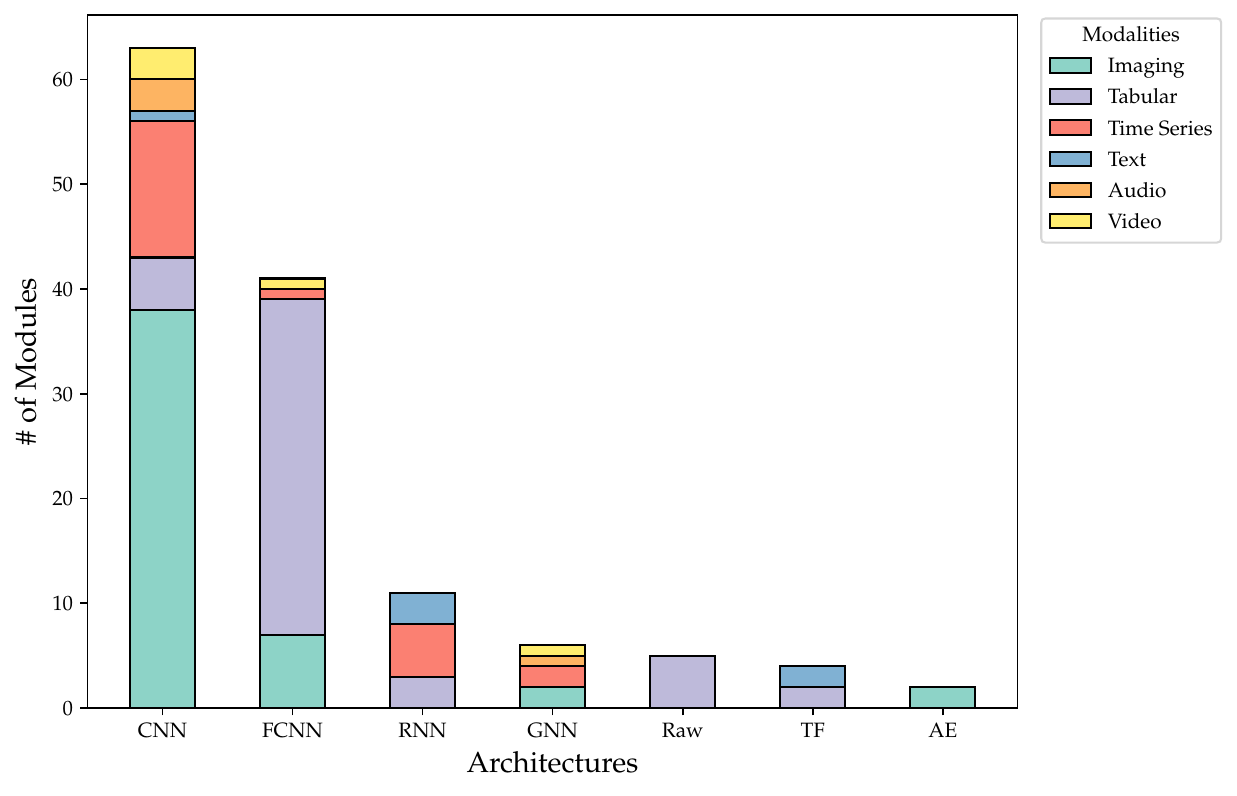}
\caption{Distribution of modalities according to type of architectures.}
\label{fig:unimodal_arch_modal}
\end{figure}

We now turn our attention to the specific modalities employed by these unimodal architectures. \figurename~\ref{fig:unimodal_arch_modal} illustrates the distribution of modalities across each architecture type. 
These modalities represent the original data type which, in some instances, undergoes feature extraction through either handcrafted methods or deep learning algorithms as described in Section~\ref{Features_Extraction}.
Regarding the utilization of architecture types across different modalities, the distribution is as follows: CNNs are utilized in 63 modules, with the imaging modality accounting for 38 of these. 
Time series data follows with 13 modules, tabular data is used in 5 modules, and audio, video, and text data are employed in 3, 3, and 1 modules respectively. 
Among these, 8 modules use handcrafted features for time series data, while 1 module each for text and video modalities employs learned features, with the remaining modules processing raw data.
FCNNs are utilized across 41 modules, with 32 dedicated to tabular and 7 to the imaging modality. 
Time series and video modalities each account for one module. Given the inefficiency of processing raw images with FCNNs, a feature extraction step is typically employed, with handcrafted methods used in 5 instances and learned features in 2. 
For time series data, handcrafted methods are employed, whereas video modality features are extracted using deep learning methods.
RNNs serve 11 modules, 5 of which process time series data, and the remaining 6 are equally divided between tabular and text modalities. 
Ten of these modules use raw data, with only one, pertaining to text data, employing learned features.
GNNs are favored for 6 modules, including 2 each for imaging and time series data, and 1 each for audio and video modalities. 
While imaging modalities are processed in their raw form, the others use extracted features, with one time series modality using handcrafted methods and the other using learned methods. 
Audio features are extracted via deep learning, whereas video features employ handcrafted methods.
TFs are utilized in 4 modules: 2 for tabular data and 2 for text data, with learned features employed for one of the text modalities and the rest using raw data.
AEs are employed in 2 modules, both for the imaging modality; one uses raw data and the other handcrafted features.
Lastly, in all 5 instances where raw data is processed without any unimodal architecture, the data belongs to the tabular modality.
In summary, the exploration of unimodal architectures reveals a broad utilization across various data modalities, with a clear preference for certain architectures in specific contexts. 
CNNs dominate in versatility, especially in imaging, while FCNNs are preferred for tabular data, and RNNs excel in time series and textual data processing. 
GNNs and TFs show specificity in handling complex and sequential data structures, respectively. 
The deployment of these architectures is significantly influenced by the nature of the data, raw or processed through handcrafted or learned features, underscoring the importance of feature extraction and the interaction between data types and architectural choices in designing effective deep learning systems.

\subsubsection{Architectural Combinations}
\label{sec:unimodal_network}

\begin{figure}[t]
\centering
\includegraphics[width=\textwidth]{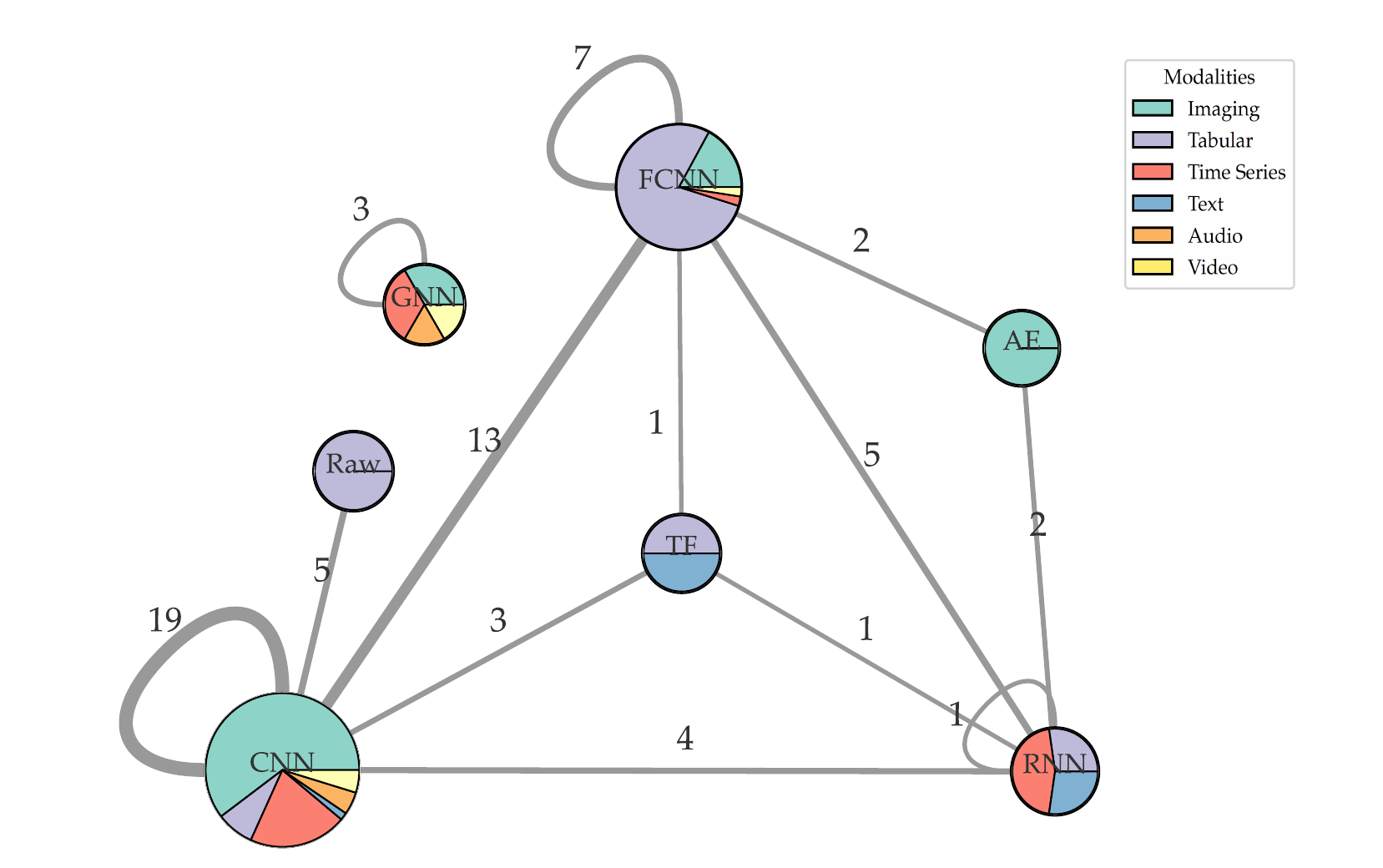}
\caption{Network graph illustrating the interconnectivity between different types of architectures. The numbers on the edges represent the frequency of articles in
which two architectures, indicated as nodes, are employed together. The width of the
edges is drawn proportional to the corresponding number of articles and the size of the nodes is arranged according
to the number of modules that the architecture is used for. The pie charts inside the nodes indicate the proportions of the different modalities that are used with the corresponding architecture.}
\label{fig:unimodal_network}
\end{figure}

Another analysis that we have conducted involved determining the types of architectures used in conjunction as unimodal modules. 
\figurename~\ref{fig:unimodal_network} presents a graph illustrating the combinations of networks utilized. 
The numbers on the edges represent the frequency of articles in which two architectures, indicated as nodes, are employed together. 
The width of the edges is proportional to the corresponding number of articles, and the sizes of the nodes are also arranged according to the number of unimodal modules that the specific architecture is used for. 
Finally, the pie charts inside the nodes represent the different types of modalities.
As observed in the graph, the most prevalent connections are CNNs with themselves, followed by CNNs with FCNNs; 19 articles utilized at least two CNNs together as unimodal modules, while 13 articles employed one or more CNNs with one or more FCNNs. 
The next most frequent connection is within FCNN itself, employed by 7 articles. 
5 articles preferred to use an RNN accompanied by an FCNN, and 4 by a CNN. 
In all the cases where raw data is used for at least one modality, CNNs are chosen for the other modalities.
There are two occurrences of AE and FCNN used together, and two with AE and RNN. 
In fact, these two instances correspond to the same two articles~\cite{bib:14_rahaman2023deep, bib:53_rahaman2022two} that employ three architectures together: FCNN, AE, and RNN. 
When examining the use of TFs, three articles chose to use them with CNN, while among the other two, one preferred to use them with an FCNN and the other with an RNN. 
One interesting observation is that GNNs are exclusively used in conjunction with each other and are not employed with any other type of architecture. 
This trend can be attributed to the specialized nature of graph-structured data and the need to maintain consistent graph representations throughout the model, preserving crucial structural information that other architectures might struggle to handle effectively.
In conclusion, our analysis of the types of architectures used in conjunction as unimodal modules provides valuable insights into the trends and preferences within the field: CNNs emerge as the most commonly utilized architecture, often paired with themselves or with FCNNs, again owing to the prevalence of medical imaging and clinical tabular data which are mostly employed together as seen in \figurename~\ref{fig:bimodal_modalities}. 
The exclusive pairing of GNNs with each other highlights their distinct role within the landscape of neural network architectures.

\subsubsection{Dimensionality Transformation}
\label{sec:condensation}

\begin{figure}[t]
\centering
\includegraphics[width=\textwidth]{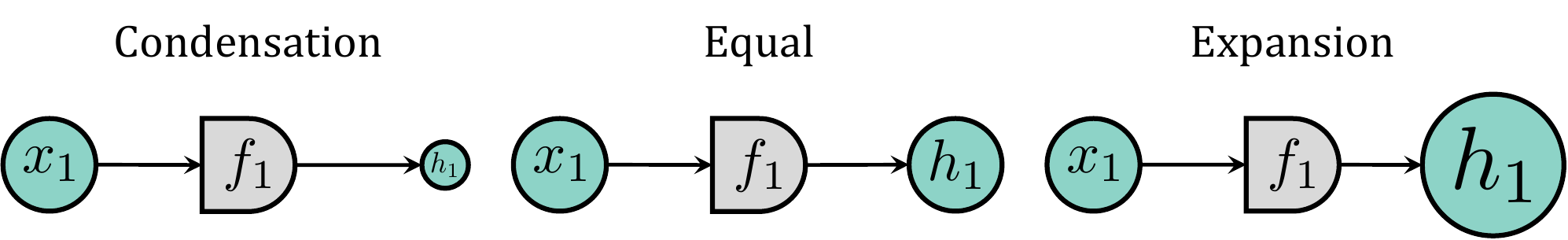}
\caption{Schematic representation of condensation/expansion: $x_1$ represents the input and $h_1$ represents the marginal representation. The size of the circles represents the size of the vectors. The information is considered condensed if the size of the marginal representation is smaller than the input size, and expanded if it is larger. The dimension of the information remains the same if the size of the marginal representation is equal to the input size.}
\label{fig:condensation_schema}
\end{figure}

Unimodal modules are responsible for extracting features, often referred to as marginal representations, from individual modalities which are subsequently utilized by the fusion block.
A primary advantage of this approach is the ability to represent high-dimensional inputs through more compact unimodal features, thereby condensing the information into a vector of reduced dimensionality. Conversely, it can also expand the dimension of the information by creating new information from the interaction of low-dimensional input features.
The information is considered condensed if the marginal representation size is smaller than the input size, expanded if the marginal representation size is greater, and equal if the sizes are the same.
\figurename~\ref{fig:condensation_schema} depicts a schematic representation of this concept.

\begin{figure}[t]
\centering
\includegraphics[width=\textwidth]{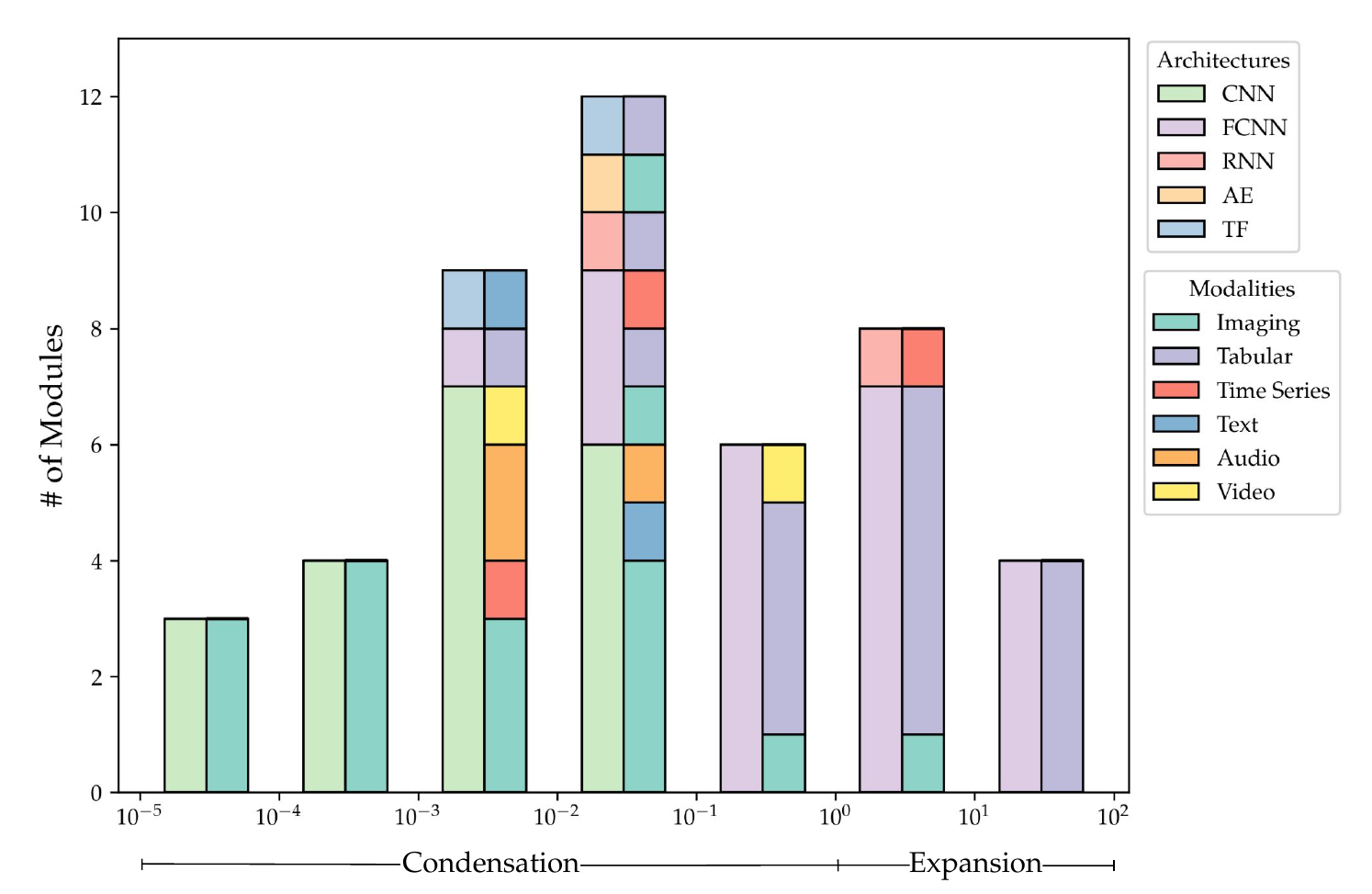}
\caption{Histogram of the condensation/expansion ratios divided according to architecture (bars on the left) and modality (bars on the right).}
\label{fig:condensation}
\end{figure}

\begin{figure}[t]
\centering
\includegraphics[width=\textwidth]{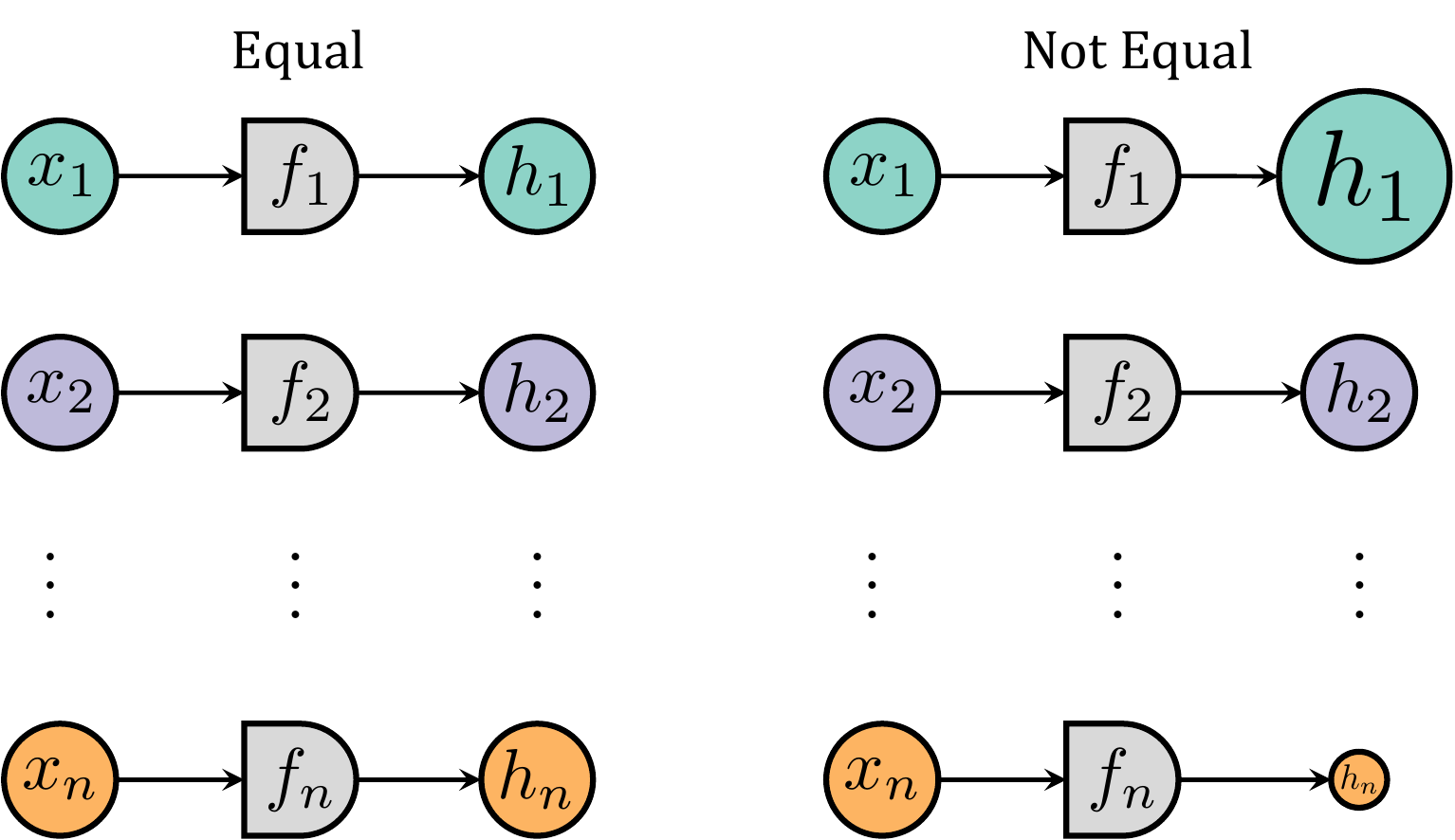}
\caption{Distribution of the marginal representations: $h_1, h_2, ..., h_n$ represent the marginal representations and the size of the circles represents the size of the feature vectors. Marginal representations are distributed equally if all of them have the same size and not equally if at least two of them have different sizes.}
\label{fig:unimodal_distribution}
\end{figure}

\figurename~\ref{fig:condensation} illustrates the degree of condensation/expansion, with the data categorized based on the architecture type and modality involved. It is calculated by dividing the marginal representation size by the input size. 
Among the articles surveyed for this review, only 39\% provide information about the input size and marginal representation size for at least one modality. 
\figurename~\ref{fig:condensation} depicts the total number of modules where the corresponding ratio is applied in these articles. 
The majority of modules condense information by a factor of one to two. 
CNNs notably condense more information than other architectures, particularly in imaging modalities, which inherently entail larger sizes. 
Conversely, smaller condensation ratios, or even expansions, are predominantly observed in modules utilizing FCNNs with tabular modalities. 
This observation aligns with the nature of tabular data, primarily composed of clinical and demographic information with substantially smaller sizes compared to other modalities. 
There are only two papers~\cite{bib:24_ostertag2023long, bib:43_ostertag2020predicting}, both implemented by the same author, that prefer a marginal representation dimension equal to the input dimension, hence a ratio of 1, both utilizing an FCNN for tabular data.
Several modules exhibit extreme expansion ratios, a less common occurrence in the literature. 
One contributing factor to this phenomenon is the endeavor to maintain equal marginal representation sizes for each modality. 
For example, in the case of~\cite{bib:16_holste2021end}, where one modality is imaging and the other is tabular data, the imaging input size is 50,000 while the input size of the tabular data is only 18. 
The authors opt for an equal marginal representation size of 512. 
Consequently, while the condensation ratio for the imaging modality is 0.01, the expansion ratio for the tabular modality is 28.44. 
Similarly, in ~\cite{bib:46_kohankhaki2022radiopaths}, the tabular modality has a considerably smaller size (15) compared to the other two modalities, imaging (50,176) and text (393,216), with authors again selecting an equal marginal representation size of 512. 
Consequently, the tabular modality exhibits an expansion ratio of 34 alongside condensation ratios of 0.01 for imaging and 0.001 for text. 
\figurename~\ref{fig:unimodal_distribution} represents a schematic of the distribution of the marginal representations. Marginal representations are distributed equally if all of them have the same size and not equally if at least two have different sizes.

In fact, many studies prefer equally distributed marginal representations.
Among the 18 articles providing marginal representation sizes for all modalities, only five~\cite{bib:21_guida2023improving, bib:24_ostertag2023long, bib:25_hayat2022medfuse, bib:32_bhattacharya2022multi, bib:43_ostertag2020predicting} feature non-equally distributed marginal representations. 
Cahan et al.~\cite{bib:37_cahan2023multimodal}, who prefer to have equal marginal representation sizes for each modality, state that the rationale behind this approach is to prevent any unintended bias towards one modality over others. 
While this rationale is sensible, it is uncertain whether all the different modalities can be represented with an equal number of features. 
All in all, the degree of information condensation/expansion depends on the input modality and the architecture that processes the corresponding input. 
Our analysis indicates that the imaging modality, when processed with CNNs, exhibits the highest condensation ratios, while tabular data processed with FCNNs demonstrates the highest expansion ratios.

Unimodal representations are extracted using separate unimodal modules. 
These representations are then forwarded to the next step, which we refer to as the fusion module, where they are fused to create a common multimodal representation. 
The following section presents the newly proposed notation and taxonomy alongside the analysis of the fusion modules adopted in the literature.

\subsection{Fusion Module}\label{sec:fusion_module}

Intermediate multimodal fusion involves integrating data from multiple modalities (e.g., text, image, audio, video) at an intermediate data representation of the modalities rather than at the data or the output level, as in early and late fusion approaches respectively. 
This approach enables an interaction between the modalities that are inherently unattainable in early fusion methods, where the early integration of modalities hinders the disentanglement of contributions for each data type. 
Likewise, it overcomes the limitations of late fusion techniques, which often rely on independently learned disjoint spaces for each modality.

At the core of the intermediate fusion approach there is the fusion module, denoted as \Fusion{} in \figurename~\ref{fig:joint_fusion}: it fuses the hidden representations extracted from the $n$ unimodal modules $h_1, \ldots, h_n$ into the unified multimodal representation $h$. 
In other words, the fusion module aims to map the features of different modalities into a common embedding space, which can be obtained through different strategies exploiting the hidden representation spaces of the modalities. 
These approaches, characterized by different degrees of abstraction and characterization of the feature spaces, intend to maximize the complementarity of the modalities during the learning phase.
Indeed, there is no unified approach to maximize the informativeness of the extracted representations, but each approach uses more or less complex fusion structures to this end.
These may involve different fusion operations, achieved by a single fusion or multiple fusions, which may combine all available information, eventually also using $h_1, \ldots, h_n$ multiple times. 

Recognizing the central role of the fusion module we hereby introduce a formal notation specifically designed to highlight the fusion process detailed in the next Section~\ref{sec:notation}. 
This notation framework aims to facilitate a thorough understanding of the critical aspects of multimodal fusion, namely \textit{what}, \textit{how many}, \textit{when} and \textit{how} to fuse: these three issues will be further analyzed in Sections~\ref{sec:what}, ~\ref{sec:how_many_fusion}, ~\ref{sec:when_fusion} and ~\ref{sec:how_fusion}.
Hence, we will categorize the existing intermediate multimodal fusion approaches and offer a framework to distinguish the various architectural design choices available when designing an intermediate fusion multimodal model.
It is worth noting that our notation framework applies beyond the medical domain, i.e., the focus of this survey.
Moreover, Section~\ref{sec:inter_uni_fus} explores the correlation between the unimodal modules and the fusion module architecture adopted.

\subsubsection{Systematic Notation for Multimodal Fusion}\label{sec:notation}

We now introduce a comprehensive mathematical notation system to facilitate a rigorous analysis of the multimodal intermediate fusion strategies employed in the revised studies. 
It enables the detailed description and precise representation of the nature, localization and timing of the fusion mechanisms designed in the architectures. 
A fusion is defined as the operation that integrates input data, as modalities and/or previous fusions, producing a unified representation. 

We denote a generic single fusion as:
\begin{equation}
    \Fusion{i} = \bullet (\Nlayers[l]{\alpha_j}, \Nlayers[m]{\alpha_k}, \ldots)\out
\end{equation}
where the components of the notation are:

\begin{itemize}
    \item \textbf{Fusion function} (\Fusion{}): identifies the fusion function that takes as input multiple feature representations, e.g., modalities or outputs of other fusions, and returns a single fused representation.
    \item \textbf{Fusion ordering} ($i$): Denotes the position of the fusion within the architecture, following a sequential progression from input to output. 
    Therefore, the highest fusion ordering index denotes the number of fusions performed in the fusion module.
    \item \textbf{Inputs} ($\alpha_j$, $\alpha_k$, etc.): Represent the input modalities that can be the sample $x$ represented in the original feature space, or the output of previous fusions \Fusion{}. 
    For example, two different inputs $j$ and $k$ could be represented as $x_j$ and $x_k$, $\Fusion{j}$ and $\Fusion{k}$, or a combination such as $x_j$ and $\Fusion{k}$.
    \item \textbf{Layer depth} ($l$ and $m$): Indicate the number of trainable layers through which inputs $\alpha_j$ and $\alpha_k$ have been processed before their integration. 
    This indicates when the fusion of information occurs. 
    An example could be a fusion function that takes as input $\Nlayers[2]{\alpha_j}$, $\Nlayers[3]{\alpha_k}$, implying that the two input representations passed through two and three layers, respectively, before the fusion.
    \item \textbf{Fusion operation} ($\bullet$): Symbolizes the type of operation performed between inputs $\alpha_j$, $\alpha_k$, etc. 
    To standardize the analysis, we identified 5 fusion categories: \textit{Concatenation}, \textit{Attention}, \textit{Tensor-operation}, \textit{Calibration}, and \textit{Knowledge-sharing}, each represented by specific symbols listed in \tablename~\ref{tab:fusion_symbols} and further detailed in Section~\ref{sec:how_fusion}. 
    The analysis of these operations provides insight into how each fusion is performed.
    \item \textbf{Final Fusion} ($\out$): if present, it specifies that the considered fusion function is the last in the architecture and its output coincides with the latent fused representation $h$, given as input to the multimodal module $f$. Conversely, if not present, it denotes that the fusion under consideration is at the beginning or in the middle of the fusion structure. 
\end{itemize}

\begin{table}[t]
    \centering
    \begin{tabular}{c|c}
    \toprule
        \textbf{Fusion Category} & \textbf{Symbol} \\
    \midrule 
        Concatenation & \concat \\
        Attention & \attention \\
        Tensor-operation & \tensorOP \\
        Calibration & \calibr \\
        Knowledge-sharing & \kShare \\
        \bottomrule
    \end{tabular}
    \caption{Symbols representing the fusion categories.}
    \label{tab:fusion_symbols}
\end{table}

After identifying all the components used in the model, it is possible to draw a directional bipartite graph $G(N, E)$ representing how the modalities and fusions are combined in each fusion approach, where $N$ is composed of 2 types of nodes, modalities and fusions, respectively, whilst $E$ is composed of directed edges which connect the inputs to a fusion. 

\begin{figure}[!ht]
    \centering
    \includegraphics[height=3cm]{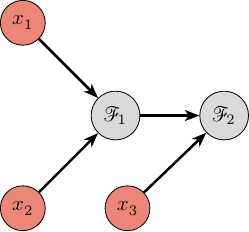}
    \caption{Graph representation of the fusion strategy applied in~\cite{bib:21_guida2023improving}, used as example to present our formal notation. Here, the modalities and the fusion nodes are represented in red and gray, respectively.}
    \label{fig:fusion_strategy_graph}
\end{figure}

To better illustrate the use of the proposed notation in the analysis of a multimodal fusion approach, we examine the intermediate fusion strategy from~\cite{bib:21_guida2023improving}, which integrates 3 modalities, specifically MRI, X-Ray and tabular data, represented with $x_1$, $x_2$, and $x_3$, respectively, through the following fusion operations:

\begin{equation}
\begin{aligned}
\Fusion{1} &= \concat(\Nlayers[21]{x_1}, \Nlayers[25]{x_2}) \\
\Fusion{2} &= \concat(\Nlayers[0]{\Fusion{1}}, \Nlayers[0]{x_3})\out
\end{aligned}
\end{equation}
This structure features 2 fusion operations: \Fusion{1} and \Fusion{2}. 
The first fusion \Fusion{1} concatenates the modalities $x_1$ and $x_2$ after processing them through 21 and 25 convolutional layers, respectively. 
The second fusion $\Fusion{2}$ combines the output of \Fusion{1}, directly without any additional trainable layer, with the $x_3$ modality.
The latter modality, since it does not go through any unimodal module, is used in its raw form, \Nlayers[0]{x_3}. 
Fusion \Fusion{2} yields the final multimodal representation of the module, as indicated by the $\to$ symbol, which is then fed to the multimodal module.
This complete set of operations inside the fusion module is also represented by its graph in \figurename~\ref{fig:fusion_strategy_graph}.

We adopt this notation and graph representation for the comprehensive analysis of all 54 articles reviewed, which is reported in Supplementary Material B.
This document examines each method offering four key pieces of information: first, we report the model image, taken from its respective article; second we formalize the fusion strategy with our approach thus reporting a set of equations; third, we offer the graph detailing the fusion structure; fourth, it also reports further taxonomies we have defined, which will be detailed in the next section. 

\subsubsection{Fusion Module Taxonomy}\label{sec:fus_tax}

The systematical classification of the reviewed fusion methodologies by using the notation introduced so far not only offers to the readers a systematic analysis and categorization of the literature, but it also permits us to highlight \textit{what} is fused, \textit{how many} times the fusion is performed, \textit{when} the fusion occurs, and \textit{how} the fusion is performed.

These categorizations are intended to group the fusion methods according to their fundamental characteristics and principles, thus seeking an answer to some fundamental questions about intermediate fusion strategies for multimodal inputs: 
\begin{itemize}
    \item What elements are involved in the fusion?
    \item How many times does the fusion occur between the modalities? 
    \item When is the fusion performed between the modalities?
    \item How is the fusion among the modalities performed? 
\end{itemize}

We aim to provide a coherent and generalized understanding of the main fusion approaches applied in the field, facilitating a deeper comprehension of their implications and applications in multimodal research and proposing suitable and coherent taxonomies for the classification and categorization of any work in this field. 
However, the proposed taxonomies may not cover all the possible multimodal intermediate fusion methodologies, but it is applicable to the works considered in this review. 
Nevertheless, the introduction of our notation system aims at simplifying the analysis of any multimodal fusion method, thus making more straightforward the extension of the taxonomies presented in this review. 

In the subsequent sections, we present our proposed taxonomies for multimodal fusion strategies.
To illustrate these taxonomies, we draw upon select articles from the literature as examples. 
However, it is important to note that our discussion of these examples is intended to be illustrative rather than exhaustive. 
We aim to provide sufficient context to elucidate our classification system, while an exhaustive in-depth analysis of each specific approach can be found in Supplementary Material B.

\begin{figure}[t]
\centering
\resizebox{\textwidth}{!}{
    \begin{tikzpicture}

        \node[label=below:(a)] (single) at (0, 0) {\includegraphics[height=3cm]{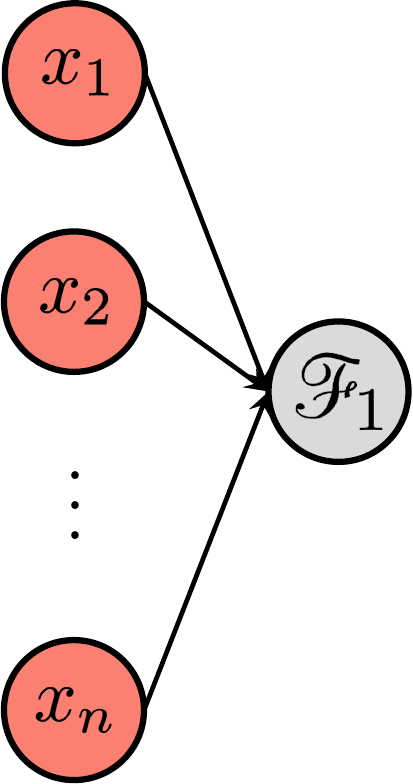}};
        
        \node[label={[yshift=0.1cm]below:(b)}] (sudden) [right=4cm of single] {\includegraphics[height=3.2cm]{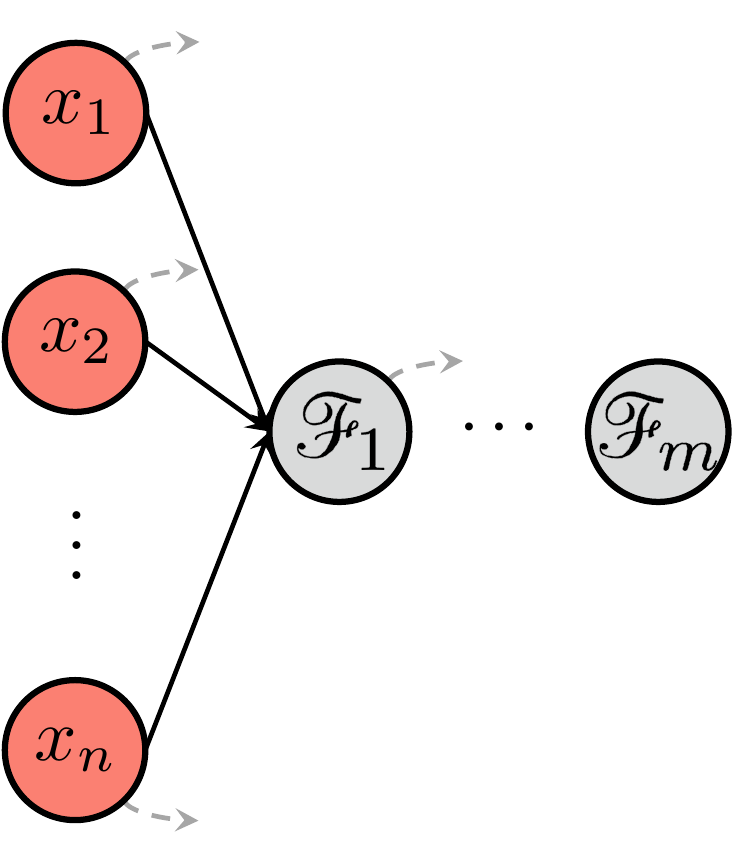}};
             
        \node[label=below:(c)] (gradual) [below=1cm of single] {\includegraphics[height=2.5cm]{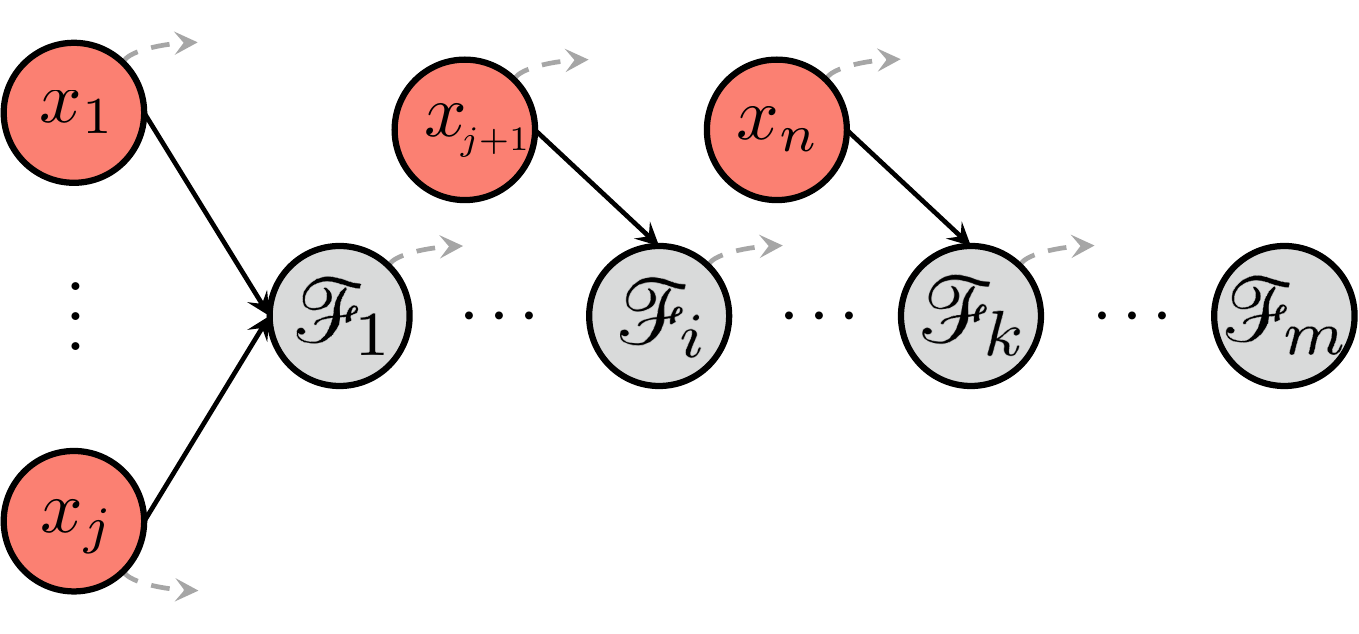}};
        
        \node[label={[yshift=-0.25cm]below:(d)}] (multi_flow) [below=1cm of sudden] {\includegraphics[height=2.5cm]{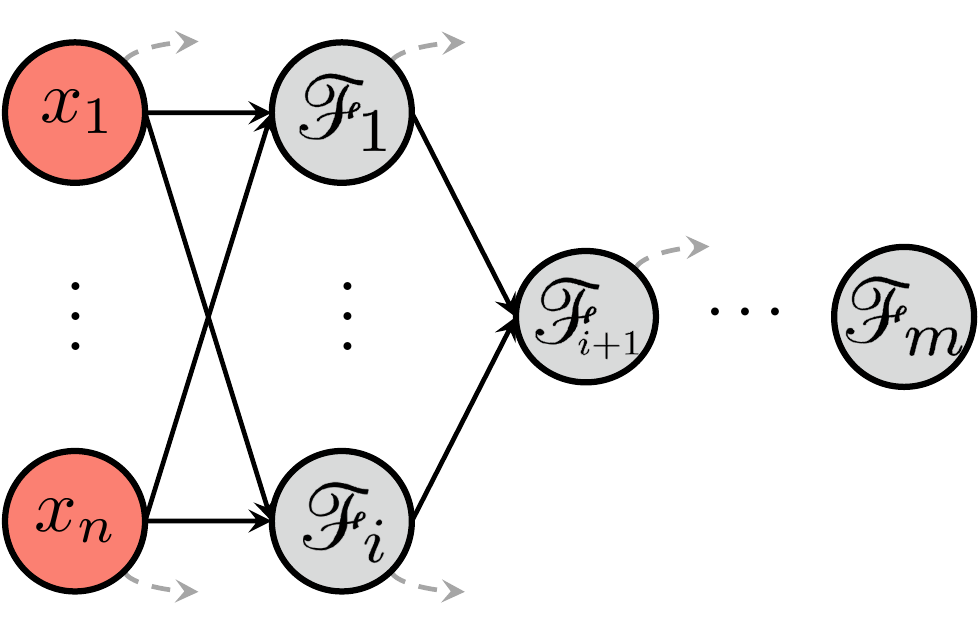}};
        
    \end{tikzpicture}}
    \caption{Structures types: (a) Single Sudden; (b) Multiple Sudden; (c) Multiple Gradual; (d) Multiple Multiflow. First row: all Sudden architectures; Second row: gradual architecture. The dashed gray arrows for the Multiple Sudden, Multiflow and Gradual indicate that each node might be used as additional inputs for subsequent fusion nodes.}
    \label{fig:taxonomies}
\end{figure}

\subsubsection{What?}\label{sec:what}

Regarding what is being fused, in section~\ref{sec:type_modalities}, we identified the main categories of modalities, including imaging, tabular, text, time-series, video, and audio. 
Furthermore, our proposed notation allows for clear visualization of the fusions performed between different modalities and other multimodal representations before getting to the final output of the proposed model. 
For a more comprehensive understanding, readers can refer to Supplementary Material B, where we have specified the modalities used by each fusion function in the respective methods.

\subsubsection{How Many?}\label{sec:how_many_fusion}

Going forward, we quantitatively assess the fusion strategies adopted within the corpus of the reviewed literature, by categorizing the architectures based on whether they opt for a \textit{single} fusion mechanism or \textit{multiple} fusions throughout the model's structure.
The predominance of \textit{single} studies (35 out of 54) demonstrates a preference for this simpler approach over those employing \textit{multiple} fusions (19 out of 54). 
This observation underscores a preference towards intermediate fusion strategies that favor simpler and straightforward modalities integration, relying on robust unimodal representations. 
This trend is related to the vast typologies of data adopted in multimodal fusion in the biomedical field, where data curation and processing play a central role, and the fusion techniques are not fully investigated.
Furthermore, these single fusion approaches often investigate more sophisticated multimodal modules able to analyze complex multimodal representations; for an in-depth discussion on the characteristics and implications of these multimodal modules, refer to Section~\ref{sec:multimodal_module}.

\subsubsection{When?}\label{sec:when_fusion}

Upon evaluating when the fusion occurs among different modalities within multimodal architectures, two primary categorizations emerged from our analysis. 

The first categorization focuses on the timing of the fusions that integrate the different modalities: whether all modalities are fused simultaneously in a single operation, gradually across multiple ones, or if there are multiple components of the network elaborating the same inputs. 
Firstly, we distinguish \textit{sudden} architectures, characterized by the simultaneous fusion of all modalities in a single fusion, from \textit{gradual} architectures, where modalities are fused progressively and incorporated in subsequent and distinct fusions. 
Secondly, we identify the \textit{multi-flow} categorization as a specific group of methods that employ multiple sequences of fusion operations that process information from a set of considered modalities, defined as \textit{fusion flow}.
Therefore, the \textit{multi-flow} structure is characterized by the generation of multiple distinct multimodal representations, each reflecting a unique integration and elaboration of all the input modalities, which are then fused into a unique fused representation. 
The graphical representation of single sudden, multiple sudden, multiple gradual and multi-flow strategies are shown in \figurename~\ref{fig:taxonomies} in panel a, b, c and d, respectively. 

To elucidate the application of the above categorization, we present an example drawn from the list of reviewed articles for each category of the taxonomy. 

The set of fusions in the architecture proposed in~\cite{bib:03_ahmad2023aatsn} can be represented as follows:

\begin{minipage}{0.5\linewidth}
\begin{equation}
\begin{aligned}
\Fusion{1} &= \attention(\Nlayers[4]{x_1}, \Nlayers[5]{x_2}) \\
\Fusion{2} &= \attention(\Nlayers[3]{x_1}, \Nlayers[1]{\Fusion{1}}) \\
\Fusion{3} &= \attention(\Nlayers[2]{x_1}, \Nlayers[1]{\Fusion{2}})\out
\end{aligned}
\end{equation}
\end{minipage}
\begin{minipage}{0.5\linewidth}
\centering
\includegraphics[height=2cm]{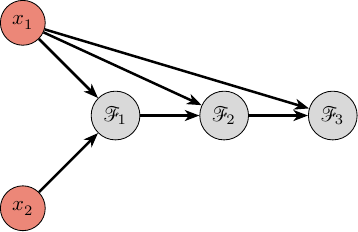}
\vspace*{.3cm}
\end{minipage}
Here, \Fusion{1} qualifies the approach as sudden due to the integration of both modalities \Nlayers{x_1} and \Nlayers{x_2} in a single fusion. 
Subsequent fusions \Fusion{2} and \Fusion{3}, which integrate modality \Nlayers{x_1} with previous fusion outputs, do not alter this classification, as the initial fusion already incorporated all the considered modalities, even if \Fusion{2} and \Fusion{3} are adopted to fuse different unimodal representations ($\Nlayers[3]{x_1}$ and $\Nlayers[2]{x_1}$) and other fusion functions ($\Nlayers[1]{\Fusion{1}}$ and $\Nlayers[1]{\Fusion{2}}$).

Conversely, the architecture presented in~\cite{bib:30_meng2022msmfn} exemplifies the gradual approach and can be described with the following fusion functions:

\hspace{-.5cm}
\begin{minipage}{.7\linewidth}
\begin{equation}
\begin{aligned}
\Fusion{1} &= \tensorOP(\Nlayers[1]{x_1}, \Nlayers[3]{x_2}) &\quad
\Fusion{2} &= \tensorOP(\Nlayers[3]{x_2}, \Nlayers[1]{x_3}) \\
\Fusion{3} &= \concat(\Nlayers[3]{x_2}, \Nlayers[2]{\Fusion{1}}, \Nlayers[2]{\Fusion{2}})  &\quad
\Fusion{4} &= \kShare(\Nlayers[x]{x_4}, \Nlayers[2]{\Fusion{3}}) \\
\Fusion{5} &= \concat(\Nlayers[0]{x_5}, \Nlayers[7]{\Fusion{4}})\out 
\end{aligned}
\end{equation}
\end{minipage}
\begin{minipage}{.3\linewidth}
    \centering
    \includegraphics[height=2cm]{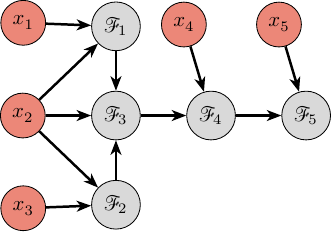}
    \vspace{.2cm}
\end{minipage}

\Fusion{1} fuses the modalities \Nlayers{x_1}, \Nlayers{x_2} and \Nlayers{x_3}, while the fusions \Fusion{4} and \Fusion{5} gradually integrate \Nlayers{x_4} and \Nlayers{x_5}, respectively, thus depicting the gradual approach for the integration of the modalities. 
Through this fusion approach the integration between the modalities is performed at different abstraction levels of the architecture, resulting in a hierarchical processing ability of the model.

As an example of multi-flow method, in~\cite{bib:27_mohammed2023mmhfnet} the authors proposed a multi-flow architecture, whose fusions can be represented as follows:

\vspace{-.1cm}
\begin{minipage}{.7\linewidth}
\begin{equation}
\begin{aligned}
\Fusion{1} &= \concat(\Nlayers[1]{x_1}, \Nlayers[1]{x_2}) & \Fusion{2} &= \concat (\Nlayers[2]{x_1}, \Nlayers[2]{x_2}) \\
\Fusion{3} &= \concat(\Nlayers[3]{x_1}, \Nlayers[3]{x_2}) & \Fusion{4} &= \concat(\Nlayers[4]{x_1}, \Nlayers[4]{x_2}) \\
\Fusion{5} &= \concat(\Nlayers[3]{\Fusion{1}}, \Nlayers[3]{\Fusion{2}}, \Nlayers[3]{\Fusion{3}}, \Nlayers[3]{\Fusion{4}})\out
\end{aligned}
\end{equation}
\end{minipage}
\begin{minipage}{.3\linewidth}
    \centering
    \includegraphics[height=2cm]{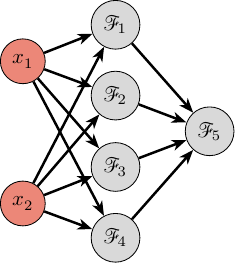}
\end{minipage}\vspace{.2cm}
This model, integrating 2 modalities, $x_1$ and $x_2$, through 4 parallel and independent fusions \Fusion{1}, \Fusion{2}, \Fusion{3}, and \Fusion{4}, exemplifies a multi-flow methodology.
Indeed, it presents 4 distinct fusion flows, each elaborating all the modalities on different abstraction levels of the unimodal representations modules. 
Finally, \Fusion{5} integrates each different fusion flow in a single representation, denoting a multi-level fusion features space. 

The second categorization delves into the abstraction level of the modalities at which the fusion occurs, as denoted by the superscript index indicating the number of trainable layers each modality, or fusion,  has passed through before being fed to the configured fusion module.
We differentiate between \textit{synchronous} and \textit{asynchronous} fusion methods. 
The former is characterized by the processing of each modality through an identical number of trainable layers prior to fusion, as indicated by matching superscripts. 
Conversely, the latter integrates modalities that have passed through different numbers of trainable layers, thus fusing at varying levels of abstraction. 
It is worth noting that the gradual fusion approach intrinsically aligns with asynchronous methodologies: the sequential integration of modalities at distinct stages inevitably leads to varying degrees of abstraction among the modalities being fused, thereby precluding a synchronous fusion process for gradual architectures.

Examples of synchronous and asynchronous methods are found in~\cite{bib:10_njoku2022deep} (eq.~\ref{eq:sync}) and~\cite{bib:11_oh2023deep} (eq.~\ref{eq:async}), respectively. 
Despite their use of the same modalities and fusion function, the distinction lies in the superscript numbers for each modality before fusion:

\begin{minipage}{.5\linewidth}
\begin{align}\label{eq:sync}
    \Fusion{1} &= \concat(\Nlayers[1]{x_1}, \Nlayers[1]{x_2})\out \\ 
    \label{eq:async}
    \Fusion{1} &= \concat(\Nlayers[3]{x_1}, \Nlayers[2]{x_2})\out 
\end{align}
\end{minipage}
\begin{minipage}{.5\linewidth}
\centering
\includegraphics[height=2cm]{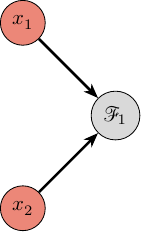}
\end{minipage}\vspace{.3cm}
The synchronous method~\cite{bib:10_njoku2022deep}, integrates modalities $x_1$ and $x_2$ after they have both been processed through a singular trainable layer, ensuring uniform abstraction levels at the fusion point. 
Conversely, the asynchronous method~\cite{bib:11_oh2023deep}, processes the modalities $x_1$ and $x_2$ through different unimodal modules, composed of 3 layers for $x_1$ and 2 for $x_2$ before their fusion.

\begin{table}[t]
    \centering
    \footnotesize
    \begin{tabularx}{\linewidth}{C{1.5cm}|C{1.1cm}|C{1.3cm}|Z}
    \toprule[1pt]
        \textbf{When?}    & \textbf{How Many?} &  \textbf{\# of Articles} & \textbf{References}  \\
    \midrule[1pt]
    \multirow{4}{*}{Sudden}     &  \multirow{3}{*}{Single} & \multirow{3}{*}{35} & 
   \cite{bib:05_li2022aviper,bib:06_lobantsev2020comparative,bib:07_menegotto2021computer,bib:08_menegotto2020computer,bib:11_oh2023deep,bib:14_rahaman2023deep,bib:15_zheng2022dyhealth,bib:16_holste2021end,bib:22_zhang2023itcep,bib:25_hayat2022medfuse,bib:26_zeng2022miftp,bib:28_wang2021modeling,bib:31_rahaman2021multi,bib:32_bhattacharya2022multi,bib:33_guarrasi2023multi,bib:34_perez2022multi,bib:35_steyaert2023multimodal,bib:37_cahan2023multimodal,bib:44_ma2023predicting,bib:46_kohankhaki2022radiopaths,bib:47_sousa2023single,bib:51_wu2023transformer,bib:53_rahaman2022two,bib:54_cahan2022weakly},
   \Sync{\cite{bib:10_njoku2022deep,bib:52_alsherbiny2021trustworthy,bib:45_seo2021predicting,bib:38_subramanian2020multimodal,bib:39_huang2020multimodal,bib:42_shetty2023multimodal,bib:36_han2022multimodal,bib:20_sedghi2020improving,bib:12_niu2023deep,bib:13_radhika2021deep,bib:49_rashid2023tinym2net}}
   \\
    \cmidrule(r){2-4} 
    & 
    
    \multirow{1}{*}{Multiple} & \multirow{1}{*}{11} & 
    \cite{bib:09_hou2023deep,bib:03_ahmad2023aatsn,bib:02_li2022dynamic},
\Sync{\cite{bib:17_cen2022exploring,bib:18_jiao2023gmrlnet,bib:19_he2021hierarchical,bib:23_pan2021liver,bib:29_chen2022ms,bib:40_kuttala2023multimodal,bib:41_li2022multimodal,bib:48_radhika2021stress}}
    \\\midrule
    \multirow{1}{*}{Gradual} & \multirow{1}{*}{Multiple} & \multirow{1}{*}{5} & \cite{bib:04_liu2022attention,bib:21_guida2023improving,bib:24_ostertag2023long,bib:30_meng2022msmfn,bib:43_ostertag2020predicting}
    \\\midrule
    \multirow{1}{*}{Multi-flow} &  \multirow{1}{*}{Multiple} & \multirow{1}{*}{3} & 
    \cite{bib:01_li2022bi,bib:27_mohammed2023mmhfnet,bib:50_sun2023toward} \\\bottomrule
    \end{tabularx}
    \caption{Classification of articles by the number of fusion (single/multiple), and fusion timing categorization (sudden/gradual/multi-flow). The articles with the symbol $\ddagger$, incorporate a synchronous fusion strategy, whereas those without the symbol use an asynchronous approach.}
    \label{tab:taxonomies}
\end{table}

To summarize the considerations reported so far, \tablename~\ref{tab:taxonomies} reports the list of the reviewed articles categorized in sudden, gradual and multi-flow, and distinguished by the adoption of single and multiple fusions approaches. 
On the one hand, the single fusion approaches are all categorized in sudden strategy, given the adoption of single fusion operation; on the other hand, we observe that most of the multiple methods apply the sudden strategy (11 articles) over the more complex gradual (5 articles) and multi-flow (3 articles) approaches. 
This trend highlights the preference for merging the modalities all at once with sudden approaches in the biomedical field, underlying the will to focus intermediate fusion learning on fine-grained and exhaustive unimodal features, simplifying the subsequent fusion operations.
The prevalence of this approach may be attributed to the absence of a priori knowledge regarding modality hierarchies in the biomedical domain, precluding the direct establishment of a definitive fusion strategy.
Consequently, the proposed methodologies predominantly leverage well-established unimodal architectures capable of extracting salient information, rather than exploring more intricate fusion strategies. 
This conservative approach mitigates the risk of introducing unintended biases during the training process, which could potentially compromise the efficacy of multimodal feature learning.

Within single sudden fusion architectures, asynchronous methods predominate (24 articles out of 35), compared to synchronous instances. 
This trend reflects the widespread adoption of a single fusion to fuse different data types (23 articles out of 35), such as imaging and tabular data, which typically require different preprocessing and feature extraction processes, resulting in heterogeneous unimodal architectures.
In contrast, architectures that implement multiple fusions exhibit a more balanced distribution, with 10 asynchronous and 9 synchronous implementations.

\subsubsection{How?}\label{sec:how_fusion}

We identify five types of approaches that can be used to fuse the data, that we already introduced in \tablename~\ref{tab:fusion_symbols}. 
They are: \textit{Concatenation}, \textit{Attention}, \textit{Tensor-operation}, \textit{Calibration}, and \textit{Knowledge-Sharing}.
In the following, we deepen each one as we remark on the need to carefully consider the method's selection and implementation. 
Indeed, understanding the strengths and limitations of each fusion class is a crucial step for researchers willing to make informed choices that optimize the integration of unimodal data, potentially leading to more sophisticated and effective multimodal systems.

\paragraph{Concatenation}
It is by far the most commonly used approach in the reviewed methods, where multiple input representations or features are combined by joining them along a specific axis or dimension. 
This method creates a single, larger representation that preserves all the information of its original components.
Notably, the majority of single fusion works (29 out of 35) utilize concatenation to merge features processed by unimodal modules.
An example is~\cite{bib:14_rahaman2023deep}, where the authors used concatenation to merge 3 unimodal representations derived from structural and functional neuroimaging and genomic data.
However, this approach introduces several limitations for multimodal methods, particularly concerning the high dimensionality of the fused feature space.
Indeed, in studies that employ multiple fusions, concatenation is generally adopted as the primary technique to elaborate the intermediate fused representations, thereby inheriting the same limitations identified in single fusion methods.
An example is~\cite{bib:27_mohammed2023mmhfnet}, which introduces a multi-flow network where each flow is represented by a concatenation function that merges unimodal features at different levels of abstraction.
However, most of the studies (13 over 19) employ concatenation for a structural necessity, because they use it to produce the final version of the fused features after all the other fusion operations have taken place. 
This further processing of the representations usually results in a reduced and more compact form with respect to those methods that employ the concatenation immediately after the unimodal modules, limiting also possible dimensionality issues and poor correlation between the fused representations.

\paragraph{Tensor-operation}
Tensor-operation represents a versatile category within our categorization, and groups together a wide range of mathematical strategies for synthesizing multimodal representations from interactions between unimodal representations. 
These operations, e.g., Hadamard product, outer product, pooling operations, and aggregation functions, as minimum, maximum and average, enable the generation of representations with varying dimensionalities, potentially lower, equal, or greater than those of the input representations.

In single fusion approaches, only a few studies~\cite{bib:25_hayat2022medfuse, bib:45_seo2021predicting, bib:42_shetty2023multimodal} delve into this fusion category. 
For instance, in~\cite{bib:42_shetty2023multimodal} the authors present an approach that employs an element-wise interaction between tensors of imaging features from CXRs and textual features from radiology reports, that ensures an element-wise interaction between the different modalities involved. 
This technique generates an interactive map integrating modalities that also retains the input's dimensionality. 

In multiple fusion approaches, tensor-operation is mostly used in combination with other fusion operations (13 works out of 19).
For example, in~\cite{bib:29_chen2022ms} the authors introduce a multiple fusion architecture that incorporates both modality-specific networks and modality-shared networks. 
Combining the knowledge-sharing and the average operation between the modality-specific representations, the model yields a fused representation that exploits the relationships between the features across all the patients and, through the average, maintains the information from the 2 inputs in a final space with the same inputs' dimensionality. 
However, the effectiveness of approaches employing tensor-operation fusion strategies may be constrained by the limitations inherent to the lack of optimized and dynamic mechanisms that can improve how the modalities interact with each other. 
Indeed, these fusion operations are not able to dynamically obtain a rich multimodal representation, thus they rely on different parts of the architecture for the elaboration of the multimodal representation.

\paragraph{Attention}
The Attention category contains all the methods that use an inter-modality attention mechanism, either in its classic form or a customized version. 
The characteristic of these approaches lies in the generation of fused features that emerge from the interactions between different modalities, guided by an attention mechanism. 
By prioritizing features based on their significance in inter-modality relationships, the attention mechanism ensures that the fusion process promotes a multimodal synthesis where the contribution of each modality is weighted according to its relevance and complementarity to the others.
This mechanism directs the network's learning focusing on the most pertinent features across the modalities, with a particular emphasis on information reciprocity. 
Among the reviewed articles, only~\cite{bib:15_zheng2022dyhealth} proposes a single fusion method based on a variation of the attention mechanism. 
Specifically, the authors propose a multimodal fusion module based on a modality-based attention mechanism.
This approach allows for the weighted combination of the input unimodal representations, resulting in a latent representation that captures the combined insights from all modalities while adjusting their relative importance in a context-specific manner. 
On the other end, the majority of the works that adopt the \textit{Attention} operation to fuse the unimodal representation adopt a multiple fusion approach. 
This trend is related to the nature of the attention mechanism, which maximizes its effectiveness if integrated multiple times.
As an example, in~\cite{bib:50_sun2023toward} the authors proposed a multiple fusion network designed with cross-attention mechanisms for the inter-modality encoders, able to exploit the inter-modality dependencies among couples of unimodal representations. 
Another example is~\cite{bib:03_ahmad2023aatsn}, where the authors proposed a multiple fusion network, which extracts anatomical CT features, and then intermediately fuses them with PET features through a cross-attention mechanism at multiple depths of the models. 

\paragraph{Calibration}
The calibration category includes those approaches that merge the unimodal representations into a shared representation of reduced dimensions (Squeeze), generating an intermediate multimodal representation.
Then, this representation is used as a calibration signal (Excitation) on all the fusion inputs that generate the signal itself. 
The approaches that employ calibration fusion are classified as multiple fusion approaches, given the necessity to employ other fusions to merge the calibration signal with the original input data. 
Moreover, the generation of the intermediate fused representation is performed generally on features processed with the same number of layers, as evidenced by the majority of synchronous fusion approaches among those that adopt the calibration fusion function~\cite{bib:19_he2021hierarchical, bib:23_pan2021liver, bib:48_radhika2021stress, bib:40_kuttala2023multimodal}. 
As an example, in~\cite{bib:19_he2021hierarchical} a multiple intermediate fusion learning strategy is applied for glioma-grading studies. 
The authors introduced a module to obtain multimodal information interactions, where the output feature maps from the two unimodal representations are first merged, and then a high-order attention mechanism is applied.
Subsequently, the calibration signal obtained is fused with each unimodal input representation.
On the contrary, in two other articles~\cite{bib:09_hou2023deep, bib:04_liu2022attention}, the calibration is performed in an asynchronous manner, generating intermediate fused representation on features processed with different numbers of layers.
As an example in~\cite{bib:04_liu2022attention} the authors adopt the fusion of two modalities through a calibration process, and at the final layers of the architecture, tabular data are concatenated, resulting in a gradual and asynchronous approach to fuse the modalities.

\paragraph{Knowledge-Sharing}
This knowledge-sharing category includes all methodologies that learn to extract a useful representation independently of the input representation. 
In particular, these approaches do not implement a direct interaction between the various unimodal components of the model, instead, they learn the fused multimodal representation through an optimization step, whether it uses a specific loss, a normalization rule (e.g., orthogonalization) or, more simply, back-propagating on a portion of the network that processes different modalities. 
Creating an intermediate embedding space involves mapping features from different modalities into a common multidimensional space where they can be compared or combined. 
This approach often involves training with objective functions that encourage similarities or dissimilarities between the content of different modalities, guaranteeing semantic consistency among the feature embeddings. 
An example is the model proposed in~\cite{bib:29_chen2022ms}, where 2 modalities representations influence each other using a similarity loss, that forces both representations to maintain a fused information consistency between the features spaces. 
Conversely, in~\cite{bib:30_meng2022msmfn} the authors proposed a self-supervised module using an orthogonalization loss to guide the network to learn modality-specific features and maximize the differentiation capability for the final task.
A further example is the model reported in~\cite{bib:20_sedghi2020improving}, where the authors used a dual-stream U-Net structure to implement an intermediate fusion between TeUS and MRI modalities. 
Both streams share the encoding layer weights in the middle of the network, forcing the backpropagation step to optimize the same bottleneck weights of the network for both unimodal-specific decoder updates.

\begin{figure}[t]       
    \resizebox{\linewidth}{!}{
    \begin{tikzpicture}
    \node[label=below:(a)] (single) at (0,0) {
    \includegraphics[height=5cm, trim={4.9cm 0cm 4.9cm 0}, clip]{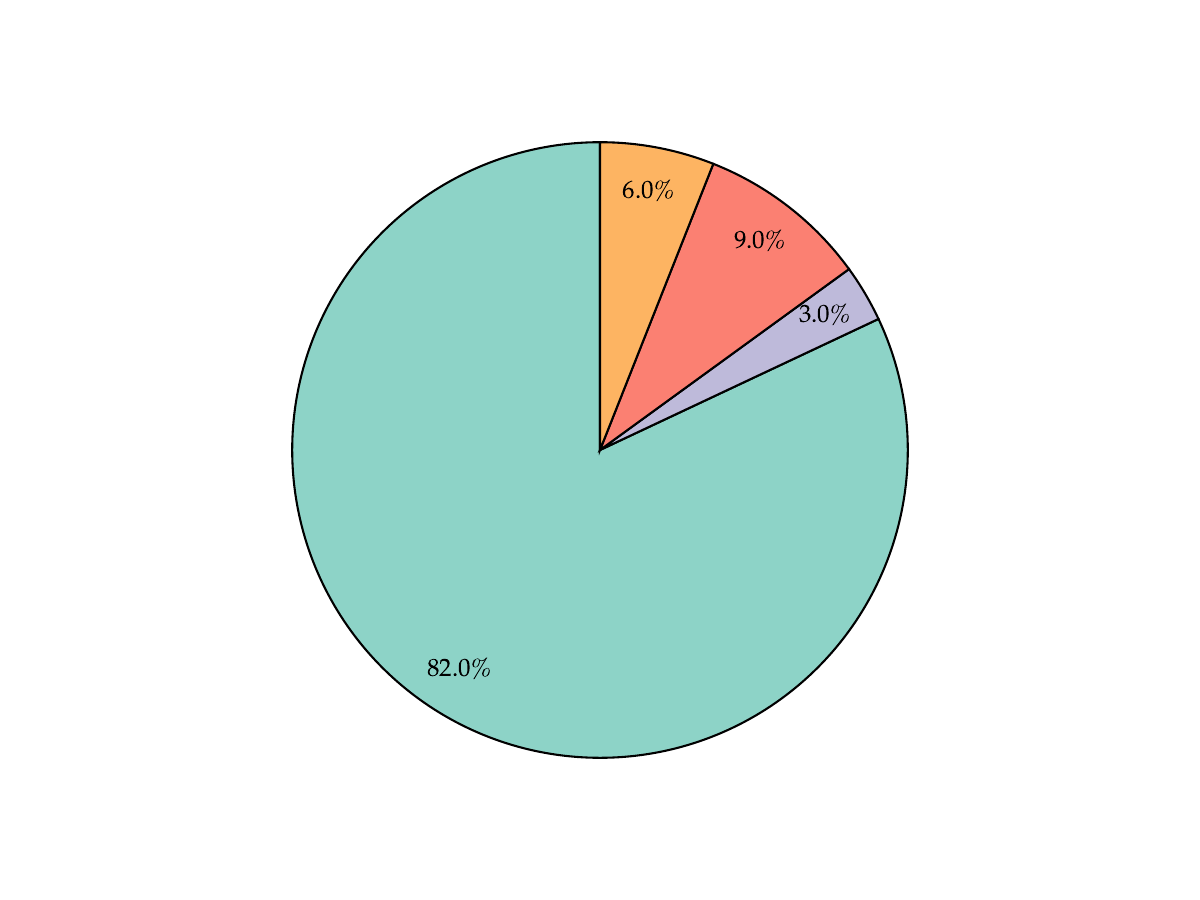}};
    \node[label={[xshift=-.6cm]below:(b)}] [right=.2cm of single] {
    \includegraphics[height=5cm, trim={4.9cm 0cm 0cm 0}, clip]{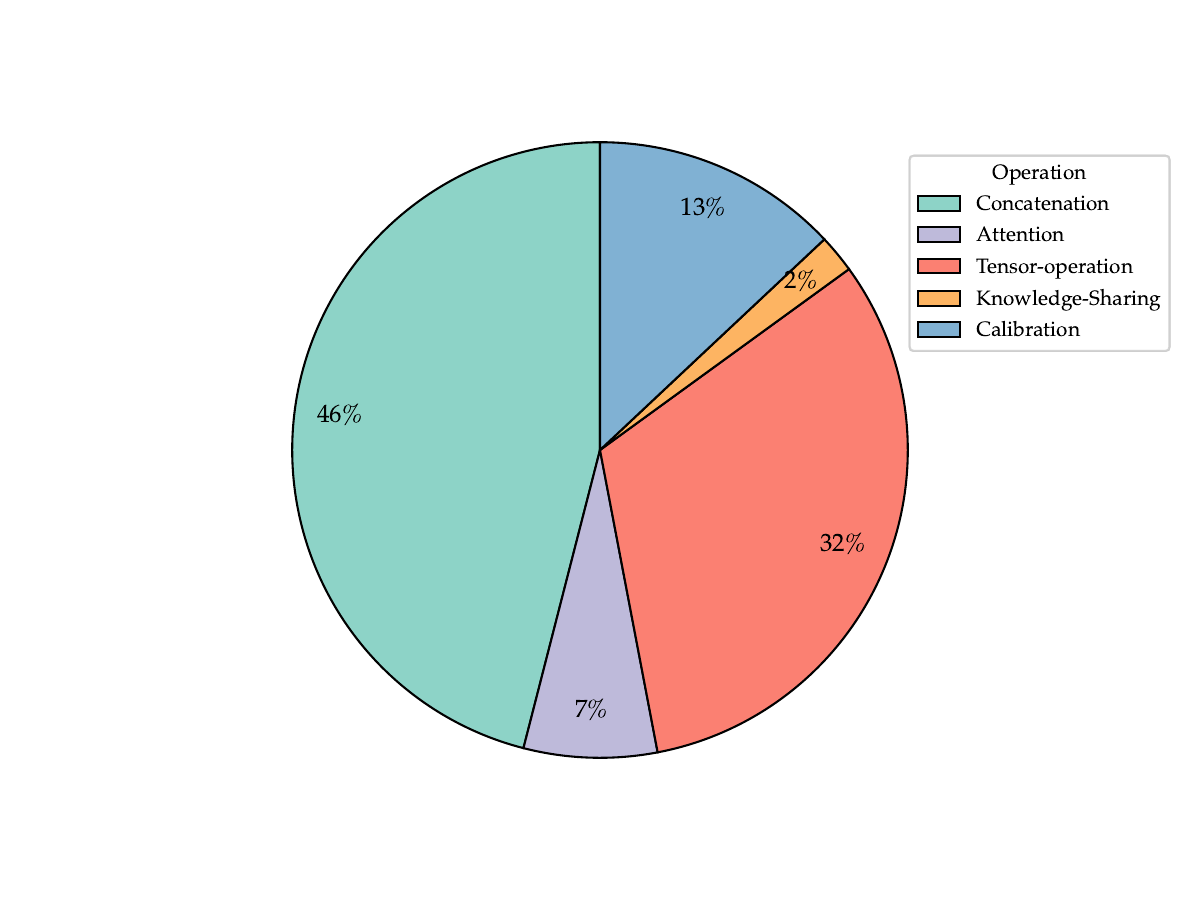}};
    \end{tikzpicture}}
    \caption{Percentage of fusion operation for all the categories, divided by single (a) and multiple (b) fusion approaches.}
    \label{fig:single_multiple}
\end{figure}

\figurename~\ref{fig:single_multiple} presents the distributions of the different categories of fusion operations, distinguished between single and multiple approaches. 
The left panel in the figure displays the percentages of use of specific types of fusion in relation to the total number of single approaches. 

Among the articles employing a single fusion operation, concatenation was the most preferred method, accounting for 82\% of cases. 
This dominance can be attributed to its simplicity and intuitive nature, allowing for straightforward integration of features from different modalities.
Other fusion methods were used less frequently: knowledge-sharing (6\%), tensor operations (9\%), calibration (0\%), and attention (3\%).
This significant imbalance likely stems from concatenation's simplicity and effectiveness in many scenarios, suggesting a data-oriented approach that prioritizes optimizing individual modality representations over complex inter-modal interactions. 
In fact, concatenation allows the subsequent layers of the network to learn the most relevant cross-modal features, which can be advantageous when the optimal fusion strategy is not known a priori.

In contrast, articles using multiple fusion methods still favored concatenation (46\%), but saw a notable increase in tensor operations, rising from 9\% in single fusion methods to 32\%. 
This shift reflects the need for multiple fusion methods for diverse fusion operations throughout the whole fusion module, which often involve tensor or vector manipulations, that lead to a higher level of interactions between the modalities.
Regarding other types of fusion, those based on attention mechanisms are increasingly attracting researchers' interest, accounting for 3\% and 7\% of the fusions in single and multiple architectures, respectively. 
The attention is preferred in multiple fusion approaches given its effectiveness when applied at different stages of the fusion process, and possesses the ability to achieve a rich representation conditioned by multiple modalities, maintaining a certain degree of explainability thanks to attention maps.

The calibration category, which involves using the calibration signal generated in other subsequent fusions, represents an exclusive approach of multiple architectures.
Used in 13\% of cases, it is the third most used method in multiple architectures, as it allows a degree of interactivity between unimodal flows within the model that can be managed according to needs.

Indeed, the same module can be used at a single point to generate a single calibration signal for the modalities, or it can be employed multiple times in succession to gradually learn multimodal representations with increasing degrees of abstraction.

Finally, the knowledge-sharing category, which accounts for 6\% and 2\% of fusions in single and multiple approaches, respectively, emerges as an interesting yet still under-explored approach. 
The higher percentage of knowledge-sharing operations in single fusion methods compared to multiple fusion approaches can be attributed to the unique characteristics of these methodologies.
The preference for knowledge-sharing in single fusion approaches stems from the desire to develop intermediate multimodal methods that centralize this type of fusion, allowing for a more precise evaluation of the effectiveness of this type of fusion.
Conversely, in multiple fusion methods, knowledge-sharing operations often play a supplementary role, enriching intermediate representations rather than serving as the primary fusion mechanism, contributing to a richer and multi-faceted approach to multimodal integration.

These patterns underscore the evolving nature of multimodal fusion strategies, where researchers are increasingly exploring sophisticated combinations of techniques to optimize the integration of diverse modalities. 
As the field progresses, we may see further refinement in the application of the fusion methods presented, potentially leading to more sophisticated and effective fusion architectures.

\subsubsection{Unimodal Module Influence on the Fusion Module}\label{sec:inter_uni_fus}

One of the main aspects that emerges from the analysis is the importance of the dimensionality of the fused space, a crucial factor for the effective integration of unimodal information, because it directly impacts the model's ability to capture and leverage multimodal interactions.
Considering the proportions of the unimodal representations in the fused feature space, only 18 out of the 54 articles reported unimodal feature sizes for all the modalities involved in the fusion. 
Among these 18 articles, 13 of them employ an equal dimension for each unimodal representation, whereas the rest do not adopt this approach and have different dimensions for each unimodal feature vector, as reported in Section~\ref{sec:unimodal_module}. 
The choice of homogeneity of the unimodal modules is not a reason for the equal distribution, given that, among these 13 articles, 7 employ homogeneous unimodal modules, whereas the remaining 6 have heterogeneous ones. 
This almost exact separation between heterogeneous and homogeneous modules shows the trend from the authors to seek a uniform feature space no matter the type of unimodal module involved in processing the unimodal data. 
As described in~\cite{bib:37_cahan2023multimodal}, who opts for using equal marginal representation sizes for each modality, this preference could be to avoid bias toward any modality.
Although it is known that it can help prevent bias, it remains unclear whether using an equal number of features for each modality is a good choice, given the different nature of the modalities involved.

Furthermore, focusing on the 13 articles that prefer an equal distribution of the unimodal representations, we observe that 12 of them utilize a single sudden approach, whereas 1 employs a multiple sudden approach~\cite{bib:41_li2022multimodal}. 
This marked prevalence of single sudden approaches using unimodal representations with equal sizes easily correlates with the goal of uniformly analyzing all the modalities involved, giving them equal importance in the final task.
Moreover, 19 out of 24 homogeneous approaches employ a synchronous approach. 
The choice to fuse unimodal representations with homogeneous and synchronous approaches, further confirms a common interest in exploiting uniform and balanced representation spaces. 
Conversely, the majority of works that employ a heterogeneous approach, 28 out of 30 employ an asynchronous approach, depicting various scenarios, where the fusion occurs between different architectures, data types and fusion operations. 

In the next Section, we will delve into the last module of an intermediate fusion approach, called multimodal module and represented as $f$ in \figurename~\ref{fig:joint_fusion}.
This module, which is provided with the combined features obtained by the fusion module, processes it to extract insights that are only apparent when considering all modalities together.

\subsection{Multimodal Module}\label{sec:multimodal_module}

The next component of the intermediate fusion framework, after the fusion operations that generated a multimodal representation, is the multimodal module, denoted with the $f$ block in \figurename~\ref{fig:joint_fusion}. 
This module leverages the representation $h$ derived from the fusion module \Fusion{}, to obtain the desired target $y$. 
The role of the multimodal module strongly depends on the type of architecture employed and the task for which it is used.
Following our definition, no fusion operation takes place in this module, but only the final processing of the multimodal representation $h$ to perform the task at hand.
In scenarios where the fusion module employs multiple fusion operations, enabling the interaction of various intermediate representations to generate the final fused representation, the multimodal module's role is limited to processing a shallow representation that aligns with the task's representational requirements.
Conversely, when the fusion architecture is simpler, such as in single sudden approaches, the multimodal module assumes a more significant role in processing and refining the multimodal representation.

\begin{figure}[t]
    \centering
    \includegraphics[width=\textwidth]{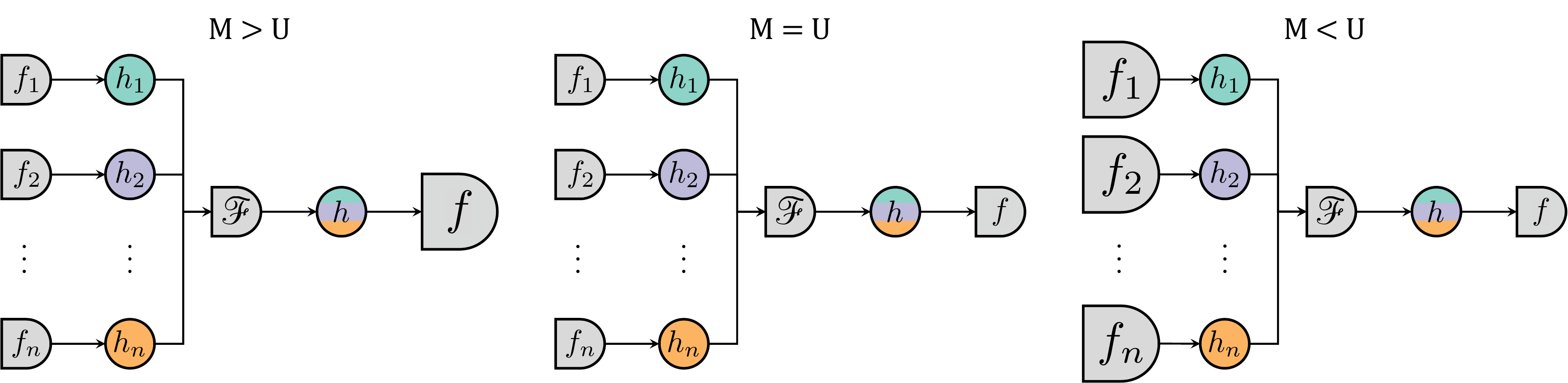}
    \caption{Three cases illustrating size relationships between unimodal and multimodal modules: M$>$U indicates the multimodal module is larger than the unimodal module; M$=$U denotes equal sizes for both modules; M$<$U signifies the multimodal module is smaller than the unimodal module.}
    \label{fig:size}
\end{figure}

In analyzing the multimodal module, a crucial aspect is the proportion of trainable parameters with respect to those used in the unimodal module's architectures, to characterize its impact on feature learning and task optimization.  
In this context, we distinguish 3 cases (\figurename~\ref{fig:size}) with respect to the unimodal module, which are the cases where unimodal modules have a larger (M$<$U) number of parameters, comparable (M$=$U), or smaller (M$>$U) than the multimodal module.
This analysis reflects the role of the module, whether it serves only for final task purposes or acts as a foundational component to process the multimodal features for the final task.

We also characterize the multimodal module on the base of the architecture that it implements, which influences the ability of the model to analyze the multimodal representation $h$. 
Firstly, we categorized the multimodal modules' architecture into the following groups:

\begin{itemize}

\item \textbf{FCNN}: Predominantly employed in both single (29 out of 35) and multiple (17 out of 19) fusion methods, the fully connected architecture plays a crucial role in mapping the feature space to the target space on which the final loss function is computed. This category of multimodal module typically comprises single or multiple fully connected layers. When stacked and combined with non-linear activation functions, FCNNs possess the theoretical capability to approximate any continuous function, rendering them versatile for mapping complex feature spaces to task-specific output spaces. Moreover, FCNNs excel at integrating information from the entirety of the fused feature space simultaneously, making them particularly well-suited for tasks requiring global feature integration. 

\item \textbf{RNN}: This architecture comprises methods that use a multimodal module based on recurrent layers for the prediction task, incorporating single or multiple LSTM or GRU layers to capture dependencies between modalities~\cite{bib:17_cen2022exploring, bib:25_hayat2022medfuse, bib:27_mohammed2023mmhfnet, bib:12_niu2023deep}.
RNNs are characterized by their cyclic connections, allowing them to maintain an internal state or memory that can capture temporal dependencies in the input data. This architecture can process inputs sequentially, making them particularly adept at handling variable-length input sequences and for temporal dependency modeling.

\item \textbf{CNN}: This approach, which features a multimodal module composed of convolutional layers, exploits the ability of the convolutional layers to extract representations from high dimensional data with a lower computational cost compared to the FCNN approach. 
Only 3 works use this type of architecture, 2 employing convolutional layers and a final fully connected layer~\cite{bib:13_radhika2021deep, bib:03_ahmad2023aatsn}, whereas the latter uses only convolutional layers~\cite{bib:20_sedghi2020improving}.
CNNs offer unique strengths in processing structured grid-like data, particularly in extracting hierarchical spatial features. Originally designed for image processing, CNNs have shown remarkable abilities in extracting relevant spatial features from input data, making them particularly effective for modalities with inherent spatial structure. In the multimodal module, CNNs could be considered an alternative way to extrapolate positional-like patterns within the modalities that can match with the final task output.

\item \textbf{Attention}: It includes methods employing attention-based models. 
This approach, exploiting the potential of the attention mechanism, can learn how to leverage information from the interaction of the different modalities. 
Among these works, 2 approaches adopt channel or spatial attention blocks inside their multimodal module~\cite{bib:23_pan2021liver, bib:46_kohankhaki2022radiopaths}, whereas 3 different works adopt state-of-the-art or customized TF architectures~\cite{bib:37_cahan2023multimodal, bib:51_wu2023transformer, bib:54_cahan2022weakly}. 
Attention allows the model to dynamically weigh the importance of different parts of the input or intermediate representations, enabling more efficient and effective information processing. This approach can be implemented in various forms, including self-attention, cross-attention, and multi-head attention.

\item \textbf{None}: This is a peculiar category where no trainable layers are employed for the multimodal module.
Among the reviewed works, we found a single study that employs this strategy~\cite{bib:28_wang2021modeling}.
Here, the authors propose a new intermediate fusion method that optimizes a loss function, considering model uncertainty while estimating correlations among predictions produced by different unimodal modules. 
\end{itemize}

\begin{figure}[t]
    \centering
    \includegraphics[width=12cm, trim={2cm 0 0 0}, clip]{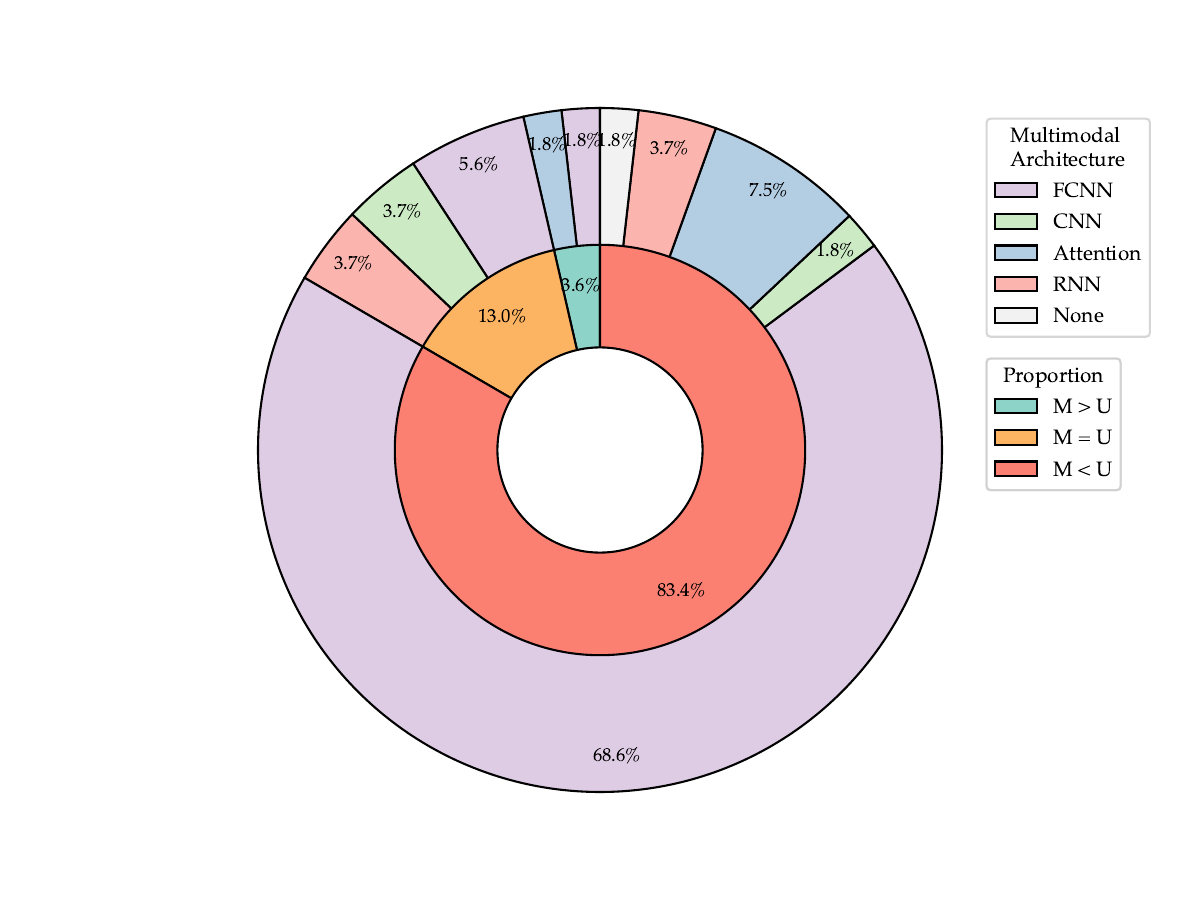}
    \caption{Multimodal modules architectures distributions separated by unimodal modules larger (M$<$U), comparable (M$=$U), or smaller (M$>$U) than the multimodal module.}
    \label{fig:arch_vs_scale}
\end{figure}

\figurename~\ref{fig:arch_vs_scale} illustrates the distribution of all these architectures categorized by M$<$U, M$=$U, M$>$U approaches. 
We notice that most of the reviewed studies employ unimodal models with larger dimensions than those of the multimodal component (45 over 54). 
In contrast, only a small portion of the approaches adopt a comparable dimension (7 over 54) or a multimodal module larger than the unimodal modules' dimension (2 over 54). 

The vast majority of M$<$U approaches adopt a \textit{FCNN} (37 articles) architecture for the multimodal module, whereas 4 methods adopt \textit{Attention} mechanism to improve the ability to elaborate the fused features representation relatively to the final objective tasks.
In these works, the responsibility for learning and elaborating the features representation spaces is mainly shifted to the fusion module. 
Indeed, in this context, the multimodal module concentrates on fine-tuning the learned representations for the specific objectives of the final task, acting as the last step in a series of transformations to process the fused features for the downstream task.

Conversely, a small portion of the works adopts an unimodal module of equal or smaller dimensions with respect to the multimodal module (red and green slices of the inner circle in \figurename~\ref{fig:arch_vs_scale}). 
In these few cases, the multimodal module's role is to transform the initially fused data into a more sophisticated representation useful for the final task, which is capable of exploiting the underlying similarities between the combined modalities in the most effective way. 
The general trend among all the works underlines the central role of the fusion module rather than the multimodal module. 
Indeed, this component serves as an integration point but also as a dynamic learning environment where the interplay of different modalities can be explored and optimized.
In contrast, most of the approaches use the multimodal module mainly to make the architecture compatible with the downstream tasks.

In the next Section, we will further discuss the downstream biomedical tasks, deepening some other meaningful characteristics of the methods, comprehending transfer learning, explainability, missing modality robustness and the experimental configurations.

\subsection{Target} \label{sec:target}
\begin{figure}[t]
\centering
\includegraphics[width=10cm]{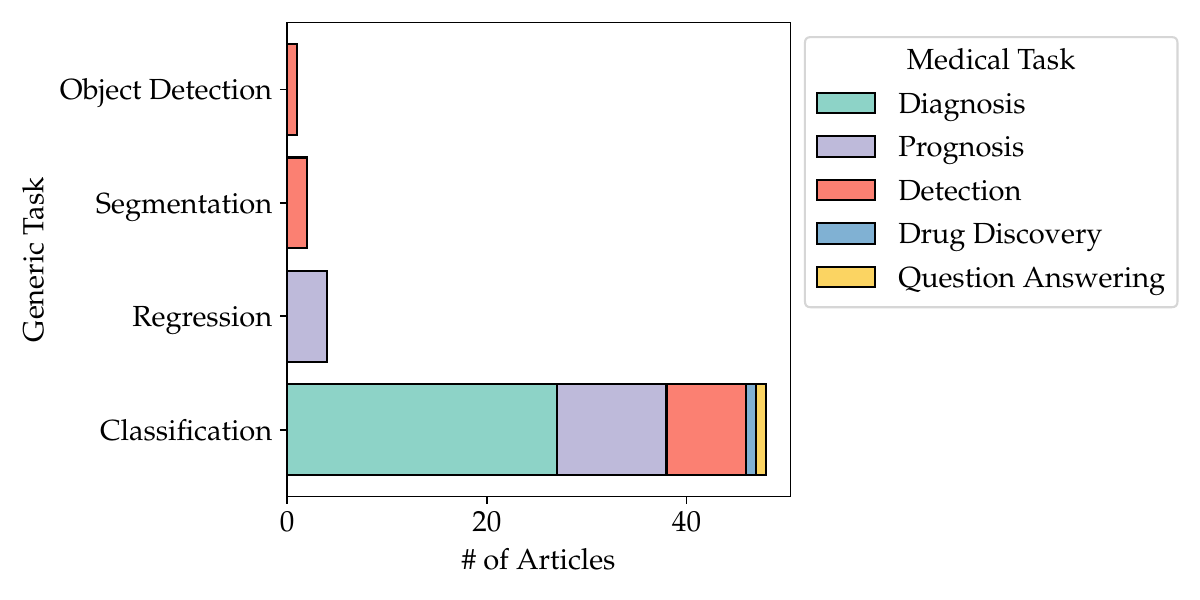}
\caption{Tasks involved in intermediate fusion studies. Each bar of the bar plot refers to the generic task type (classification, regression, segmentation, object detection) whilst the color legend refers to the medical task involved (diagnosis, prognosis, detection, drug discovery, or question answering). Each task addressed within the same article is considered an individual work. As a consequence, the total count depicted in the bar plot may surpass the number of selected articles.}
\label{fig:tasks}
\end{figure}

In MDL models, the target represents the output $y$, as outlined in \figurename~\ref{fig:joint_fusion}, which is functional for the task's execution.
In the context of biomedical applications, this could be a diagnostic classification, a prognostic regression, or any other relevant task that requires the definition of an outcome.
All selected articles perform supervised tasks, involving the use of annotated datasets in which all data instances are labeled. We categorized each article based on the generic task and the medical task. For the first, we identified 4 categories: classification, regression, segmentation, and object detection. The second instead divides into 5 categories: diagnosis, prognosis, detection, drug discovery, and question answering. \figurename~\ref{fig:tasks} shows that the majority perform classification tasks (47 out of 54) with a prevalence of diagnostic tasks (26 out of 54) and prognostics tasks (12 out of 54). Some perform classification tasks for detection purposes (7 out of 54), as in~\cite{bib:40_kuttala2023multimodal} where the authors design a model for stress detection or
in~\cite{bib:17_cen2022exploring} where they develop a system for detecting protective behaviors in patients with chronic pain. Only one article addresses classification for drug discovery~\cite{bib:45_seo2021predicting}, while another focuses on classification for question answering~\cite{bib:01_li2022bi}.
In total, 3 articles perform regression, addressing prognostic tasks~\cite{bib:11_oh2023deep,bib:35_steyaert2023multimodal,bib:38_subramanian2020multimodal}.
All proposed applications are single-task, except for 2 works: in~\cite{bib:20_sedghi2020improving}, the authors propose an approach for prostate cancer detection by integrating segmentation and prediction of the cancer likelihood for each pixel; in~\cite{bib:23_pan2021liver}, the authors develop a system for liver tumor detection including classification and bounding box regression for the detection stage.

\begin{figure}[t]
\centering
\includegraphics[width=13cm]{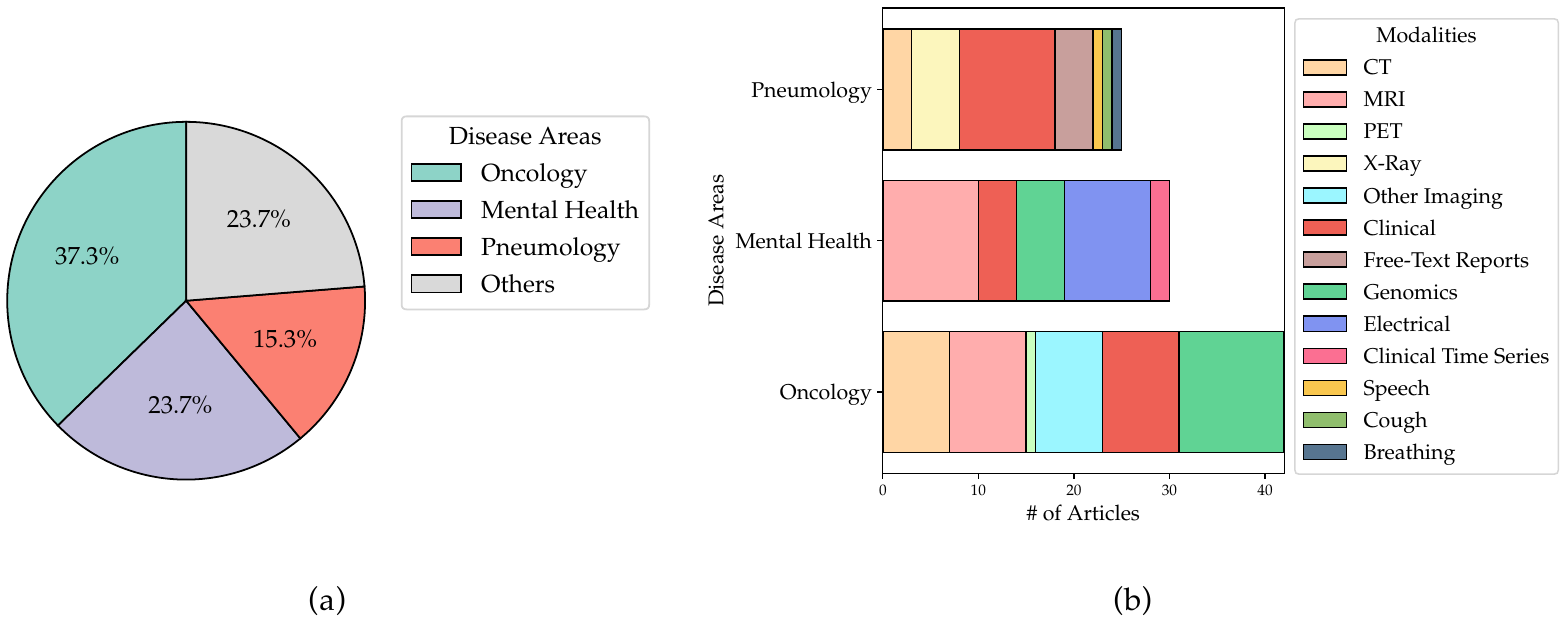}
\caption{(a) Disease areas involved in intermediate fusion studies. Others: visual question answering~\cite{bib:01_li2022bi}, visual impairment~\cite{bib:05_li2022aviper}, overall survival form discharge~\cite{bib:12_niu2023deep}, kidney injury~\cite{bib:15_zheng2022dyhealth}, chronic pain~\cite{bib:17_cen2022exploring}, placental insufficency~\cite{bib:18_jiao2023gmrlnet}, osteoartritis~\cite{bib:21_guida2023improving}, voice pathology~\cite{bib:27_mohammed2023mmhfnet}, heart failure~\cite{bib:32_bhattacharya2022multi}, glaucoma~\cite{bib:41_li2022multimodal}, diabetic rethynopathy ~\cite{bib:41_li2022multimodal}, heart failure mortality prediction~\cite{bib:44_ma2023predicting}, phenotype classification~\cite{bib:25_hayat2022medfuse}, all-cause mortality prediction~\cite{bib:25_hayat2022medfuse}; (b) Exploring data modalities across top three disease areas.}
\label{fig:disease_areas}
\end{figure}

Focusing on the disease areas involved in intermediate fusion studies, we identified three main clusters, as shown in \figurename~\ref{fig:disease_areas} (a): oncology (22 out of 54), mental health (14 out of 54), and pneumology (9 out of 54). Due to the diverse range of application areas, the remaining publications were grouped into a single cluster labeled as ``Others'' reporting the specific application for each article. 
Moreover, our analysis highlights how the nature of the disease influences the choice and use of different data modalities. As shown in \figurename~\ref{fig:disease_areas} (b), oncology relies mostly on genomics data, followed by clinical data, CT, and MRI imaging. Conversely, mental health prioritizes MRI and electrical data, e.g., EEG, ECG, whilst pneumology relies mainly on clinical data and X-ray imaging.

\subsection{Learning} \label{sec:learning}
As we described the key components of an MDL model in the previous sections, we now turn our attention to pivotal techniques for training and deploying MDL models in real-world scenarios.

\subsubsection{Transfer Learning} \label{subsec:transfer_learning}
\begin{figure}[t]
\centering
\includegraphics[width=\textwidth]{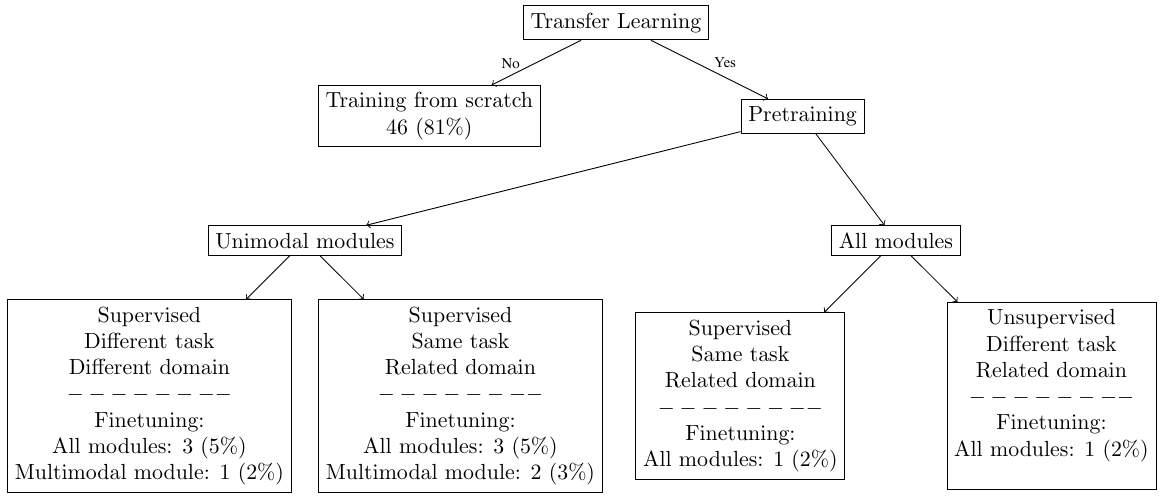}
\caption{Transfer Learning in intermediate fusion methods. The tree diagram distinguishes between articles training from scratch the entire MDL model and articles pretraining the unimodal modules in isolation or pretraining the entire network.}
\label{fig:Transfer_Learning}
\end{figure}

Although deep learning has achieved remarkable success, the hunger for data hinders its full potential in the medical field where data scarcity is fueled by high costs, privacy, and ethical issues~\cite{bib:qiu2023pre}.
To this end, transfer learning has been proposed to mitigate the problem. Inspired by human learning, transfer learning allows the exploitation of knowledge acquired in a similar or different domain, reducing the need to train the entire MDL model from scratch and thus limiting the need for large amounts of data in the domain of interest. Transfer learning consists of two macro-steps: first, the model is trained on large datasets, a task usually referred to as pre-training, to make the model learn general unimodal and/or multimodal feature representations that could be valuable across different related tasks. Second, the pre-trained model is tailored to a specific task, usually leveraging smaller datasets. We refer to this phase as finetuning. 
Following the tree diagram in \figurename~\ref{fig:Transfer_Learning}, the majority (46 out of 54) train their models from scratch, whilst the remaining (11 out of 54) utilize pre-training, dividing into those pre-training only the unimodal modules (9 out of 54) and those pre-training the entire model (2).
With respect to the task at hand, there are different approaches for pre-training a MDL model each with its variations. Some opt to pre-train the unimodal modules in a different supervised task and a different domain (4 out of 54). Following this step, they proceed with either finetuning all modules (3 out of 54) or freezing the unimodal modules while finetuning the multimodal module (1 out of 54). Conversely, some others choose to pre-train the unimodal modules in the same supervised task and related domain (5 out of 54). Subsequently, they either finetune all modules (3 out of 54) or freeze the unimodal modules and finetune the multimodal module in isolation (2 out of 54).
Among those pre-training the entire model, we identify those pretraining in the same supervised task and related domain (1 out of 54), and those pretraining in a different unsupervised task and related domain (1 out of 54), followed by finetuning the entire  MDL model.
Though the lack of large datasets in the medical domain might make transfer learning the optimal solution to address data scarcity, our analysis shows that transfer learning is not widely adopted. Indeed, although it has been successfully employed in several tasks, transfer learning has raised some skepticism especially when employed in the medical field~\cite{bib:qiu2023pre}. In recent work, Matsoukas et al.~\cite{bib:matsoukas2022makes} highlight that the core principle of transfer learning is the increasing reuse of learned representations. Therefore, optimal outcomes are achieved when pre-training occurs in domains similar to the specific domain under consideration. This discourages the use of pre-trained models on generic domains like the well-known ImageNet~\cite{russakovsky2015imagenet}. Although this approach might contribute to speeding up model convergence, it does not guarantee superior performance or more meaningful learned representations. Therefore, the lack of pre-trained backbone architectures on medical domains makes transfer learning a sub-optimal approach to address data scarcity, which paves the way for alternative solutions such as generative data augmentation~\cite{bib:ramachandram2017deep}. While our work does not cover generative data augmentation, interested readers can refer to~\cite{mumuni2022data} for further information on this topic.
Focusing on the articles that perform transfer learning, all utilize one or more pre-trained CNNs~\cite{bib:06_lobantsev2020comparative,bib:08_menegotto2020computer,bib:21_guida2023improving,bib:24_ostertag2023long, bib:25_hayat2022medfuse, bib:26_zeng2022miftp, bib:33_guarrasi2023multi, bib:35_steyaert2023multimodal, bib:46_kohankhaki2022radiopaths, bib:51_wu2023transformer}. Two articles use a pre-trained RNN~\cite{bib:06_lobantsev2020comparative, bib:25_hayat2022medfuse}, one uses a pre-trained FCNN~\cite{bib:24_ostertag2023long}, and another employs a TF~\cite{bib:46_kohankhaki2022radiopaths}. We observed that among those pre-training their networks, 7 utilize well-known CNN architectures, while 3 opted to pre-train their custom models~\cite{bib:21_guida2023improving, bib:24_ostertag2023long, bib:51_wu2023transformer}. The well-known CNN architectures include ResNet-based models~\cite{bib:06_lobantsev2020comparative, bib:25_hayat2022medfuse, bib:26_zeng2022miftp, bib:33_guarrasi2023multi, bib:35_steyaert2023multimodal}, InceptionV3~\cite{bib:08_menegotto2020computer} and EfficientNet~\cite{bib:46_kohankhaki2022radiopaths}. While the pre-trained RNNs and FCNN are custom-designed, for the TF, a BERT-based model is used.

\subsubsection{Multimodal optimization}
In a MDL model, the optimization strategy can significantly influence the multimodal representation, \emph{h}, returned by the fusion module 
\Fusion{}, either through indirect or direct effects. Indirect effects occur when the loss function operates on the output $y$, thus indirectly influencing the update of the multimodal representation \emph{h}. Conversely, direct effects involve loss functions that act directly on the feature representations, promoting their interaction and complementarity.
Based on our analysis, 51 articles fall into indirect effects, with the majority of them using standard loss functions (e.g., cross-entropy, mean-squared error, focal loss).
Among the indirect effects, it is notable that some approaches consider the multimodal nature of the model when formulating the loss function. This is done with the explicit intent of guiding and balancing the influence of each modality in defining the multimodal representation, \emph{h}.  In~\cite{bib:18_jiao2023gmrlnet}, the authors use a contrastive learning strategy, using the triplet loss to decouple modality shared and modality-specific feature spaces. To balance the informativeness of each modality before fusion, \cite{bib:36_han2022multimodal} quantifies the multimodal confidence by computing the true class probability which is then used to weight each unimodal representation before feeding to \Fusion{}. Similarly, \cite{bib:28_wang2021modeling} models the uncertainty of individual modalities produced by different learners, thus capturing the pairwise correlation between predictions. To ensure all modalities can equally contribute to the model's output, preventing the model from being biased towards a subset of modalities, \cite{bib:14_rahaman2023deep,bib:31_rahaman2021multi} use a multimodal regularization~\cite{bib:gat2020removing} based on functional entropy whilst~\cite{bib:35_steyaert2023multimodal} uses different learning rates for the unimodal modules to balance the learning of different modalities.
Turning our attention to direct effects, we classify them into intra-modal and inter-modal effects. The former involves architectures that use multiple unimodal modules for each modality, aiming to extract distinct representations for a single modality. For instance, in~\cite{bib:29_chen2022ms}, the authors impose an orthogonalization constraint within the individual unimodal embeddings to help the separation between modal shared and modal specific features. As for the inter-modal effects, we find methodologies that fall into the category of knowledge-sharing fusion~\cite{bib:19_he2021hierarchical,bib:29_chen2022ms, bib:30_meng2022msmfn} as we describe in Section~\ref{sec:fusion_module}.

\subsubsection{Missing Modalities}\label{sec:missing}
\begin{figure}[t]
\centering
\includegraphics[width=8cm]{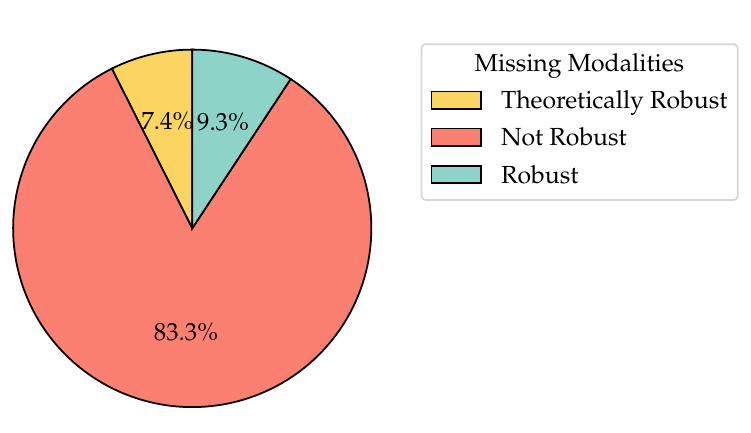}
\caption{Robustness to missing modalities. The pie chart highlights the articles whose architectures do not handle missing modalities, those that are designed to be robust against missing modalities, and those that although not specified by the authors could potentially handle them.}
\label{fig:missing_modalities}
\end{figure}

The term missing modalities denotes the partial or complete absence of one or more modalities within a multimodal instance. 
In the context of multimodal learning, the presence of missing modalities is a major concern that can undermine the applicability of a model, especially in data-in-the-wild scenarios where incomplete instances are prevalent.
In the medical field, this problem is even more pronounced as the combination of missing modalities and data scarcity severely limits the availability of large amounts of data. 
Looking at \figurename~\ref{fig:missing_modalities}, the majority of the articles (45 out of 54) showcase models that lack robustness to missing modalities. The main criticality lies in the way the modalities are fused with a prevalence of simple concatenation-based fusion that requires complete data instances.
Only a few (5 out of 54) claim their models can handle missing modalities. In their work, Kohankhaki et al.~\cite{bib:46_kohankhaki2022radiopaths}  propose an attention-based mechanism for the fusion layer followed by a pooling operation, which is permutation invariant to its inputs, making the absence of some modalities unproblematic. Alternatively, Hayat et al.~\cite{bib:25_hayat2022medfuse} treat the multimodal representation as a sequence of unimodal tokens, such that the fusion module performs the aggregation through a recurrence mechanism, e.g., LSTM. A further approach is introduced by Zheng et al.~\cite{bib:15_zheng2022dyhealth}, where an exponential decreasing or increasing mechanism is used to gradually plug out or plug in a specific data modality. To simulate a missing modality scenario, Ostertag et al.~\cite{bib:43_ostertag2020predicting} propose a training strategy that randomly imputes values from the training set with zeros. They indeed demonstrate that training the multimodal architecture with randomly missing data increases its performance. Leveraging a GNN, Jiao et al.~\cite{bib:18_jiao2023gmrlnet} first build a graph-based manifold knowledge on complete data samples which is then transferred to incomplete data scenarios, so that when a modality is missing, its knowledge is derived from the constructed manifold information.  
The remaining 4 articles, while not explicitly addressing the problem, present approaches that potentially could handle missing modalities. This is achieved either through fusion using attention-based techniques~\cite{bib:37_cahan2023multimodal,bib:51_wu2023transformer, bib:54_cahan2022weakly} or through element-wise tensor operations~\cite{bib:19_he2021hierarchical}.

\subsubsection{Explainability} \label{sec:xai}
To date, we are witnessing a surge in the performance of DL models due to larger databases and increased computational power ~\cite{bib:samek2019towards}.
Just as task complexity is increasing, model architectures are also becoming more complex, featuring nested nonlinear structures that turn them into black boxes hiding the decision-making process.
Lack of interpretability and transparency hinders the use of DL-based systems in many real-world applications, such as in the medical domain. Indeed, in medical applications, the possibility of establishing a connection between explainability and causality is crucial as it allows human experts to understand the decision-making process of an AI system~\cite{bib:33_guarrasi2023multi} but also acts as a prerequisite for new insights~\cite{bib:samek2019towards}. Indeed, the possibility to unveil new patterns in data as well as novel associations, e.g., between genes and diseases, through model understanding and introspection is often more crucial than the prediction itself. 

To date, explainable AI (XAI) offers several tools to improve the trust and transparency of AI models, but it is still in its infancy, especially in the context of MDL.
This is confirmed by our analysis with the majority of articles (34 out of 54) not addressing XAI in their works or only mentioning it as a future development. 

The interested reader will find several taxonomies of XAI techniques in the literature, each of which groups them according to a certain criterion~\cite{bib:joshi2021review}. For example, an explanation can be model agnostic or model specific based on its dependence on the model it is trying to explain, or it can be local or global depending on whether the scope is to derive predictions for a specific data point or the entire model. Moreover, explainability can be introduced at different stages of model development, producing pre-modeling, during modeling, and post-hoc modeling explanations. 

The specific focus of our review on MDL, leads us to a further categorization of XAI techniques based on the module of the MDL model on which the explanations are produced. This classification results in a distinction between unimodal and multimodal explanations. The former aim to highlight salient features within a single data modality, whilst the latter focus on modality importance, quantifying the heterogeneity of information across modalities as well as their interactions. In \tablename~\ref{tab:xai}, we list unimodal and multimodal XAI approaches used in the selected articles with their respective reference.
Due to their extensive adoption and established success in various domains, CNNs~\cite{bib:03_ahmad2023aatsn,bib:04_liu2022attention,bib:05_li2022aviper,bib:20_sedghi2020improving, bib:26_zeng2022miftp,bib:30_meng2022msmfn,bib:32_bhattacharya2022multi,bib:33_guarrasi2023multi,bib:35_steyaert2023multimodal,bib:36_han2022multimodal,bib:37_cahan2023multimodal,bib:40_kuttala2023multimodal,bib:46_kohankhaki2022radiopaths,bib:48_radhika2021stress,bib:50_sun2023toward} are the most explained architecture type, followed by FCNNs~\cite{bib:14_rahaman2023deep,bib:31_rahaman2021multi,bib:32_bhattacharya2022multi,bib:33_guarrasi2023multi,bib:35_steyaert2023multimodal,bib:46_kohankhaki2022radiopaths,bib:50_sun2023toward}.
Only three publications provide explanations for RNNs~\cite{bib:14_rahaman2023deep, bib:15_zheng2022dyhealth, bib:31_rahaman2021multi}, two for TFs~\cite{bib:37_cahan2023multimodal, bib:46_kohankhaki2022radiopaths}, one for GNNs~\cite{bib:18_jiao2023gmrlnet}, and another for AEs~\cite{bib:14_rahaman2023deep}.

\begin{table}[t]
    \centering
    \resizebox{\linewidth}{!}{
    \begin{tabular}{|c|c|c|}
    \hline
    \textbf{Unimodal XAI} & \textbf{Multimodal XAI} & \textbf{Articles} \\ \hline
    Saliency Maps & Modality-wise Attention, Correlation of Interaction & ~\cite{bib:14_rahaman2023deep, bib:31_rahaman2021multi} \\ \hline
    GradCAM, Integrated Gradient & - & ~\cite{bib:33_guarrasi2023multi} \\ \hline
    - & Modality Ablation & ~\cite{bib:12_niu2023deep} \\ \hline
    Ablation, LIME & - & ~\cite{bib:32_bhattacharya2022multi} \\ \hline
    SHAP, Saliency Maps & - & ~\cite{bib:35_steyaert2023multimodal} \\ \hline
    GradCAM & - & ~\cite{bib:05_li2022aviper, bib:26_zeng2022miftp} \\ \hline
    3D t-SNE & 3D t-SNE & ~\cite{bib:48_radhika2021stress} \\ \hline
    CAM & - & ~\cite{bib:04_liu2022attention} \\ \hline
    Guided-backprojection & - & ~\cite{bib:03_ahmad2023aatsn} \\ \hline
    LayerCAM & - & ~\cite{bib:30_meng2022msmfn} \\ \hline
    Attention Maps & - & ~\cite{bib:50_sun2023toward} \\ \hline
    CAM, UMAP & CAM & ~\cite{bib:18_jiao2023gmrlnet} \\ \hline
    Feature Importance & Modality Importance & ~\cite{bib:36_han2022multimodal} \\ \hline
    GradCAM++, Integrated Gradients & - & ~\cite{bib:46_kohankhaki2022radiopaths} \\ \hline
    GradCAM, Feature Importance, t-SNE & t-SNE & ~\cite{bib:37_cahan2023multimodal} \\ \hline
    Masking & - & ~\cite{bib:20_sedghi2020improving} \\ \hline
    t-SNE & - & ~\cite{bib:40_kuttala2023multimodal} \\ \hline
    Feature Importance & Modality-Wise Attention & ~\cite{bib:15_zheng2022dyhealth} \\ \hline
    \end{tabular}}
    \caption{XAI approaches for unimodal and multimodal explanations used in the selected articles.}
    \label{tab:xai}    
\end{table}

\subsubsection{Experimental Configuration} \label{subsec:experimental_configuration}
Designing a comprehensive experimental setup is an essential requirement for objectively assessing MDL models. Based on our analysis, we deem that a thorough experimental setting should include: (\textbf{A}) conducting all the experiments using a robust resampling technique (e.g., cross-validation, bootstrap sampling, jackknife sampling, leave-one-out cross-validation, stratified sampling, random sampling); (\textbf{B}) comparing unimodal and multimodal configurations; (\textbf{C}) comparing various fusion strategies, i.e., early, intermediate, and late; (\textbf{D}) conducting structured statistical tests to determine the statistical significance of model performance; (\textbf{E}) provide for the use of external validation, thus assessing the model's generalization.

\tablename~\ref{tab:experimental_configuration} shows the experimental setting adopted by the selected articles, bringing out that the literature is still lacking in using comprehensive experimental settings. First of all, the use of robust resampling technique (\textbf{A}), which is pivotal for ensuring the reliability and robustness of experimental findings, is limited to just 19 articles. When comparing different experiments, it's typical to compute numerical metrics (e.g., accuracy, F1 score, dice score, etc.) relevant to the task at hand and compare their averages. However, merely comparing average scores is not enough to determine the superiority of one approach over another. A higher average does not necessarily signify a better approach since a wider dispersion of individual results might indicate greater instability. This is why, the average performance of an experiment should always be reported along with their standard deviations. 

\begin{figure}[t]
\centering
\includegraphics[width=12cm]{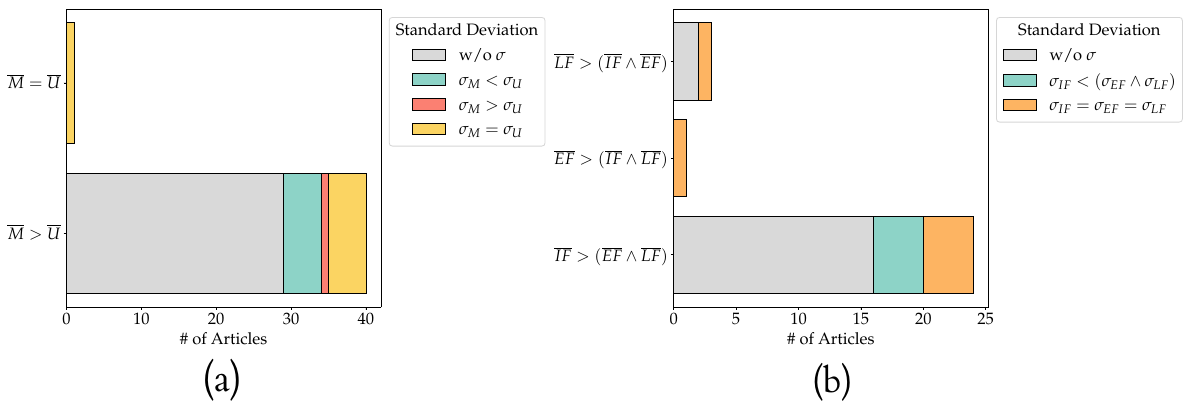}
\caption{(a) Average performance of multimodal approaches compared to unimodal approaches. $\overline{U}$ denotes average unimodal performance, $\overline{M}$ denotes average multimodal performance, \emph{$\sigma$} denotes the standard deviation; (b) Average performance of intermediate fusion approaches compared to other fusion techniques. $\overline{EF}$ denotes average early fusion performance, $\overline{IF}$ denotes average intermediate fusion performance, $\overline{LF}$ denotes average late fusion performance, \emph{$\sigma$} denotes the standard deviation.}
\label{fig:uni_multi}
\end{figure}

Although our analysis has focused on works addressing fusion techniques, the question  ``Is multimodal better than unimodal?'' is still central in the research community and is preliminary to the investigation of how to represent and combine different data modalities.
Indeed, 42 articles include \textbf{B} in their setting. 
In \figurename~\ref{fig:uni_multi} (a), we show that 40 works demonstrate the superiority of the multimodal approach over the unimodal approach. 
However, 29 out of these 40 report only the average performance without standard deviation. Among those reporting the standard deviation, 5 confirm the superiority of the multimodal approach by showing lower standard deviations than the unimodal approach, 5 report equal standard deviations, and 1 shows higher standard deviations for the multimodal approach.
Going back to \tablename~\ref{tab:experimental_configuration}, \textbf{C} is second in position with 31 
articles performing a comparison between fusion strategies. In \figurename~\ref{fig:uni_multi} (b),
we show that intermediate fusion outperforms early and late fusion in 24 works. Again, the majority do not report standard deviations (16 out of 24). Among those reporting the standard deviation, 4 corroborate the superiority of intermediate fusion by showing lower standard deviations whilst the other 4 report equal standard deviations for all fusion strategies, i.e., early, intermediate, and late.
Only 1 article shows that early fusion outperforms intermediate and late fusion with no difference in standard deviation, whilst 3 articles claim late fusion as the best fusion technique of which 2 do not report standard deviations and 1 shows equal standard deviations between early, intermediate, and late fusion.
Only 14 articles assess the statistical significance of their results through structured statistical tests (\textbf{D}). Furthermore, only 5 use an external validation set in their experiments (\textbf{E}). 
The use of a complete and comprehensive experimental procedure including \textbf{A}, \textbf{B}, \textbf{C}, \textbf{D} and \textbf{E} is limited to only 4 articles. 
In addition, only 11 articles make their code public thus not facilitating transparency and reproducibility. 
 
\begin{table}[t]
    \centering
    \resizebox{\linewidth}{!}{
    \begin{tabular}{|l|c|c|}
    \hline
    \textbf{Experimental Configuration} & \textbf{\# of Articles} & \textbf{References}\\ \hline
    \textbf{$\varnothing$} & 2 &~\cite{bib:40_kuttala2023multimodal,bib:48_radhika2021stress} \\ \hline
    \textbf{A} & 2 &~\cite{bib:22_zhang2023itcep,bib:24_ostertag2023long} \\ \hline
    \textbf{B} & 13 &~\cite{bib:02_li2022dynamic,bib:05_li2022aviper,bib:08_menegotto2020computer,bib:09_hou2023deep,bib:11_oh2023deep,bib:12_niu2023deep,bib:13_radhika2021deep,bib:20_sedghi2020improving,bib:26_zeng2022miftp,bib:27_mohammed2023mmhfnet,bib:38_subramanian2020multimodal,bib:40_kuttala2023multimodal,bib:47_sousa2023single,bib:49_rashid2023tinym2net} \\ \hline
    \textbf{C} & 4 &~\cite{bib:07_menegotto2021computer,bib:10_njoku2022deep,bib:44_ma2023predicting,bib:46_kohankhaki2022radiopaths} \\ \hline
    \textbf{D} & 1 &~\cite{bib:01_li2022bi} \\ \hline
    \textbf{A} $\land$ \textbf{B} & 1 &~\cite{bib:42_shetty2023multimodal} \\ \hline
    \textbf{B} $\land$ \textbf{C} & 6 &~\cite{bib:06_lobantsev2020comparative,bib:15_zheng2022dyhealth,bib:16_holste2021end,bib:23_pan2021liver,bib:37_cahan2023multimodal,bib:53_rahaman2022two} \\ \hline
    \textbf{B} $\land$ \textbf{D} & 3  &~\cite{bib:18_jiao2023gmrlnet,bib:29_chen2022ms,bib:32_bhattacharya2022multi,bib:43_ostertag2020predicting} \\ \hline
    \textbf{C} $\land$ \textbf{D} & 1 &~\cite{bib:36_han2022multimodal} \\ \hline
    \textbf{A} $\land$ \textbf{B} $\land$ \textbf{C} & 7 &~\cite{bib:04_liu2022attention,bib:14_rahaman2023deep,bib:21_guida2023improving,bib:25_hayat2022medfuse,bib:28_wang2021modeling,bib:31_rahaman2021multi,bib:34_perez2022multi} \\ \hline
    \textbf{A} $\land$ \textbf{B} $\land$ \textbf{D} & 1 &~\cite{bib:30_meng2022msmfn} \\ \hline
    \textbf{A} $\land$ \textbf{C} $\land$ \textbf{D} & 2 &~\cite{bib:52_alsherbiny2021trustworthy,bib:45_seo2021predicting}  \\ \hline
    \textbf{B} $\land$ \textbf{C} $\land$ \textbf{D} & 4 &~\cite{bib:18_jiao2023gmrlnet,bib:39_huang2020multimodal,bib:41_li2022multimodal,bib:54_cahan2022weakly} \\ \hline
    \textbf{A} $\land$ \textbf{B} $\land$ \textbf{C} $\land$ \textbf{E} & 1 &~\cite{bib:03_ahmad2023aatsn} \\ \hline
    \textbf{A} $\land$ \textbf{B} $\land$ \textbf{C} $\land$ \textbf{D} & 2 &~\cite{bib:17_cen2022exploring,bib:50_sun2023toward} \\ \hline
    \textbf{A} $\land$ \textbf{B} $\land$ \textbf{C} $\land$ \textbf{D} $\land$ \textbf{E} & 4 &~\cite{bib:19_he2021hierarchical,bib:33_guarrasi2023multi,bib:35_steyaert2023multimodal,bib:51_wu2023transformer} \\ \hline
    \end{tabular}}
    \caption{Experimental configurations used in intermediate fusion studies. \textbf{A}: cross-validation scheme, \textbf{B}: comparison between unimodal and multimodal configurations, \textbf{C}: comparison between different fusion strategies, i.e., early, intermediate, and late, \textbf{D}: statistical tests, \textbf{E}: external validation set, \textbf{$\varnothing$}: none of the previous but only comparisons of intermediate fusion configurations. The symbol $\land$ denotes the logic operator AND. The central column shows the number of articles using a specific experimental setting whilst the right column reports the references.}
    \label{tab:experimental_configuration}
\end{table}

\section{Discussion and Conclusion}

As we draw this systematic review to a close, it is essential to step back and contemplate the current limitations, the trajectory for future research as well as the practical impact of our findings.
This section synthesizes the insights garnered from our in-depth analysis of intermediate fusion methods in MDL for biomedical applications.
As mentioned above, this survey includes two supplementary documents for detailed information on individual papers, enhancing readability: Supplementary Material A contains a table summarizing all fields discussed in Sections X to X for each paper, Supplementary Material B analyzes the fusion mechanisms of each paper, showing the corresponding fusion notation and graphical representation.

\subsection{Limitations, Research Gaps and Future Directions}

\subsubsection{Limitations}

\begin{figure}[t]
\centering
\includegraphics[width=\textwidth]{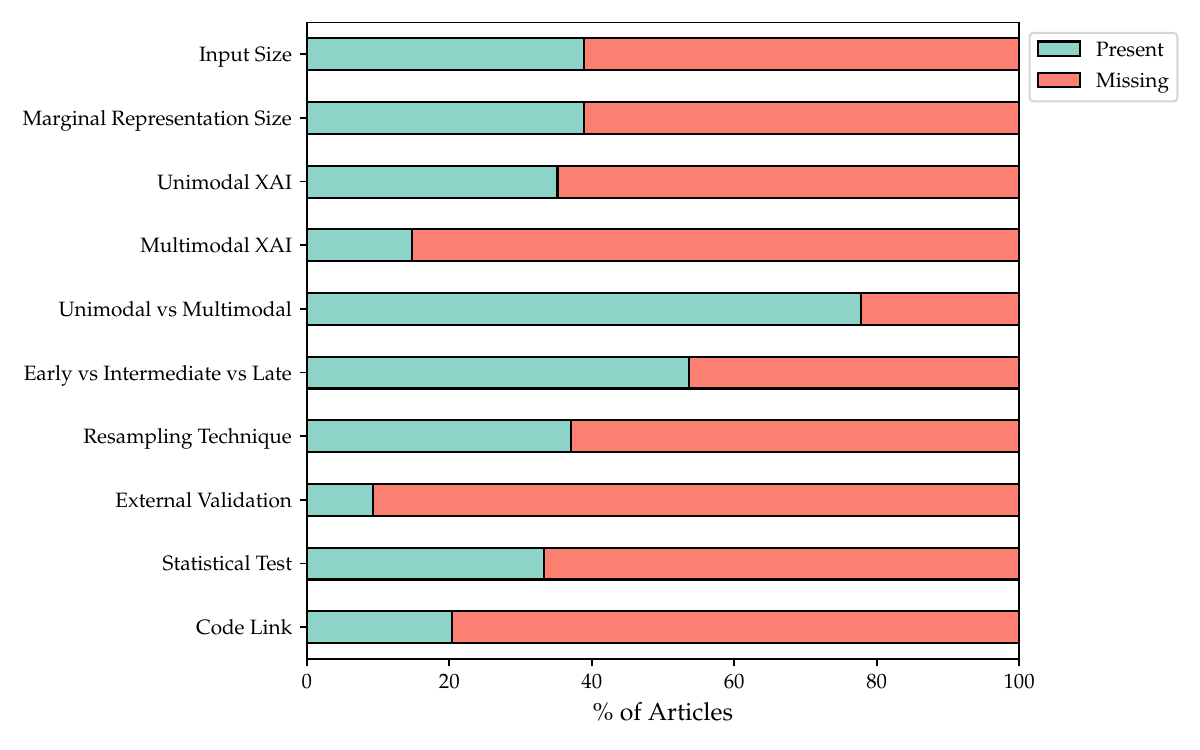}
\caption{Distribution of Reported Information in Analyzed Articles. Presents stacked bar charts for various fields, indicating the percentage of information explicitly stated in the articles (green) versus the information missing or not stated (red).}
\label{fig:present_vs_missing}
\end{figure}

While conducting the analyses, we identified significant gaps in the reporting of essential information, leading to reduced methodological transparency in the field of MDL, as shown in \figurename~\ref{fig:present_vs_missing}.
These gaps are not trivial; they are central to ensuring reproducibility, understanding model behavior, and validating the robustness of the proposed solutions.
Regarding the modalities' information, the input size (Section~\ref{sec:modalities}) and the marginal representation size (Section~\ref{sec:condensation}) are insufficiently reported. Without clarity on input dimensions and the size of representations post unimodal processing, it becomes challenging to grasp the data's distribution and the extent of feature extraction prior to fusion.
Regarding XAI (Section~\ref{sec:xai}), the lack of reporting on unimodal XAI and even more so on multimodal XAI limits the understanding of the multimodal models.
In particular, explainability is fundamental in biomedical applications where understanding model decisions can impact clinical outcomes, essential for increasing trust and adoption of AI-based systems.
Regarding the comparisons (Section~\ref{subsec:experimental_configuration}), a minority of studies fail to compare unimodal and multimodal approaches, which is a relatively lower omission but still significant for establishing the value of multimodal strategies.
A more substantial gap is seen in the comparison of different fusion strategies (Section~\ref{subsec:experimental_configuration}), with studies not reporting comparisons between early, intermediate, and late fusion, which are pivotal for contextualizing the effectiveness of intermediate fusion.
In terms of validation (Section~\ref{subsec:experimental_configuration}), many studies do not use sampling techniques, and even more overlook external validation. Such omissions call into question the generalizability of the models.
Furthermore, statistical tests (Section~\ref{subsec:experimental_configuration}) to establish the significance of findings are omitted in many of cases, raising concerns about the statistical robustness of the reported outcomes.
Finally, regarding the absence of a public code link in publications significantly hinders the field's progression. The availability of code is essential for peer validation, understanding model intricacies, and fostering collaborative improvements.
The concerning levels of missing information in these critical areas suggest a need for more rigorous reporting standards in MDL research, and also highlights an opportunity for future work to address these gaps, enhancing the quality and transparency of research within the domain.

\subsubsection{Potential Research Gaps}

Moving forward, we hereby outline the primary research gaps identified:
\begin{itemize}
    \item \textbf{Limited modalities combination}: As discussed in Section~\ref{sec:modalities_combination}, most studies utilizing intermediate fusion techniques in MDL within the medical field tend to focus on integrating only two types of modalities, typically imaging and tabular data~\cite{bib:caruso2022multimodal}. More complex combinations that involve integrating more than two modalities, including text, video, time series, and audio, are relatively uncommon. This limited bimodal approach, centered on imaging and tabular modalities, may be due to the complexity and challenges associated with merging data from diverse modalities, as well as the need for effective data structuring and preprocessing.    
    However, a more comprehensive multimodal analysis could offer a more detailed view of the patient, enhancing the performance of the models.
    \item \textbf{Lack of Multi-scale Data}: In the literature on MDL using intermediate fusion techniques within the biomedical field, there is a notable gap in integrating data of different dimensional scales. Combining macroscopic data (e.g., radiological images) with microscopic data (e.g., histopathology) can offer a more comprehensive understanding of the health status of patients. Such integration would enable the direct correlation of observable macroscopic changes with underlying molecular alterations, thereby improving the understanding of various pathologies.
    \item \textbf{Lack of benchmark datasets}: As shown in Section~\ref{sec:modalities}, a notable research gap in the field of intermediate MDL for biomedical applications is the absence of standardized, publicly available benchmark datasets. Similar to the role of ImageNet~\cite{bib:deng2009imagenet} in evaluating novel models for computer vision classification tasks, the availability of such datasets would allow researchers to compare and validate their deep learning models across different biomedical domains.
    \item \textbf{Suboptimal marginal representation dimensions}: A significant research gap exists regarding the dimensionality of marginal representations. Our analysis reveals no clear trend in how these dimensions are chosen, with many studies arbitrarily defaulting to powers of two, likely influenced by binary representations. More critically, as mentioned in Section~\ref{sec:condensation} the majority of studies employ equal-dimensional marginal representations across modalities, an approach that often lacks justification and may be suboptimal, considering that different modalities inherently contain varying amounts and types of information. The lack of a systematic approach to determining appropriate dimensions for each modality potentially leads to suboptimal fusion performance, loss of critical modality-specific information, and inefficient use of computational resources.
    \item \textbf{Insufficient consideration of multimodal fusion strategies}: As shown in Section~\ref{sec:fusion_module}, our analysis reveals that there is a persistent tendency in the field to prioritize unimodal feature extraction approaches over addressing the complexities of multimodal fusion. 
    This bias has led to the development of oversimplified architectures for intermediate fusion, leaving critical aspects unexplored, as shown in section~\ref{sec:how_fusion} where we analyzed the fusion operations performed in the fusion module.
    Key challenges remain unresolved, such as determining optimal methods for combining network branches that process different modalities, managing varying input sizes and dimensionalities, and ensuring efficient information flow across the fused network.
    Furthermore, our investigation indicates that the majority of fusion operations are performed without strong theoretical foundations or assumptions regarding the relationships between modality features. 
    This lack of a consistent approach results in methodologies that do not take full advantage of the complementarity and correlation of the characteristics of the different modalities.
    Consequently, the potential for integrating information from different sources remains largely unrealized, limiting the effectiveness of current multimodal fusion strategies in biomedical applications.
    Addressing these limitations requires a paradigm shift in research focus, emphasizing the development of more sophisticated, theoretically grounded fusion techniques that can effectively capture and utilize inter-modal relationships.
    \item \textbf{Limited Multimodal Module exploration}: The multimodal module, responsible for the final elaboration of fused information from the fusion module and its mapping to the task representation space, is a critical component in intermediate MDL. Our analysis, as detailed in section~\ref{sec:multimodal_module}, reveals a significant research gap in understanding the most effective architectures for extracting valuable task-specific representations from fused features.
    The vast majority of studies rely heavily on FCNN architectures for this crucial stage. While FCNNs offer different advantages, this predominant focus overlooks the potential benefits of more advanced approaches that could optimize the final learning elaboration from fused feature representations. 
    There is a notable scarcity of research exploring alternative architectures such as RNNs, CNNs, attention mechanisms, and graph-based methods in this context.
    This limitation is particularly concerning given the diverse nature of biomedical tasks and datasets. Different architectural approaches may offer unique advantages in capturing complex temporal dependencies (RNNs), spatial patterns (CNNs), feature importance (attention mechanisms), or relational structures (graph-based methods) within the fused representations. The lack of comprehensive comparative studies exploring these varied approaches hinders our ability to determine optimal architectures for specific biomedical applications.
    \item \textbf{Limited interpretability and explainability}: Deep learning models in biomedical applications often lack interpretability, making it challenging to understand and explain the decisions made by the models~\cite{bib:guarrasi2024multimodal}.
    As shown in Section~\ref{sec:xai}, our analysis shows that XAI is still an understudied topic in most of the current state-of-the-art of biomedical multimodal intermediate fusion approaches. This oversight highlights significant research gaps: there is a lack of standardized methods for evaluating model interpretability, inadequate methods to enhance the transparency of complex models, and insufficient integration of XAI methods tailored to multimodal data. Bridging these gaps is essential to ensure AI models are both accurate and interpretable, fulfilling the requirements for deployment in safety-critical applications such as healthcare.
    \item \textbf{Transferability and generalization across biomedical domains}: Many existing studies in intermediate MDL focus on specific biomedical domains or tasks, which limits the transferability and generalization of the models developed. 
    There exists a notable research gap exploring strategies that facilitate transfer learning and domain adaptation to enhance the generalization capabilities of these models across various biomedical fields. 
    Moreover, investigating the transferability of learned representations and developing techniques for detecting and mitigating domain shifts are critical areas that remain underexplored.
    \item \textbf{Robustness to missing modalities}: The ability to handle missing modalities is paramount for the deployment of MDL models in real-world scenarios, especially in such a complex and heterogeneous field as healthcare. As shown in Section~\ref{sec:missing}, our analysis reveals that most of the current biomedical multimodal intermediate fusion approaches are still inadequate to be deployed in real-world medical applications~\cite{bib:rofena2024deep}. This is evidenced by the prevalence of simple concatenation-based fusion strategies, which fail to make models robust to missing modalities. Consequently, most existing models lack inherent robustness to missing modalities during both training and inference, significantly hindering their applicability in real-world healthcare settings.
    \item \textbf{Real-time and resource-constrained implementation}: Despite the promise of intermediate MDL in biomedical applications, the real-time implementation and deployment of these models on resource-constrained devices or environments is an ongoing research challenge. Future research could investigate techniques that enable efficient computation, model compression, and optimization to accommodate real-time requirements and resource limitations. Exploring hardware acceleration techniques or designing specialized architectures suitable for intermediate MDL in resource-constrained settings could help bridge this research gap.
    \item \textbf{Ethical considerations and potential biases}: As intermediate MDL models become more pervasive in biomedical applications, it is crucial to address ethical considerations and potential biases that may arise from these models. Research could aim to investigate the potential biases, fairness, and transparency issues associated with the deployment of these models, with a focus on ensuring patient privacy, data anonymization, and reducing biases related to race, gender, or socioeconomic factors. Designing frameworks that enable the identification and mitigation of biases within intermediate MDL models would contribute towards ensuring the ethical use of these technologies in biomedical applications.
\end{itemize}

While there is a clear path forward, marked by research gaps and opportunities for enhancement, the potential for these sophisticated computational techniques to aid patient care and biomedical research remains beyond question.
It is with a collective effort toward addressing these gaps, through rigorous research, transparent reporting, and collaborative innovation, that we can harness the full potential of MDL.
As we stand on the peak of these advancements, the convergence of technology, medicine, and data science prospects a future where healthcare is more precise, personalized, and preventive.

\subsubsection{Future Research}

The exploration of intermediate fusion methods in MDL has uncovered numerous avenues for future research, especially pertinent to advancing the capabilities and applications within the biomedical sphere. These potential directions are critical for progressing the field and addressing the identified limitations that currently challenge the efficacy and applicability of these methods.

\begin{itemize}
    \item \textbf{Multimodal dimension and modalities combination}: Future research should prioritize investigating trimodal and high-modal combinations, as well as incorporating diverse modalities like time series, text, audio, and video. Our analysis has shown that most current studies are bimodal, predominantly utilizing imaging and tabular data.  
    Exploring a wider range of data types and multimodal dimensions could be essential for gaining a more comprehensive understanding of clinical scenarios, thereby enhancing diagnostic and predictive accuracy.
    \item \textbf{Integration of Multi-scale Data}: One area of future research could be focused on integrating multi-scale data from different modalities for better analysis and understanding in biomedical applications. This could involve developing deep learning architectures that can effectively handle and fuse data at different spatial and temporal scales, such as combining microscopic and macroscopic data in cancer research.
    \item \textbf{Creation of benchmark datasets}: Future research should focus on the creation and curation of large-scale, high-quality benchmark datasets that encompass multiple biomedical domains, ensuring a wide range of data modalities, such as imaging, genomics, clinical data, and others.
    These datasets should be annotated with standardized labels and metadata to support various deep learning tasks, enabling consistent evaluation and comparison of deep learning models.    
    \item \textbf{Optimization of marginal representation dimensions}: Future research should investigate the optimal dimensionality of marginal representations in intermediate fusion. Current research shows no clear trend in how these dimensions are chosen which indicates that the selection might be suboptimal. This presents a significant research opportunity to explore the impact of varying dimensionalities on fusion performance; methods for determining optimal dimensions for each modality, taking into account the inherent complexity and information density of different data types; and strategies for balancing the representation of modalities with significantly different feature spaces without losing critical information. Addressing this gap could lead to more efficient and effective multimodal fusion models that better capture the unique characteristics of each modality.
    \item \textbf{Deepening Multimodal Architectures}: Future research in intermediate MDL for biomedical applications should prioritize the development of innovative and integrated fusion strategies. 
    By exploring novel approaches to combine information from diverse sources, taking also into account the data heterogeneity in the biomedical field, more advanced techniques to harmoniously exploit the most informative features could be obtained.
    Moreover, the actual research lacks of advanced techniques able to map the multimodal features into the final task output, limiting the effective capability of the model to elaborate the multimodal nature of the input data.
    This paradigm shift in research focus should emphasize creating more sophisticated, theoretically grounded fusion techniques capable of effectively capturing and leveraging inter-modal relationships. 
    Such advancements will be critical in maximizing the impact of multimodal analysis on diagnostic accuracy, treatment planning, and overall patient care in the biomedical field.
    \item \textbf{Explainability and Interpretability}: Deep learning models often lack interpretability, a crucial aspect in biomedical applications where decision-making is critical. Future research should aim to improve the explainability of these models in biomedical contexts. This involves creating methods that provide clear interpretations or explanations of the decisions made by MDL models, enabling clinicians and researchers to understand and trust their predictions. To achieve this, there is plenty of room for advancement. Future research should prioritize integrating XAI principles starting from the initial design phase, favoring inherently explainable architectures.  Moreover, developing standardized evaluation metrics for model interpretability, and enhancing user interfaces to better communicate model decisions can further advance this field. To this end, a joint effort between researchers and clinicians will also be crucial to ensure that the developed solutions are both technically robust and practically applicable in real-world biomedical settings.
    \item \textbf{Transfer Learning and Domain Adaptation}: Biomedical datasets are often limited in size and differ in distribution across different hospitals or research centers. Future research should focus on investigating transfer learning and domain adaptation techniques in MDL to overcome the challenge of limited data and domain shift. 
    By exploring these techniques,  models trained on one dataset or domain could be effectively applied to other datasets or domains, thereby overcoming data scarcity and improving generalizability.  This research direction has the potential to improve the robustness, applicability, and resilience of multimodal deep-learning models in biomedical applications.
    \item \textbf{Privacy and Ethical Considerations}: In future research on MDL for biomedical applications, it's essential to address both privacy and ethical considerations, especially under the upcoming AI Act. This involves developing and implementing robust privacy-preserving techniques that ensure data confidentiality and adhere to informed consent principles. Moreover, it's critical to establish clear guidelines for handling and sharing biomedical data. These measures will not only comply with legal frameworks like the AI Act but also foster trust and safety in the use of AI technologies in healthcare contexts.
    \item \textbf{Real-time Applications and Clinical Decision Support}: Investigating real-time applications of MDL in clinical decision support systems is a promising area of future research. This could involve developing models that can process and analyze multimodal data streams in real-time, providing timely insights or recommendations to healthcare professionals for accurate diagnosis or treatment planning.
    \item \textbf{Federated Learning}: As healthcare data grows not only in volume but also in sensitivity, maintaining privacy and data security becomes crucial. Federated learning presents a promising solution by enabling the training of models on decentralized datasets without needing to share the data itself. Future research could explore how federated learning can be applied to intermediate fusion in MDL, particularly in biomedical applications. Investigating this approach would address privacy concerns, particularly with patient data, and could enhance model training by leveraging diverse datasets from multiple institutions without compromising data security.
    \item \textbf{Multimodal Medical Foundation Models}: 
    Advances in MDL have made great strides in improving diagnosis, prognosis, and treatment decisions. However, current methodologies remain largely task-specific (narrow AIs), resulting in poor generalization and lacking broad medical understanding and contextual information. In contrast, the recent rise of foundation AI models offers a chance to rethink biomedical AI systems, enriching them with more general capabilities. Foundation models have gained popularity in language, and their scope has gradually expanded to other domains, such as computer vision. These models are often trained with self-supervision and contrastive learning approaches, fully exploiting the informativeness of large-scale unlabeled data. This enhances scalability but also compels the models to predict parts of the inputs, making them richer and potentially more useful than models trained on a limited labeled space. Foundation models can be effectively adapted to several downstream tasks and settings using in-context learning or few-shot fine-tuning, with the same set of model weights.
    Despite their promise, foundation models in healthcare are still in an early stage, facing several limitations such as insufficient domain-specific training data, challenges in handling the diverse and complex nature of medical information, and the need for rigorous validation in clinical settings.
    Given the potential of foundation models, it is imperative to investigate their multimodal capabilities, leading to the opportunity to define a unified medical AI system that can interpret complex multimodal data. The new paradigm offered by foundation models in conjunction with multimodal learning expertise leads to the discovery of inter/cross-modality information but also potential knowledge intersection between seemingly different tasks and data. Moreover, there will be opportunities to explore the pivotal role of text as the ``inter-lingua''  of all modalities~\cite{poon2023multimodal}. This will result in a holistic orchestration between data modalities and tasks that equips the AI system with robust biomedical knowledge, to be deployed in real clinical applications.
\end{itemize}

\subsection{Implications for Practice}

Our review sheds light on specific practical implications that can significantly advance research and application in biomedical fields through multimodal AI. Here are detailed benefits and impacts:
\begin{itemize}
    \item \textbf{Enhanced Diagnostic Accuracy}: By integrating diverse data modalities, through intermediate fusion, diagnostic models become more precise. This approach captures a comprehensive view of patient health, enabling earlier and more accurate detection of conditions that are crucial for effective treatment strategies.
    \item \textbf{Personalized Patient Care}: The deep insights provided by intermediate fusion into complex patient data are vital for personalizing treatment plans. Such data-driven personalization ensures treatments are specifically tailored to individual patient profiles, thereby enhancing the effectiveness of medical interventions.
    \item \textbf{Efficient Data Utilization}: With the ever-increasing volumes of data in healthcare, intermediate fusion methods stand out by enabling the synthesis of heterogeneous data streams into actionable insights, thus preventing data overload and improving the efficiency of medical data analysis.
    \item \textbf{Advancement in Research}: Applying intermediate fusion techniques can lead to significant breakthroughs in biomedical research. It enables the exploration of complex, multifaceted medical conditions that might remain opaque under unimodal analysis, thus accelerating the development of novel therapeutics and diagnostics.
    \item \textbf{Education and Training}: The integration of intermediate fusion methods into educational curricula for healthcare professionals highlights the shift towards data-centric medical practices. This training equips medical personnel with the skills needed to apply cutting-edge technology in clinical diagnostics and treatment planning.
    \item \textbf{Resource Optimization}: Improvements in diagnostic and treatment accuracy reduce unnecessary procedures and optimize the allocation of medical resources. This not only cuts costs but also improves patient care by focusing on the most effective medical interventions.
    \item \textbf{Ethical and Privacy Considerations}: As multimodal fusion becomes more embedded in clinical practice, it is imperative to navigate the ethical and privacy aspects diligently. The secure and ethical handling of patient data ensures trust and compliance with regulatory standards, which is foundational in healthcare.
\end{itemize}

By highlighting these aspects, this review emphasizes the critical role of intermediate fusion in transforming the biomedical landscape. The adoption of these advanced computational techniques paves the way for more effective, personalized, and data-driven approaches in healthcare, positioning the field to leverage the full potential of AI in improving patient outcomes.

\section*{Author Contributions}
\textbf{Valerio Guarrasi}:
Conceptualization,
Data curation,
Formal analysis,
Funding acquisition,
Investigation,
Methodology,
Project administration,
Resources,
Software,
Supervision,
Validation,
Visualization,
Writing – original draft,
Writing – review and editing;
\textbf{Fatih Aksu}:
Conceptualization,
Data curation,
Formal analysis,
Investigation,
Methodology,
Software,
Validation,
Visualization,
Writing – original draft,
Writing – review and editing;
\textbf{Camillo Maria Caruso}:
Conceptualization,
Data curation,
Formal analysis,
Investigation,
Methodology,
Software,
Validation,
Visualization,
Writing – original draft,
Writing – review and editing;
\textbf{Francesco Di Feola}:
Conceptualization,
Data curation,
Formal analysis,
Investigation,
Methodology,
Software,
Validation,
Visualization,
Writing – original draft,
Writing – review and editing;
\textbf{Aurora Rofena}:
Conceptualization,
Data curation,
Formal analysis,
Investigation,
Methodology,
Software,
Validation,
Visualization,
Writing – original draft,
Writing – review and editing;
\textbf{Filippo Ruffini}:
Conceptualization,
Data curation,
Formal analysis,
Investigation,
Methodology,
Software,
Validation,
Visualization,
Writing – original draft,
Writing – review and editing;
\textbf{Paolo Soda}:
Conceptualization,
Funding acquisition,
Methodology,
Project administration,
Resources,
Supervision,
Writing – review and editing;

\section*{Acknowledgment}
Fatih Aksu, Camillo Maria Caruso, Aurora Rofena and Filippo Ruffini are Ph.D. students enrolled in the National Ph.D. in Artificial Intelligence, course on Health and Life Sciences, organized by Università Campus Bio-Medico di Roma.

This work was partially founded by: 
i) Università Campus Bio-Medico di Roma under the program ``University Strategic Projects'' within the project ``AI-powered Digital Twin for next-generation lung cancEr cAre (IDEA)''; 
ii) from PRIN 2022 MUR 20228MZFAA-AIDA (CUP C53D23003620008); 
iii) from PNRR MUR project PE0000013-FAIR.

\bibliographystyle{elsarticle-num-names} 
\bibliography{mybibfile}

\begin{thebibliography}{117}
\expandafter\ifx\csname natexlab\endcsname\relax\def\natexlab#1{#1}\fi
\providecommand{\url}[1]{\texttt{#1}}
\providecommand{\href}[2]{#2}
\providecommand{\path}[1]{#1}
\providecommand{\DOIprefix}{doi:}
\providecommand{\ArXivprefix}{arXiv:}
\providecommand{\URLprefix}{URL: }
\providecommand{\Pubmedprefix}{pmid:}
\providecommand{\doi}[1]{\href{http://dx.doi.org/#1}{\path{#1}}}
\providecommand{\Pubmed}[1]{\href{pmid:#1}{\path{#1}}}
\providecommand{\bibinfo}[2]{#2}
\ifx\xfnm\relax \def\xfnm[#1]{\unskip,\space#1}\fi
\bibitem[{LeCun et~al.(2015)LeCun, Bengio, and Hinton}]{bib:lecun2015deep}
\bibinfo{author}{Y.~LeCun}, \bibinfo{author}{Y.~Bengio}, \bibinfo{author}{G.~Hinton},
\newblock \bibinfo{title}{Deep learning},
\newblock \bibinfo{journal}{nature} \bibinfo{volume}{521} (\bibinfo{year}{2015}) \bibinfo{pages}{436--444}.
\bibitem[{Ramachandram and Taylor(2017)}]{bib:ramachandram2017deep}
\bibinfo{author}{D.~Ramachandram}, \bibinfo{author}{G.~W. Taylor},
\newblock \bibinfo{title}{{Deep multimodal learning: A survey on recent advances and trends}},
\newblock \bibinfo{journal}{IEEE signal processing magazine} \bibinfo{volume}{34} (\bibinfo{year}{2017}) \bibinfo{pages}{96--108}.
\bibitem[{Lahat et~al.(2015)Lahat, Adali, and Jutten}]{bib:lahat2015multimodal}
\bibinfo{author}{D.~Lahat}, \bibinfo{author}{T.~Adali}, \bibinfo{author}{C.~Jutten},
\newblock \bibinfo{title}{{Multimodal data fusion: an overview of methods, challenges, and prospects}},
\newblock \bibinfo{journal}{Proceedings of the IEEE} \bibinfo{volume}{103} (\bibinfo{year}{2015}) \bibinfo{pages}{1449--1477}.
\bibitem[{Sui et~al.(2012)Sui, Adali, Yu, Chen, and Calhoun}]{bib:sui2012review}
\bibinfo{author}{J.~Sui}, \bibinfo{author}{T.~Adali}, \bibinfo{author}{Q.~Yu}, \bibinfo{author}{J.~Chen}, \bibinfo{author}{V.~D. Calhoun},
\newblock \bibinfo{title}{{A review of multivariate methods for multimodal fusion of brain imaging data}},
\newblock \bibinfo{journal}{Journal of neuroscience methods} \bibinfo{volume}{204} (\bibinfo{year}{2012}) \bibinfo{pages}{68--81}.
\bibitem[{Viswanath et~al.(2017)Viswanath, Tiwari, Lee, Madabhushi, and Initiative}]{bib:viswanath2017dimensionality}
\bibinfo{author}{S.~E. Viswanath}, \bibinfo{author}{P.~Tiwari}, \bibinfo{author}{G.~Lee}, \bibinfo{author}{A.~Madabhushi}, \bibinfo{author}{A.~D.~N. Initiative},
\newblock \bibinfo{title}{{Dimensionality reduction-based fusion approaches for imaging and non-imaging biomedical data: concepts, workflow, and use-cases}},
\newblock \bibinfo{journal}{BMC medical imaging} \bibinfo{volume}{17} (\bibinfo{year}{2017}) \bibinfo{pages}{1--17}.
\bibitem[{Ching et~al.(2018)Ching, Himmelstein, Beaulieu-Jones, Kalinin, Do, Way, Ferrero, Agapow, Zietz, Hoffman et~al.}]{bib:ching2018opportunities}
\bibinfo{author}{T.~Ching}, \bibinfo{author}{D.~S. Himmelstein}, \bibinfo{author}{B.~K. Beaulieu-Jones}, \bibinfo{author}{A.~A. Kalinin}, \bibinfo{author}{B.~T. Do}, \bibinfo{author}{G.~P. Way}, \bibinfo{author}{E.~Ferrero}, \bibinfo{author}{P.-M. Agapow}, \bibinfo{author}{M.~Zietz}, \bibinfo{author}{M.~M. Hoffman}, et~al.,
\newblock \bibinfo{title}{{Opportunities and obstacles for deep learning in biology and medicine}},
\newblock \bibinfo{journal}{Journal of the royal society interface} \bibinfo{volume}{15} (\bibinfo{year}{2018}) \bibinfo{pages}{20170387}.
\bibitem[{Ahmad et~al.(2018)Ahmad, Eckert, and Teredesai}]{bib:ahmad2018interpretable}
\bibinfo{author}{M.~A. Ahmad}, \bibinfo{author}{C.~Eckert}, \bibinfo{author}{A.~Teredesai},
\newblock \bibinfo{title}{{Interpretable machine learning in healthcare}},
\newblock in: \bibinfo{booktitle}{Proceedings of the 2018 ACM international conference on bioinformatics, computational biology, and health informatics}, \bibinfo{year}{2018}, pp. \bibinfo{pages}{559--560}.
\bibitem[{Vayena et~al.(2018)Vayena, Blasimme, and Cohen}]{bib:vayena2018machine}
\bibinfo{author}{E.~Vayena}, \bibinfo{author}{A.~Blasimme}, \bibinfo{author}{I.~G. Cohen},
\newblock \bibinfo{title}{{Machine learning in medicine: addressing ethical challenges}},
\newblock \bibinfo{journal}{PLoS medicine} \bibinfo{volume}{15} (\bibinfo{year}{2018}) \bibinfo{pages}{e1002689}.
\bibitem[{Strubell et~al.(2020)Strubell, Ganesh, and McCallum}]{bib:strubell2019energy}
\bibinfo{author}{E.~Strubell}, \bibinfo{author}{A.~Ganesh}, \bibinfo{author}{A.~McCallum},
\newblock \bibinfo{title}{{Energy and policy considerations for modern deep learning research}},
\newblock in: \bibinfo{booktitle}{Proceedings of the AAAI conference on artificial intelligence}, volume~\bibinfo{volume}{34}, \bibinfo{year}{2020}, pp. \bibinfo{pages}{13693--13696}.
\bibitem[{Min et~al.(2017)Min, Lee, and Yoon}]{bib:min2017deep}
\bibinfo{author}{S.~Min}, \bibinfo{author}{B.~Lee}, \bibinfo{author}{S.~Yoon},
\newblock \bibinfo{title}{{Deep learning in bioinformatics}},
\newblock \bibinfo{journal}{Briefings in bioinformatics} \bibinfo{volume}{18} (\bibinfo{year}{2017}) \bibinfo{pages}{851--869}.
\bibitem[{Guo et~al.(2019)Guo, Wang, and Wang}]{bib:guo2019deep}
\bibinfo{author}{W.~Guo}, \bibinfo{author}{J.~Wang}, \bibinfo{author}{S.~Wang},
\newblock \bibinfo{title}{{Deep multimodal representation learning: A survey}},
\newblock \bibinfo{journal}{Ieee Access} \bibinfo{volume}{7} (\bibinfo{year}{2019}) \bibinfo{pages}{63373--63394}.
\bibitem[{Jabeen et~al.(2023)Jabeen, Li, Amin, Bourahla, Li, and Jabbar}]{bib:jabeen2023review}
\bibinfo{author}{S.~Jabeen}, \bibinfo{author}{X.~Li}, \bibinfo{author}{M.~S. Amin}, \bibinfo{author}{O.~Bourahla}, \bibinfo{author}{S.~Li}, \bibinfo{author}{A.~Jabbar},
\newblock \bibinfo{title}{{A review on methods and applications in multimodal deep learning}},
\newblock \bibinfo{journal}{ACM Transactions on Multimedia Computing, Communications and Applications} \bibinfo{volume}{19} (\bibinfo{year}{2023}) \bibinfo{pages}{1--41}.
\bibitem[{Gao et~al.(2020)Gao, Li, Chen, and Zhang}]{bib:gao2020survey}
\bibinfo{author}{J.~Gao}, \bibinfo{author}{P.~Li}, \bibinfo{author}{Z.~Chen}, \bibinfo{author}{J.~Zhang},
\newblock \bibinfo{title}{{A survey on deep learning for multimodal data fusion}},
\newblock \bibinfo{journal}{Neural Computation} \bibinfo{volume}{32} (\bibinfo{year}{2020}) \bibinfo{pages}{829--864}.
\bibitem[{Bayoudh et~al.(2022)Bayoudh, Knani, Hamdaoui, and Mtibaa}]{bib:bayoudh2022survey}
\bibinfo{author}{K.~Bayoudh}, \bibinfo{author}{R.~Knani}, \bibinfo{author}{F.~Hamdaoui}, \bibinfo{author}{A.~Mtibaa},
\newblock \bibinfo{title}{{A survey on deep multimodal learning for computer vision: advances, trends, applications, and datasets}},
\newblock \bibinfo{journal}{The Visual Computer} \bibinfo{volume}{38} (\bibinfo{year}{2022}) \bibinfo{pages}{2939--2970}.
\bibitem[{Liang et~al.(2024)Liang, Zadeh, and Morency}]{bib:liang2022foundations}
\bibinfo{author}{P.~P. Liang}, \bibinfo{author}{A.~Zadeh}, \bibinfo{author}{L.-P. Morency},
\newblock \bibinfo{title}{{Foundations \& trends in multimodal machine learning: Principles, challenges, and open questions}},
\newblock \bibinfo{journal}{ACM Computing Surveys} \bibinfo{volume}{56} (\bibinfo{year}{2024}) \bibinfo{pages}{1--42}.
\bibitem[{Ngiam et~al.(2011)Ngiam, Khosla, Kim, Nam, Lee, and Ng}]{bib:ngiam2011multimodal}
\bibinfo{author}{J.~Ngiam}, \bibinfo{author}{A.~Khosla}, \bibinfo{author}{M.~Kim}, \bibinfo{author}{J.~Nam}, \bibinfo{author}{H.~Lee}, \bibinfo{author}{A.~Y. Ng},
\newblock \bibinfo{title}{{Multimodal deep learning}},
\newblock in: \bibinfo{booktitle}{Proceedings of the 28th international conference on machine learning (ICML-11)}, \bibinfo{year}{2011}, pp. \bibinfo{pages}{689--696}.
\bibitem[{Zhang et~al.(2020)Zhang, Yang, He, and Deng}]{bib:zhang2020multimodal}
\bibinfo{author}{C.~Zhang}, \bibinfo{author}{Z.~Yang}, \bibinfo{author}{X.~He}, \bibinfo{author}{L.~Deng},
\newblock \bibinfo{title}{{Multimodal intelligence: Representation learning, information fusion, and applications}},
\newblock \bibinfo{journal}{IEEE Journal of Selected Topics in Signal Processing} \bibinfo{volume}{14} (\bibinfo{year}{2020}) \bibinfo{pages}{478--493}.
\bibitem[{Xu et~al.(2023)Xu, Zhu, and Clifton}]{bib:xu2023multimodal}
\bibinfo{author}{P.~Xu}, \bibinfo{author}{X.~Zhu}, \bibinfo{author}{D.~A. Clifton},
\newblock \bibinfo{title}{{Multimodal learning with transformers: A survey}},
\newblock \bibinfo{journal}{IEEE Transactions on Pattern Analysis and Machine Intelligence}  (\bibinfo{year}{2023}).
\bibitem[{Baltru{\v{s}}aitis et~al.(2018)Baltru{\v{s}}aitis, Ahuja, and Morency}]{bib:baltruvsaitis2018multimodal}
\bibinfo{author}{T.~Baltru{\v{s}}aitis}, \bibinfo{author}{C.~Ahuja}, \bibinfo{author}{L.-P. Morency},
\newblock \bibinfo{title}{{Multimodal machine learning: A survey and taxonomy}},
\newblock \bibinfo{journal}{IEEE transactions on pattern analysis and machine intelligence} \bibinfo{volume}{41} (\bibinfo{year}{2018}) \bibinfo{pages}{423--443}.
\bibitem[{Summaira et~al.(2021)Summaira, Li, Shoib, Li, and Abdul}]{bib:summaira2021recent}
\bibinfo{author}{J.~Summaira}, \bibinfo{author}{X.~Li}, \bibinfo{author}{A.~M. Shoib}, \bibinfo{author}{S.~Li}, \bibinfo{author}{J.~Abdul},
\newblock \bibinfo{title}{{Recent advances and trends in multimodal deep learning: A review}},
\newblock \bibinfo{journal}{arXiv preprint arXiv:2105.11087}  (\bibinfo{year}{2021}).
\bibitem[{Bengio et~al.(2013)Bengio, Courville, and Vincent}]{bib:bengio2013representation}
\bibinfo{author}{Y.~Bengio}, \bibinfo{author}{A.~Courville}, \bibinfo{author}{P.~Vincent},
\newblock \bibinfo{title}{{Representation learning: A review and new perspectives}},
\newblock \bibinfo{journal}{IEEE transactions on pattern analysis and machine intelligence} \bibinfo{volume}{35} (\bibinfo{year}{2013}) \bibinfo{pages}{1798--1828}.
\bibitem[{Zhu et~al.(2024)Zhu, Wu, Sebe, and Yan}]{bib:zhu2022vision+}
\bibinfo{author}{Y.~Zhu}, \bibinfo{author}{Y.~Wu}, \bibinfo{author}{N.~Sebe}, \bibinfo{author}{Y.~Yan},
\newblock \bibinfo{title}{{Vision+ x: A survey on multimodal learning in the light of data}},
\newblock \bibinfo{journal}{IEEE Transactions on Pattern Analysis and Machine Intelligence}  (\bibinfo{year}{2024}).
\bibitem[{Acosta et~al.(2022)Acosta, Falcone, Rajpurkar, and Topol}]{bib:acosta2022multimodal}
\bibinfo{author}{J.~N. Acosta}, \bibinfo{author}{G.~J. Falcone}, \bibinfo{author}{P.~Rajpurkar}, \bibinfo{author}{E.~J. Topol},
\newblock \bibinfo{title}{{Multimodal biomedical AI}},
\newblock \bibinfo{journal}{Nature Medicine} \bibinfo{volume}{28} (\bibinfo{year}{2022}) \bibinfo{pages}{1773--1784}.
\bibitem[{Stahlschmidt et~al.(2022)Stahlschmidt, Ulfenborg, and Synnergren}]{bib:stahlschmidt2022multimodal}
\bibinfo{author}{S.~R. Stahlschmidt}, \bibinfo{author}{B.~Ulfenborg}, \bibinfo{author}{J.~Synnergren},
\newblock \bibinfo{title}{{Multimodal deep learning for biomedical data fusion: a review}},
\newblock \bibinfo{journal}{Briefings in Bioinformatics} \bibinfo{volume}{23} (\bibinfo{year}{2022}) \bibinfo{pages}{bbab569}.
\bibitem[{Zhang et~al.(2020)Zhang, Dong, Wang, Yu, Yao, Zhou, Hu, Li, Jim{\'e}nez-Mesa, Ramirez et~al.}]{bib:zhang2020advances}
\bibinfo{author}{Y.-D. Zhang}, \bibinfo{author}{Z.~Dong}, \bibinfo{author}{S.-H. Wang}, \bibinfo{author}{X.~Yu}, \bibinfo{author}{X.~Yao}, \bibinfo{author}{Q.~Zhou}, \bibinfo{author}{H.~Hu}, \bibinfo{author}{M.~Li}, \bibinfo{author}{C.~Jim{\'e}nez-Mesa}, \bibinfo{author}{J.~Ramirez}, et~al.,
\newblock \bibinfo{title}{{Advances in multimodal data fusion in neuroimaging: Overview, challenges, and novel orientation}},
\newblock \bibinfo{journal}{Information Fusion} \bibinfo{volume}{64} (\bibinfo{year}{2020}) \bibinfo{pages}{149--187}.
\bibitem[{Heiliger et~al.(2023)Heiliger, Sekuboyina, Menze, Egger, and Kleesiek}]{bib:heiliger2023beyond}
\bibinfo{author}{L.~Heiliger}, \bibinfo{author}{A.~Sekuboyina}, \bibinfo{author}{B.~Menze}, \bibinfo{author}{J.~Egger}, \bibinfo{author}{J.~Kleesiek},
\newblock \bibinfo{title}{{Beyond medical imaging-A review of multimodal deep learning in radiology}},
\newblock \bibinfo{journal}{Authorea Preprints}  (\bibinfo{year}{2023}).
\bibitem[{Lipkova et~al.(2022)Lipkova, Chen, Chen, Lu, Barbieri, Shao, Vaidya, Chen, Zhuang, Williamson et~al.}]{bib:lipkova2022artificial}
\bibinfo{author}{J.~Lipkova}, \bibinfo{author}{R.~J. Chen}, \bibinfo{author}{B.~Chen}, \bibinfo{author}{M.~Y. Lu}, \bibinfo{author}{M.~Barbieri}, \bibinfo{author}{D.~Shao}, \bibinfo{author}{A.~J. Vaidya}, \bibinfo{author}{C.~Chen}, \bibinfo{author}{L.~Zhuang}, \bibinfo{author}{D.~F. Williamson}, et~al.,
\newblock \bibinfo{title}{{Artificial intelligence for multimodal data integration in oncology}},
\newblock \bibinfo{journal}{Cancer cell} \bibinfo{volume}{40} (\bibinfo{year}{2022}) \bibinfo{pages}{1095--1110}.
\bibitem[{Behrad and Abadeh(2022)}]{bib:behrad2022overview}
\bibinfo{author}{F.~Behrad}, \bibinfo{author}{M.~S. Abadeh},
\newblock \bibinfo{title}{{An overview of deep learning methods for multimodal medical data mining}},
\newblock \bibinfo{journal}{Expert Systems with Applications} \bibinfo{volume}{200} (\bibinfo{year}{2022}) \bibinfo{pages}{117006}.
\bibitem[{Zhou et~al.(2019)Zhou, Ruan, and Canu}]{bib:zhou2019review}
\bibinfo{author}{T.~Zhou}, \bibinfo{author}{S.~Ruan}, \bibinfo{author}{S.~Canu},
\newblock \bibinfo{title}{{A review: Deep learning for medical image segmentation using multi-modality fusion}},
\newblock \bibinfo{journal}{Array} \bibinfo{volume}{3} (\bibinfo{year}{2019}) \bibinfo{pages}{100004}.
\bibitem[{Xu(2019)}]{bib:xu2019deep}
\bibinfo{author}{Y.~Xu},
\newblock \bibinfo{title}{{Deep learning in multimodal medical image analysis}},
\newblock in: \bibinfo{booktitle}{Health Information Science: 8th International Conference, HIS 2019, Xi'an, China, October 18--20, 2019, Proceedings 8}, \bibinfo{organization}{Springer}, \bibinfo{year}{2019}, pp. \bibinfo{pages}{193--200}.
\bibitem[{Huang et~al.(2020)Huang, Pareek, Seyyedi, Banerjee, and Lungren}]{bib:huang2020fusion}
\bibinfo{author}{S.-C. Huang}, \bibinfo{author}{A.~Pareek}, \bibinfo{author}{S.~Seyyedi}, \bibinfo{author}{I.~Banerjee}, \bibinfo{author}{M.~P. Lungren},
\newblock \bibinfo{title}{{Fusion of medical imaging and electronic health records using deep learning: a systematic review and implementation guidelines}},
\newblock \bibinfo{journal}{NPJ digital medicine} \bibinfo{volume}{3} (\bibinfo{year}{2020}) \bibinfo{pages}{136}.
\bibitem[{Antonelli et~al.(2019)Antonelli, Guarracino, Maddalena, and Sangiovanni}]{bib:antonelli2019integrating}
\bibinfo{author}{L.~Antonelli}, \bibinfo{author}{M.~R. Guarracino}, \bibinfo{author}{L.~Maddalena}, \bibinfo{author}{M.~Sangiovanni},
\newblock \bibinfo{title}{{Integrating imaging and omics data: a review}},
\newblock \bibinfo{journal}{Biomedical Signal Processing and Control} \bibinfo{volume}{52} (\bibinfo{year}{2019}) \bibinfo{pages}{264--280}.
\bibitem[{Li et~al.(2022)Li, Huang, Xu, Cao, Lu, Wang, and Xiang}]{bib:05_li2022aviper}
\bibinfo{author}{X.~Li}, \bibinfo{author}{M.~Huang}, \bibinfo{author}{Y.~Xu}, \bibinfo{author}{Y.~Cao}, \bibinfo{author}{Y.~Lu}, \bibinfo{author}{P.~Wang}, \bibinfo{author}{X.~Xiang},
\newblock \bibinfo{title}{{AviPer: assisting visually impaired people to perceive the world with visual-tactile multimodal attention network}},
\newblock \bibinfo{journal}{CCF Transactions on Pervasive Computing and Interaction} \bibinfo{volume}{4} (\bibinfo{year}{2022}) \bibinfo{pages}{219--239}.
\bibitem[{Shetty et~al.(2023)Shetty, Mahale et~al.}]{bib:42_shetty2023multimodal}
\bibinfo{author}{S.~Shetty}, \bibinfo{author}{A.~Mahale}, et~al.,
\newblock \bibinfo{title}{{Multimodal medical tensor fusion network-based DL framework for abnormality prediction from the radiology CXRs and clinical text reports}},
\newblock \bibinfo{journal}{Multimedia Tools and Applications}  (\bibinfo{year}{2023}) \bibinfo{pages}{1--48}.
\bibitem[{Sun et~al.(2023)Sun, Guo, and Shen}]{bib:50_sun2023toward}
\bibinfo{author}{X.~Sun}, \bibinfo{author}{W.~Guo}, \bibinfo{author}{J.~Shen},
\newblock \bibinfo{title}{{Toward attention-based learning to predict the risk of brain degeneration with multimodal medical data}},
\newblock \bibinfo{journal}{Frontiers in Neuroscience} \bibinfo{volume}{16} (\bibinfo{year}{2023}) \bibinfo{pages}{1043626}.
\bibitem[{Aine et~al.(2017)Aine, Bockholt, Bustillo, Ca{\~n}ive, Caprihan, Gasparovic, Hanlon, Houck, Jung, Lauriello et~al.}]{cobre}
\bibinfo{author}{C.~Aine}, \bibinfo{author}{H.~J. Bockholt}, \bibinfo{author}{J.~R. Bustillo}, \bibinfo{author}{J.~M. Ca{\~n}ive}, \bibinfo{author}{A.~Caprihan}, \bibinfo{author}{C.~Gasparovic}, \bibinfo{author}{F.~M. Hanlon}, \bibinfo{author}{J.~M. Houck}, \bibinfo{author}{R.~E. Jung}, \bibinfo{author}{J.~Lauriello}, et~al.,
\newblock \bibinfo{title}{{Multimodal neuroimaging in schizophrenia: description and dissemination}},
\newblock \bibinfo{journal}{Neuroinformatics} \bibinfo{volume}{15} (\bibinfo{year}{2017}) \bibinfo{pages}{343--364}.
\bibitem[{Rahaman et~al.(2023)Rahaman, Chen, Fu, Lewis, Iraji, van Erp, and Calhoun}]{bib:14_rahaman2023deep}
\bibinfo{author}{M.~A. Rahaman}, \bibinfo{author}{J.~Chen}, \bibinfo{author}{Z.~Fu}, \bibinfo{author}{N.~Lewis}, \bibinfo{author}{A.~Iraji}, \bibinfo{author}{T.~G. van Erp}, \bibinfo{author}{V.~D. Calhoun},
\newblock \bibinfo{title}{{Deep multimodal predictome for studying mental disorders}},
\newblock \bibinfo{journal}{Human Brain Mapping} \bibinfo{volume}{44} (\bibinfo{year}{2023}) \bibinfo{pages}{509--522}.
\bibitem[{Rahaman et~al.(2021)Rahaman, Chen, Fu, Lewis, Iraji, and Calhoun}]{bib:31_rahaman2021multi}
\bibinfo{author}{M.~A. Rahaman}, \bibinfo{author}{J.~Chen}, \bibinfo{author}{Z.~Fu}, \bibinfo{author}{N.~Lewis}, \bibinfo{author}{A.~Iraji}, \bibinfo{author}{V.~D. Calhoun},
\newblock \bibinfo{title}{{Multi-modal deep learning of functional and structural neuroimaging and genomic data to predict mental illness}},
\newblock in: \bibinfo{booktitle}{2021 43rd Annual International Conference of the IEEE Engineering in Medicine \& Biology Society (EMBC)}, \bibinfo{organization}{IEEE}, \bibinfo{year}{2021}, pp. \bibinfo{pages}{3267--3272}.
\bibitem[{Rahaman et~al.(2022)Rahaman, Garg, Iraji, Fu, Chen, and Calhoun}]{bib:53_rahaman2022two}
\bibinfo{author}{M.~A. Rahaman}, \bibinfo{author}{Y.~Garg}, \bibinfo{author}{A.~Iraji}, \bibinfo{author}{Z.~Fu}, \bibinfo{author}{J.~Chen}, \bibinfo{author}{V.~Calhoun},
\newblock \bibinfo{title}{{Two-dimensional attentive fusion for multi-modal learning of neuroimaging and genomics data}},
\newblock in: \bibinfo{booktitle}{2022 IEEE 32nd International Workshop on Machine Learning for Signal Processing (MLSP)}, \bibinfo{organization}{IEEE}, \bibinfo{year}{2022}, pp. \bibinfo{pages}{1--6}.
\bibitem[{Liu et~al.(2022)Liu, Huang, Hu, Zhu, Wong, and Tan}]{bib:04_liu2022attention}
\bibinfo{author}{R.~Liu}, \bibinfo{author}{Z.-A. Huang}, \bibinfo{author}{Y.~Hu}, \bibinfo{author}{Z.~Zhu}, \bibinfo{author}{K.-C. Wong}, \bibinfo{author}{K.~C. Tan},
\newblock \bibinfo{title}{{Attention-like multimodality fusion with data augmentation for diagnosis of mental disorders using MRI}},
\newblock \bibinfo{journal}{IEEE Transactions on Neural Networks and Learning Systems}  (\bibinfo{year}{2022}).
\bibitem[{Johnson et~al.(2019)Johnson, Pollard, Greenbaum, Lungren, Deng, Peng, Lu, Mark, Berkowitz, and Horng}]{mimic_cxr}
\bibinfo{author}{A.~E. Johnson}, \bibinfo{author}{T.~J. Pollard}, \bibinfo{author}{N.~R. Greenbaum}, \bibinfo{author}{M.~P. Lungren}, \bibinfo{author}{C.-y. Deng}, \bibinfo{author}{Y.~Peng}, \bibinfo{author}{Z.~Lu}, \bibinfo{author}{R.~G. Mark}, \bibinfo{author}{S.~J. Berkowitz}, \bibinfo{author}{S.~Horng},
\newblock \bibinfo{title}{{MIMIC-CXR-JPG, a large publicly available database of labeled chest radiographs}},
\newblock \bibinfo{journal}{arXiv preprint arXiv:1901.07042}  (\bibinfo{year}{2019}).
\bibitem[{Lobantsev et~al.(2020)Lobantsev, Gusarova, Vatian, Kapitonov, and Shalyto}]{bib:06_lobantsev2020comparative}
\bibinfo{author}{A.~Lobantsev}, \bibinfo{author}{N.~Gusarova}, \bibinfo{author}{A.~S. Vatian}, \bibinfo{author}{A.~A. Kapitonov}, \bibinfo{author}{A.~A. Shalyto},
\newblock \bibinfo{title}{{Comparative assessment of text-image fusion models for medical diagnostics}},
\newblock \bibinfo{journal}{}  (\bibinfo{year}{2020}) \bibinfo{pages}{70--79}.
\bibitem[{Zeng et~al.(2022)Zeng, Niu, Lu, Pei, Zhou, and Guo}]{bib:26_zeng2022miftp}
\bibinfo{author}{J.~Zeng}, \bibinfo{author}{K.~Niu}, \bibinfo{author}{Y.~Lu}, \bibinfo{author}{S.~Pei}, \bibinfo{author}{Y.~Zhou}, \bibinfo{author}{Z.~Guo},
\newblock \bibinfo{title}{{MIFTP: A Multimodal Multi-Level Independent Fusion Framework with Improved Twin Pyramid for Multilabel Chest X-Ray Image Classification}},
\newblock in: \bibinfo{booktitle}{2022 IEEE 34th International Conference on Tools with Artificial Intelligence (ICTAI)}, \bibinfo{organization}{IEEE}, \bibinfo{year}{2022}, pp. \bibinfo{pages}{1112--1119}.
\bibitem[{Hayat et~al.(2022)Hayat, Geras, and Shamout}]{bib:25_hayat2022medfuse}
\bibinfo{author}{N.~Hayat}, \bibinfo{author}{K.~J. Geras}, \bibinfo{author}{F.~E. Shamout},
\newblock \bibinfo{title}{{MedFuse: Multi-modal fusion with clinical time-series data and chest X-ray images}},
\newblock in: \bibinfo{booktitle}{Machine Learning for Healthcare Conference}, \bibinfo{organization}{PMLR}, \bibinfo{year}{2022}, pp. \bibinfo{pages}{479--503}.
\bibitem[{Kohankhaki et~al.(2022)Kohankhaki, Ayad, Barhoush, Leibe, and Schmeink}]{bib:46_kohankhaki2022radiopaths}
\bibinfo{author}{M.~Kohankhaki}, \bibinfo{author}{A.~Ayad}, \bibinfo{author}{M.~Barhoush}, \bibinfo{author}{B.~Leibe}, \bibinfo{author}{A.~Schmeink},
\newblock \bibinfo{title}{{Radiopaths: Deep Multimodal Analysis on Chest Radiographs}},
\newblock in: \bibinfo{booktitle}{2022 IEEE International Conference on Big Data (Big Data)}, \bibinfo{organization}{IEEE}, \bibinfo{year}{2022}, pp. \bibinfo{pages}{3613--3621}.
\bibitem[{Jack~Jr et~al.(2008)Jack~Jr, Bernstein, Fox, Thompson, Alexander, Harvey, Borowski, Britson, L.~Whitwell, Ward et~al.}]{adni}
\bibinfo{author}{C.~R. Jack~Jr}, \bibinfo{author}{M.~A. Bernstein}, \bibinfo{author}{N.~C. Fox}, \bibinfo{author}{P.~Thompson}, \bibinfo{author}{G.~Alexander}, \bibinfo{author}{D.~Harvey}, \bibinfo{author}{B.~Borowski}, \bibinfo{author}{P.~J. Britson}, \bibinfo{author}{J.~L.~Whitwell}, \bibinfo{author}{C.~Ward}, et~al.,
\newblock \bibinfo{title}{{The Alzheimer's disease neuroimaging initiative (ADNI): MRI methods}},
\newblock \bibinfo{journal}{Journal of Magnetic Resonance Imaging: An Official Journal of the International Society for Magnetic Resonance in Medicine} \bibinfo{volume}{27} (\bibinfo{year}{2008}) \bibinfo{pages}{685--691}.
\bibitem[{Ostertag et~al.(2023)Ostertag, Visani, Urruty, and Beurton-Aimar}]{bib:24_ostertag2023long}
\bibinfo{author}{C.~Ostertag}, \bibinfo{author}{M.~Visani}, \bibinfo{author}{T.~Urruty}, \bibinfo{author}{M.~Beurton-Aimar},
\newblock \bibinfo{title}{{Long-term cognitive decline prediction based on multi-modal data using Multimodal3DSiameseNet: transfer learning from Alzheimer’s disease to Parkinson’s disease}},
\newblock \bibinfo{journal}{International Journal of Computer Assisted Radiology and Surgery} \bibinfo{volume}{18} (\bibinfo{year}{2023}) \bibinfo{pages}{809--818}.
\bibitem[{Ostertag et~al.(2020)Ostertag, Beurton-Aimar, Visani, Urruty, and Bertet}]{bib:43_ostertag2020predicting}
\bibinfo{author}{C.~Ostertag}, \bibinfo{author}{M.~Beurton-Aimar}, \bibinfo{author}{M.~Visani}, \bibinfo{author}{T.~Urruty}, \bibinfo{author}{K.~Bertet},
\newblock \bibinfo{title}{{Predicting brain degeneration with a multimodal Siamese neural network}},
\newblock in: \bibinfo{booktitle}{2020 Tenth International Conference on Image Processing Theory, Tools and Applications (IPTA)}, \bibinfo{organization}{IEEE}, \bibinfo{year}{2020}, pp. \bibinfo{pages}{1--6}.
\bibitem[{Subramanian et~al.(2016)Subramanian, Wache, Abadi, Vieriu, Winkler, and Sebe}]{ascertain}
\bibinfo{author}{R.~Subramanian}, \bibinfo{author}{J.~Wache}, \bibinfo{author}{M.~K. Abadi}, \bibinfo{author}{R.~L. Vieriu}, \bibinfo{author}{S.~Winkler}, \bibinfo{author}{N.~Sebe},
\newblock \bibinfo{title}{{ASCERTAIN: Emotion and personality recognition using commercial sensors}},
\newblock \bibinfo{journal}{IEEE Transactions on Affective Computing} \bibinfo{volume}{9} (\bibinfo{year}{2016}) \bibinfo{pages}{147--160}.
\bibitem[{Radhika and Oruganti(2021)}]{bib:13_radhika2021deep}
\bibinfo{author}{K.~Radhika}, \bibinfo{author}{V.~R.~M. Oruganti},
\newblock \bibinfo{title}{{Deep multimodal fusion for subject-independent stress detection}},
\newblock in: \bibinfo{booktitle}{2021 11th International Conference on Cloud Computing, Data Science \& Engineering (Confluence)}, \bibinfo{organization}{IEEE}, \bibinfo{year}{2021}, pp. \bibinfo{pages}{105--109}.
\bibitem[{Kuttala et~al.(2023)Kuttala, Subramanian, and Oruganti}]{bib:40_kuttala2023multimodal}
\bibinfo{author}{R.~Kuttala}, \bibinfo{author}{R.~Subramanian}, \bibinfo{author}{V.~R.~M. Oruganti},
\newblock \bibinfo{title}{{Multimodal Hierarchical CNN Feature Fusion for Stress Detection}},
\newblock \bibinfo{journal}{IEEE Access} \bibinfo{volume}{11} (\bibinfo{year}{2023}) \bibinfo{pages}{6867--6878}.
\bibitem[{Radhika and Oruganti(2021)}]{bib:48_radhika2021stress}
\bibinfo{author}{K.~Radhika}, \bibinfo{author}{V.~R.~M. Oruganti},
\newblock \bibinfo{title}{{Stress detection using CNN fusion}},
\newblock in: \bibinfo{booktitle}{TENCON 2021-2021 IEEE Region 10 Conference (TENCON)}, \bibinfo{organization}{IEEE}, \bibinfo{year}{2021}, pp. \bibinfo{pages}{492--497}.
\bibitem[{Markova et~al.(2019)Markova, Ganchev, and Kalinkov}]{clas}
\bibinfo{author}{V.~Markova}, \bibinfo{author}{T.~Ganchev}, \bibinfo{author}{K.~Kalinkov},
\newblock \bibinfo{title}{{Clas: A database for cognitive load, affect and stress recognition}},
\newblock in: \bibinfo{booktitle}{2019 International Conference on Biomedical Innovations and Applications (BIA)}, \bibinfo{organization}{IEEE}, \bibinfo{year}{2019}, pp. \bibinfo{pages}{1--4}.
\bibitem[{Keator et~al.(2016)Keator, van Erp, Turner, Glover, Mueller, Liu, Voyvodic, Rasmussen, Calhoun, Lee et~al.}]{fbirn}
\bibinfo{author}{D.~B. Keator}, \bibinfo{author}{T.~G. van Erp}, \bibinfo{author}{J.~A. Turner}, \bibinfo{author}{G.~H. Glover}, \bibinfo{author}{B.~A. Mueller}, \bibinfo{author}{T.~T. Liu}, \bibinfo{author}{J.~T. Voyvodic}, \bibinfo{author}{J.~Rasmussen}, \bibinfo{author}{V.~D. Calhoun}, \bibinfo{author}{H.~J. Lee}, et~al.,
\newblock \bibinfo{title}{{The function biomedical informatics research network data repository}},
\newblock \bibinfo{journal}{Neuroimage} \bibinfo{volume}{124} (\bibinfo{year}{2016}) \bibinfo{pages}{1074--1079}.
\bibitem[{Adhikari et~al.(2019)Adhikari, Hong, Sampath, Chiappelli, Jahanshad, Thompson, Rowland, Calhoun, Du, Chen et~al.}]{mrpc}
\bibinfo{author}{B.~M. Adhikari}, \bibinfo{author}{L.~E. Hong}, \bibinfo{author}{H.~Sampath}, \bibinfo{author}{J.~Chiappelli}, \bibinfo{author}{N.~Jahanshad}, \bibinfo{author}{P.~M. Thompson}, \bibinfo{author}{L.~M. Rowland}, \bibinfo{author}{V.~D. Calhoun}, \bibinfo{author}{X.~Du}, \bibinfo{author}{S.~Chen}, et~al.,
\newblock \bibinfo{title}{{Functional network connectivity impairments and core cognitive deficits in schizophrenia}},
\newblock \bibinfo{journal}{Human brain mapping} \bibinfo{volume}{40} (\bibinfo{year}{2019}) \bibinfo{pages}{4593--4605}.
\bibitem[{Bakr et~al.(2018)Bakr, Gevaert, Echegaray, Ayers, Zhou, Shafiq, Zheng, Benson, Zhang, Leung et~al.}]{nslc_radiogenomic}
\bibinfo{author}{S.~Bakr}, \bibinfo{author}{O.~Gevaert}, \bibinfo{author}{S.~Echegaray}, \bibinfo{author}{K.~Ayers}, \bibinfo{author}{M.~Zhou}, \bibinfo{author}{M.~Shafiq}, \bibinfo{author}{H.~Zheng}, \bibinfo{author}{J.~A. Benson}, \bibinfo{author}{W.~Zhang}, \bibinfo{author}{A.~N. Leung}, et~al.,
\newblock \bibinfo{title}{{A radiogenomic dataset of non-small cell lung cancer}},
\newblock \bibinfo{journal}{Scientific data} \bibinfo{volume}{5} (\bibinfo{year}{2018}) \bibinfo{pages}{1--9}.
\bibitem[{Hou et~al.(2023)Hou, Jia, Zhang, Wu, Zhao, Zhao, Wang, and Qiang}]{bib:09_hou2023deep}
\bibinfo{author}{G.~Hou}, \bibinfo{author}{L.~Jia}, \bibinfo{author}{Y.~Zhang}, \bibinfo{author}{W.~Wu}, \bibinfo{author}{L.~Zhao}, \bibinfo{author}{J.~Zhao}, \bibinfo{author}{L.~Wang}, \bibinfo{author}{Y.~Qiang},
\newblock \bibinfo{title}{{Deep learning approach for predicting lymph node metastasis in non-small cell lung cancer by fusing image--gene data}},
\newblock \bibinfo{journal}{Engineering Applications of Artificial Intelligence} \bibinfo{volume}{122} (\bibinfo{year}{2023}) \bibinfo{pages}{106140}.
\bibitem[{Wang et~al.(2021)Wang, Subramanian, and Syeda-Mahmood}]{bib:28_wang2021modeling}
\bibinfo{author}{H.~Wang}, \bibinfo{author}{V.~Subramanian}, \bibinfo{author}{T.~Syeda-Mahmood},
\newblock \bibinfo{title}{{Modeling uncertainty in multi-modal fusion for lung cancer survival analysis}},
\newblock in: \bibinfo{booktitle}{2021 IEEE 18th international symposium on biomedical imaging (ISBI)}, \bibinfo{organization}{IEEE}, \bibinfo{year}{2021}, pp. \bibinfo{pages}{1169--1172}.
\bibitem[{Subramanian et~al.(2020)Subramanian, Do, and Syeda-Mahmood}]{bib:38_subramanian2020multimodal}
\bibinfo{author}{V.~Subramanian}, \bibinfo{author}{M.~N. Do}, \bibinfo{author}{T.~Syeda-Mahmood},
\newblock \bibinfo{title}{{Multimodal fusion of imaging and genomics for lung cancer recurrence prediction}},
\newblock in: \bibinfo{booktitle}{2020 IEEE 17th International Symposium on Biomedical Imaging (ISBI)}, \bibinfo{organization}{IEEE}, \bibinfo{year}{2020}, pp. \bibinfo{pages}{804--808}.
\bibitem[{cpt(2018)}]{cptac-pda}
\bibinfo{title}{The {Clinical} {Proteomic} {Tumor} {Analysis} {Consortium} {Pancreatic} {Ductal} {Adenocarcinoma} {Collection} ({CPTAC}-{PDA})}, \bibinfo{howpublished}{\url{https://www.cancerimagingarchive.net/collection/cptac-pda/##citations}}, \bibinfo{year}{2018}. \bibinfo{note}{[Online; accessed 2024-04-16]}.
\bibitem[{Menegotto et~al.(2021)Menegotto, Becker, and Cazella}]{bib:07_menegotto2021computer}
\bibinfo{author}{A.~B. Menegotto}, \bibinfo{author}{C.~D.~L. Becker}, \bibinfo{author}{S.~C. Cazella},
\newblock \bibinfo{title}{{Computer-aided diagnosis of hepatocellular carcinoma fusing imaging and structured health data}},
\newblock \bibinfo{journal}{Health Information Science and Systems} \bibinfo{volume}{9} (\bibinfo{year}{2021}) \bibinfo{pages}{20}.
\bibitem[{Menegotto et~al.(2020)Menegotto, Lopes~Becker, and Cazella}]{bib:08_menegotto2020computer}
\bibinfo{author}{A.~B. Menegotto}, \bibinfo{author}{C.~D. Lopes~Becker}, \bibinfo{author}{S.~C. Cazella},
\newblock \bibinfo{title}{{Computer-aided hepatocarcinoma diagnosis using multimodal deep learning}},
\newblock in: \bibinfo{booktitle}{Ambient Intelligence--Software and Applications--, 10th International Symposium on Ambient Intelligence}, \bibinfo{organization}{Springer}, \bibinfo{year}{2020}, pp. \bibinfo{pages}{3--10}.
\bibitem[{Johnson et~al.(2016)Johnson, Pollard, Shen, Lehman, Feng, Ghassemi, Moody, Szolovits, Celi, and Mark}]{mimic-iii}
\bibinfo{author}{A.~E. Johnson}, \bibinfo{author}{T.~J. Pollard}, \bibinfo{author}{L.~Shen}, \bibinfo{author}{L.-w.~H. Lehman}, \bibinfo{author}{M.~Feng}, \bibinfo{author}{M.~Ghassemi}, \bibinfo{author}{B.~Moody}, \bibinfo{author}{P.~Szolovits}, \bibinfo{author}{L.~A. Celi}, \bibinfo{author}{R.~G. Mark},
\newblock \bibinfo{title}{{Data descriptor: MIMIC-III, a freely accessible critical care database}},
\newblock \bibinfo{journal}{Scientific data} \bibinfo{volume}{3} (\bibinfo{year}{2016}) \bibinfo{pages}{1--9}.
\bibitem[{Niu et~al.(2023)Niu, Zhang, Peng, Pan, and Xiao}]{bib:12_niu2023deep}
\bibinfo{author}{K.~Niu}, \bibinfo{author}{K.~Zhang}, \bibinfo{author}{X.~Peng}, \bibinfo{author}{Y.~Pan}, \bibinfo{author}{N.~Xiao},
\newblock \bibinfo{title}{{Deep multi-modal intermediate fusion of clinical record and time series data in mortality prediction}},
\newblock \bibinfo{journal}{Frontiers in Molecular Biosciences} \bibinfo{volume}{10} (\bibinfo{year}{2023}) \bibinfo{pages}{1136071}.
\bibitem[{Ma et~al.(2023)Ma, Hao, Zhao, Luo, Liu, and Li}]{bib:44_ma2023predicting}
\bibinfo{author}{M.~Ma}, \bibinfo{author}{X.~Hao}, \bibinfo{author}{J.~Zhao}, \bibinfo{author}{S.~Luo}, \bibinfo{author}{Y.~Liu}, \bibinfo{author}{D.~Li},
\newblock \bibinfo{title}{{Predicting heart failure in-hospital mortality by integrating longitudinal and category data in electronic health records}},
\newblock \bibinfo{journal}{Medical \& Biological Engineering \& Computing}  (\bibinfo{year}{2023}) \bibinfo{pages}{1--17}.
\bibitem[{Johnson et~al.(2023)Johnson, Johnson, Bulgarelli, Pollard, Horng, Celi, and Mark}]{mimc-iv}
\bibinfo{author}{A.~Johnson}, \bibinfo{author}{A.~Johnson}, \bibinfo{author}{L.~Bulgarelli}, \bibinfo{author}{T.~Pollard}, \bibinfo{author}{S.~Horng}, \bibinfo{author}{L.~A. Celi}, \bibinfo{author}{R.~Mark}, \bibinfo{title}{Mimic-{IV} (version 2.2)}, \bibinfo{howpublished}{https://physionet.org/content/mimiciv/2.2}, \bibinfo{year}{2023}. \bibinfo{note}{[Online; accessed 2024-04-16]}.
\bibitem[{Linehan et~al.(2016)Linehan, Gautam, Kirk, Lee, Roche, Bonaccio, Filippini, Rieger-Christ, Lemmerman, and Jarosz}]{tcga-kirp}
\bibinfo{author}{M.~Linehan}, \bibinfo{author}{R.~Gautam}, \bibinfo{author}{S.~Kirk}, \bibinfo{author}{Y.~Lee}, \bibinfo{author}{C.~Roche}, \bibinfo{author}{E.~Bonaccio}, \bibinfo{author}{J.~Filippini}, \bibinfo{author}{K.~Rieger-Christ}, \bibinfo{author}{J.~Lemmerman}, \bibinfo{author}{R.~Jarosz}, \bibinfo{title}{The {Cancer} {Genome} {Atlas} {Cervical} {Kidney} {Renal} {Papillary} {Cell} {Carcinoma} {Collection} ({TCGA}-{KIRP}) ({Version} 4)}, \bibinfo{howpublished}{https://www.cancerimagingarchive.net/collection/tcga-kirp/}, \bibinfo{year}{2016}. \bibinfo{note}{[Online; accessed 2024-04-16]}.
\bibitem[{Erickson et~al.(2016)Erickson, Kirk, Lee, Bathe, Kearns, Gerdes, Rieger-Christ, and Lemmerman}]{tcga-lihc}
\bibinfo{author}{B.~Erickson}, \bibinfo{author}{S.~Kirk}, \bibinfo{author}{Y.~Lee}, \bibinfo{author}{O.~Bathe}, \bibinfo{author}{M.~Kearns}, \bibinfo{author}{C.~Gerdes}, \bibinfo{author}{K.~Rieger-Christ}, \bibinfo{author}{J.~Lemmerman}, \bibinfo{title}{The {Cancer} {Genome} {Atlas} {Liver} {Hepatocellular} {Carcinoma} {Collection} ({TCGA}-{LIHC}) ({Version} 5)}, \bibinfo{howpublished}{https://www.cancerimagingarchive.net/collection/tcga-lihc/}, \bibinfo{year}{2016}. \bibinfo{note}{[Online; accessed 2024-04-16]}.
\bibitem[{Lucchesi and Aredes(2016)}]{tcga-stad}
\bibinfo{author}{F.~Lucchesi}, \bibinfo{author}{N.~Aredes}, \bibinfo{title}{The {Cancer} {Genome} {Atlas} {Stomach} {Adenocarcinoma} {Collection} ({TCGA}-{STAD}) ({Version} 3)}, \bibinfo{howpublished}{https://doi.org/10.7937/K9/TCIA.2016.GDHL9KIM}, \bibinfo{year}{2016}. \bibinfo{note}{[Online; accessed 2024-04-16]}.
\bibitem[{Yuan et~al.(2021)Yuan, Wu, and Zhang}]{gini}
\bibinfo{author}{Y.~Yuan}, \bibinfo{author}{L.~Wu}, \bibinfo{author}{X.~Zhang},
\newblock \bibinfo{title}{{Gini-impurity index analysis}},
\newblock \bibinfo{journal}{IEEE Transactions on Information Forensics and Security} \bibinfo{volume}{16} (\bibinfo{year}{2021}) \bibinfo{pages}{3154--3169}.
\bibitem[{Chawla et~al.(2002)Chawla, Bowyer, Hall, and Kegelmeyer}]{smote}
\bibinfo{author}{N.~V. Chawla}, \bibinfo{author}{K.~W. Bowyer}, \bibinfo{author}{L.~O. Hall}, \bibinfo{author}{W.~P. Kegelmeyer},
\newblock \bibinfo{title}{{SMOTE: synthetic minority over-sampling technique}},
\newblock \bibinfo{journal}{Journal of artificial intelligence research} \bibinfo{volume}{16} (\bibinfo{year}{2002}) \bibinfo{pages}{321--357}.
\bibitem[{Seo et~al.(2021)Seo, Kim, Han, Son, Hong, Sohn, Shim, and Hwang}]{bib:45_seo2021predicting}
\bibinfo{author}{S.~Seo}, \bibinfo{author}{Y.~Kim}, \bibinfo{author}{H.-J. Han}, \bibinfo{author}{W.~C. Son}, \bibinfo{author}{Z.-Y. Hong}, \bibinfo{author}{I.~Sohn}, \bibinfo{author}{J.~Shim}, \bibinfo{author}{C.~Hwang},
\newblock \bibinfo{title}{{Predicting successes and failures of clinical trials with outer product--based convolutional neural network}},
\newblock \bibinfo{journal}{Frontiers in Pharmacology} \bibinfo{volume}{12} (\bibinfo{year}{2021}) \bibinfo{pages}{670670}.
\bibitem[{Ren et~al.(2020)Ren, Yu, Ma, Zhao, Yi et~al.}]{ldsm_loss}
\bibinfo{author}{J.~Ren}, \bibinfo{author}{C.~Yu}, \bibinfo{author}{X.~Ma}, \bibinfo{author}{H.~Zhao}, \bibinfo{author}{S.~Yi}, et~al.,
\newblock \bibinfo{title}{{Balanced meta-softmax for long-tailed visual recognition}},
\newblock \bibinfo{journal}{Advances in neural information processing systems} \bibinfo{volume}{33} (\bibinfo{year}{2020}) \bibinfo{pages}{4175--4186}.
\bibitem[{Li et~al.(2022)Li, Long, Yang, Weng, Zeng, Huang, Wang, and Hao}]{bib:01_li2022bi}
\bibinfo{author}{Y.~Li}, \bibinfo{author}{S.~Long}, \bibinfo{author}{Z.~Yang}, \bibinfo{author}{H.~Weng}, \bibinfo{author}{K.~Zeng}, \bibinfo{author}{Z.~Huang}, \bibinfo{author}{F.~L. Wang}, \bibinfo{author}{T.~Hao},
\newblock \bibinfo{title}{{A Bi-level representation learning model for medical visual question answering}},
\newblock \bibinfo{journal}{Journal of Biomedical Informatics} \bibinfo{volume}{134} (\bibinfo{year}{2022}) \bibinfo{pages}{104183}.
\bibitem[{Cahan et~al.(2023)Cahan, Klang, Marom, Soffer, Barash, Burshtein, Konen, and Greenspan}]{bib:37_cahan2023multimodal}
\bibinfo{author}{N.~Cahan}, \bibinfo{author}{E.~Klang}, \bibinfo{author}{E.~M. Marom}, \bibinfo{author}{S.~Soffer}, \bibinfo{author}{Y.~Barash}, \bibinfo{author}{E.~Burshtein}, \bibinfo{author}{E.~Konen}, \bibinfo{author}{H.~Greenspan},
\newblock \bibinfo{title}{{Multimodal fusion models for pulmonary embolism mortality prediction}},
\newblock \bibinfo{journal}{Scientific Reports} \bibinfo{volume}{13} (\bibinfo{year}{2023}) \bibinfo{pages}{1--15}.
\bibitem[{Lin et~al.(2017)Lin, Goyal, Girshick, He, and Doll{\'a}r}]{focal_loss}
\bibinfo{author}{T.-Y. Lin}, \bibinfo{author}{P.~Goyal}, \bibinfo{author}{R.~Girshick}, \bibinfo{author}{K.~He}, \bibinfo{author}{P.~Doll{\'a}r},
\newblock \bibinfo{title}{{Focal loss for dense object detection}},
\newblock in: \bibinfo{booktitle}{Proceedings of the IEEE international conference on computer vision}, \bibinfo{year}{2017}, pp. \bibinfo{pages}{2980--2988}.
\bibitem[{Cahan et~al.(2022)Cahan, Marom, Soffer, Barash, Konen, Klang, and Greenspan}]{bib:54_cahan2022weakly}
\bibinfo{author}{N.~Cahan}, \bibinfo{author}{E.~M. Marom}, \bibinfo{author}{S.~Soffer}, \bibinfo{author}{Y.~Barash}, \bibinfo{author}{E.~Konen}, \bibinfo{author}{E.~Klang}, \bibinfo{author}{H.~Greenspan},
\newblock \bibinfo{title}{{Weakly supervised multimodal 30-day all-cause mortality prediction for pulmonary embolism patients}},
\newblock in: \bibinfo{booktitle}{2022 IEEE 19th International Symposium on Biomedical Imaging (ISBI)}, \bibinfo{organization}{IEEE}, \bibinfo{year}{2022}, pp. \bibinfo{pages}{1--4}.
\bibitem[{Stahlschmidt et~al.(2022)Stahlschmidt, Ulfenborg, and Synnergren}]{bib:missing-modality-healthcare}
\bibinfo{author}{S.~R. Stahlschmidt}, \bibinfo{author}{B.~Ulfenborg}, \bibinfo{author}{J.~Synnergren},
\newblock \bibinfo{title}{{Multimodal deep learning for biomedical data fusion: a review}},
\newblock \bibinfo{journal}{Briefings in Bioinformatics} \bibinfo{volume}{23} (\bibinfo{year}{2022}) \bibinfo{pages}{bbab569}.
\bibitem[{Holste et~al.(2021)Holste, Partridge, Rahbar, Biswas, Lee, and Alessio}]{bib:16_holste2021end}
\bibinfo{author}{G.~Holste}, \bibinfo{author}{S.~C. Partridge}, \bibinfo{author}{H.~Rahbar}, \bibinfo{author}{D.~Biswas}, \bibinfo{author}{C.~I. Lee}, \bibinfo{author}{A.~M. Alessio},
\newblock \bibinfo{title}{{End-to-end learning of fused image and non-image features for improved breast cancer classification from mri}},
\newblock in: \bibinfo{booktitle}{Proceedings of the IEEE/CVF International Conference on Computer Vision}, \bibinfo{year}{2021}, pp. \bibinfo{pages}{3294--3303}.
\bibitem[{Bhattacharya et~al.(2022)Bhattacharya, Sadasivuni, Chao, Agasthi, Ayoub, Holmes, Arsanjani, Sanyal, and Banerjee}]{bib:32_bhattacharya2022multi}
\bibinfo{author}{A.~Bhattacharya}, \bibinfo{author}{S.~Sadasivuni}, \bibinfo{author}{C.-J. Chao}, \bibinfo{author}{P.~Agasthi}, \bibinfo{author}{C.~Ayoub}, \bibinfo{author}{D.~R. Holmes}, \bibinfo{author}{R.~Arsanjani}, \bibinfo{author}{A.~Sanyal}, \bibinfo{author}{I.~Banerjee},
\newblock \bibinfo{title}{{Multi-modal fusion model for predicting adverse cardiovascular outcome post percutaneous coronary intervention}},
\newblock \bibinfo{journal}{Physiological Measurement} \bibinfo{volume}{43} (\bibinfo{year}{2022}) \bibinfo{pages}{124004}.
\bibitem[{Sousa et~al.(2023)Sousa, Matos, Silva, Freitas, Oliveira, and Pereira}]{bib:47_sousa2023single}
\bibinfo{author}{J.~V. Sousa}, \bibinfo{author}{P.~Matos}, \bibinfo{author}{F.~Silva}, \bibinfo{author}{P.~Freitas}, \bibinfo{author}{H.~P. Oliveira}, \bibinfo{author}{T.~Pereira},
\newblock \bibinfo{title}{{Single Modality vs. Multimodality: What Works Best for Lung Cancer Screening?}},
\newblock \bibinfo{journal}{Sensors} \bibinfo{volume}{23} (\bibinfo{year}{2023}) \bibinfo{pages}{5597}.
\bibitem[{He et~al.(2021)He, Han, Zhang, and Chen}]{bib:19_he2021hierarchical}
\bibinfo{author}{M.~He}, \bibinfo{author}{K.~Han}, \bibinfo{author}{Y.~Zhang}, \bibinfo{author}{W.~Chen},
\newblock \bibinfo{title}{{Hierarchical-order multimodal interaction fusion network for grading gliomas}},
\newblock \bibinfo{journal}{Physics in Medicine \& Biology} \bibinfo{volume}{66} (\bibinfo{year}{2021}) \bibinfo{pages}{215016}.
\bibitem[{Pan et~al.(2021)Pan, Zhou, Tan, Sun, Guan, Wang, Luo, and Lu}]{bib:23_pan2021liver}
\bibinfo{author}{C.~Pan}, \bibinfo{author}{P.~Zhou}, \bibinfo{author}{J.~Tan}, \bibinfo{author}{B.~Sun}, \bibinfo{author}{R.~Guan}, \bibinfo{author}{Z.~Wang}, \bibinfo{author}{Y.~Luo}, \bibinfo{author}{J.~Lu},
\newblock \bibinfo{title}{{Liver tumor detection via a multi-scale intermediate multi-modal fusion network on MRI images}},
\newblock in: \bibinfo{booktitle}{2021 IEEE international conference on image processing (ICIP)}, \bibinfo{organization}{IEEE}, \bibinfo{year}{2021}, pp. \bibinfo{pages}{299--303}.
\bibitem[{Ahmad et~al.(2023)Ahmad, Xia, Cui, and Islam}]{bib:03_ahmad2023aatsn}
\bibinfo{author}{I.~Ahmad}, \bibinfo{author}{Y.~Xia}, \bibinfo{author}{H.~Cui}, \bibinfo{author}{Z.~U. Islam},
\newblock \bibinfo{title}{{AATSN: Anatomy Aware Tumor Segmentation Network for PET-CT volumes and images using a lightweight fusion-attention mechanism}},
\newblock \bibinfo{journal}{Computers in Biology and Medicine} \bibinfo{volume}{157} (\bibinfo{year}{2023}) \bibinfo{pages}{106748}.
\bibitem[{Li et~al.(2022)Li, Zou, Wu, Dai, Bai, and Jiao}]{bib:02_li2022dynamic}
\bibinfo{author}{Y.~Li}, \bibinfo{author}{B.~Zou}, \bibinfo{author}{J.~Wu}, \bibinfo{author}{Y.~Dai}, \bibinfo{author}{H.~X. Bai}, \bibinfo{author}{Z.~Jiao},
\newblock \bibinfo{title}{{A dynamic multi-modal fusion network for ovarian tumor differentiation}},
\newblock in: \bibinfo{booktitle}{2022 IEEE International Conference on Bioinformatics and Biomedicine (BIBM)}, \bibinfo{organization}{IEEE}, \bibinfo{year}{2022}, pp. \bibinfo{pages}{767--772}.
\bibitem[{Meng et~al.(2022)Meng, Zhu, Pang, Tian, Nie, and Wang}]{bib:30_meng2022msmfn}
\bibinfo{author}{Z.~Meng}, \bibinfo{author}{Y.~Zhu}, \bibinfo{author}{W.~Pang}, \bibinfo{author}{J.~Tian}, \bibinfo{author}{F.~Nie}, \bibinfo{author}{K.~Wang},
\newblock \bibinfo{title}{{MSMFN: An Ultrasound Based Multi-Step Modality Fusion Network for Identifying the Histologic Subtypes of Metastatic Cervical Lymphadenopathy}},
\newblock \bibinfo{journal}{IEEE Transactions on Medical Imaging} \bibinfo{volume}{42} (\bibinfo{year}{2022}) \bibinfo{pages}{996--1008}.
\bibitem[{Huang et~al.(2020)Huang, Pareek, Zamanian, Banerjee, and Lungren}]{bib:39_huang2020multimodal}
\bibinfo{author}{S.-C. Huang}, \bibinfo{author}{A.~Pareek}, \bibinfo{author}{R.~Zamanian}, \bibinfo{author}{I.~Banerjee}, \bibinfo{author}{M.~P. Lungren},
\newblock \bibinfo{title}{{Multimodal fusion with deep neural networks for leveraging CT imaging and electronic health record: a case-study in pulmonary embolism detection}},
\newblock \bibinfo{journal}{Scientific reports} \bibinfo{volume}{10} (\bibinfo{year}{2020}) \bibinfo{pages}{22147}.
\bibitem[{Alsherbiny et~al.(2021)Alsherbiny, Radwan, Moustafa, Bhuyan, El-Waisi, Chang, and Li}]{bib:52_alsherbiny2021trustworthy}
\bibinfo{author}{M.~A. Alsherbiny}, \bibinfo{author}{I.~Radwan}, \bibinfo{author}{N.~Moustafa}, \bibinfo{author}{D.~J. Bhuyan}, \bibinfo{author}{M.~El-Waisi}, \bibinfo{author}{D.~Chang}, \bibinfo{author}{C.~G. Li},
\newblock \bibinfo{title}{{Trustworthy Deep Neural Network for Inferring Anticancer Synergistic Combinations}},
\newblock \bibinfo{journal}{IEEE Journal of Biomedical and Health Informatics} \bibinfo{volume}{27} (\bibinfo{year}{2021}) \bibinfo{pages}{1691--1700}.
\bibitem[{Cen et~al.(2022)Cen, Wang, Olugbade, Williams, and Bianchi-Berthouze}]{bib:17_cen2022exploring}
\bibinfo{author}{G.~Cen}, \bibinfo{author}{C.~Wang}, \bibinfo{author}{T.~A. Olugbade}, \bibinfo{author}{A.~C. d.~C. Williams}, \bibinfo{author}{N.~Bianchi-Berthouze},
\newblock \bibinfo{title}{{Exploring multimodal fusion for continuous protective behavior detection}},
\newblock in: \bibinfo{booktitle}{2022 10th International Conference on Affective Computing and Intelligent Interaction (ACII)}, \bibinfo{organization}{IEEE}, \bibinfo{year}{2022}, pp. \bibinfo{pages}{1--8}.
\bibitem[{Chen et~al.(2022)Chen, Hong, Guo, Hao, and Hu}]{bib:29_chen2022ms}
\bibinfo{author}{T.~Chen}, \bibinfo{author}{R.~Hong}, \bibinfo{author}{Y.~Guo}, \bibinfo{author}{S.~Hao}, \bibinfo{author}{B.~Hu},
\newblock \bibinfo{title}{{MS}$^2-${GNN: Exploring GNN}$-${Based Multimodal Fusion Network for Depression Detection}},
\newblock \bibinfo{journal}{IEEE Transactions on Cybernetics}  (\bibinfo{year}{2022}).
\bibitem[{Njoku et~al.(2022)Njoku, Caliwag, Lim, Kim, Hwang, and Jung}]{bib:10_njoku2022deep}
\bibinfo{author}{J.~N. Njoku}, \bibinfo{author}{A.~C. Caliwag}, \bibinfo{author}{W.~Lim}, \bibinfo{author}{S.~Kim}, \bibinfo{author}{H.~Hwang}, \bibinfo{author}{J.~Jung},
\newblock \bibinfo{title}{{Deep learning based data fusion methods for multimodal emotion recognition}},
\newblock \bibinfo{journal}{The Journal of Korean Institute of Communications and Information Sciences} \bibinfo{volume}{47} (\bibinfo{year}{2022}) \bibinfo{pages}{79--87}.
\bibitem[{Mohammed et~al.(2023)Mohammed, Omeroglu, and Oral}]{bib:27_mohammed2023mmhfnet}
\bibinfo{author}{H.~M. Mohammed}, \bibinfo{author}{A.~N. Omeroglu}, \bibinfo{author}{E.~A. Oral},
\newblock \bibinfo{title}{{MMHFNet: Multi-modal and multi-layer hybrid fusion network for voice pathology detection}},
\newblock \bibinfo{journal}{Expert Systems with Applications} \bibinfo{volume}{223} (\bibinfo{year}{2023}) \bibinfo{pages}{119790}.
\bibitem[{Sedghi et~al.(2020)Sedghi, Mehrtash, Jamzad, Amalou, Wells, Kapur, Kwak, Turkbey, Choyke, Pinto et~al.}]{bib:20_sedghi2020improving}
\bibinfo{author}{A.~Sedghi}, \bibinfo{author}{A.~Mehrtash}, \bibinfo{author}{A.~Jamzad}, \bibinfo{author}{A.~Amalou}, \bibinfo{author}{W.~M. Wells}, \bibinfo{author}{T.~Kapur}, \bibinfo{author}{J.~T. Kwak}, \bibinfo{author}{B.~Turkbey}, \bibinfo{author}{P.~Choyke}, \bibinfo{author}{P.~Pinto}, et~al.,
\newblock \bibinfo{title}{{Improving detection of prostate cancer foci via information fusion of MRI and temporal enhanced ultrasound}},
\newblock \bibinfo{journal}{International journal of computer assisted radiology and surgery} \bibinfo{volume}{15} (\bibinfo{year}{2020}) \bibinfo{pages}{1215--1223}.
\bibitem[{Guida et~al.(2023)Guida, Zhang, and Shan}]{bib:21_guida2023improving}
\bibinfo{author}{C.~Guida}, \bibinfo{author}{M.~Zhang}, \bibinfo{author}{J.~Shan},
\newblock \bibinfo{title}{{Improving knee osteoarthritis classification using multimodal intermediate fusion of X-ray, MRI, and clinical information}},
\newblock \bibinfo{journal}{Neural Computing and Applications}  (\bibinfo{year}{2023}) \bibinfo{pages}{1--10}.
\bibitem[{Oh et~al.(2023)Oh, Kang, Oh, and Kim}]{bib:11_oh2023deep}
\bibinfo{author}{S.~Oh}, \bibinfo{author}{S.-R. Kang}, \bibinfo{author}{I.-J. Oh}, \bibinfo{author}{M.-S. Kim},
\newblock \bibinfo{title}{{Deep learning model integrating positron emission tomography and clinical data for prognosis prediction in non-small cell lung cancer patients}},
\newblock \bibinfo{journal}{BMC bioinformatics} \bibinfo{volume}{24} (\bibinfo{year}{2023}) \bibinfo{pages}{1--13}.
\bibitem[{Zheng et~al.(2022)Zheng, Cai, Chua, Herschel, Zhang, and Ooi}]{bib:15_zheng2022dyhealth}
\bibinfo{author}{K.~Zheng}, \bibinfo{author}{S.~Cai}, \bibinfo{author}{H.~R. Chua}, \bibinfo{author}{M.~Herschel}, \bibinfo{author}{M.~Zhang}, \bibinfo{author}{B.~C. Ooi},
\newblock \bibinfo{title}{{DyHealth: making neural networks dynamic for effective healthcare analytics}},
\newblock \bibinfo{journal}{Proceedings of the VLDB Endowment} \bibinfo{volume}{15} (\bibinfo{year}{2022}) \bibinfo{pages}{3445--3458}.
\bibitem[{Zhang et~al.(2023)Zhang, Jian, Xu, Zhao, Lu, Lin, and Xie}]{bib:22_zhang2023itcep}
\bibinfo{author}{Y.~Zhang}, \bibinfo{author}{X.~Jian}, \bibinfo{author}{L.~Xu}, \bibinfo{author}{J.~Zhao}, \bibinfo{author}{M.~Lu}, \bibinfo{author}{Y.~Lin}, \bibinfo{author}{L.~Xie},
\newblock \bibinfo{title}{{iTCep: a deep learning framework for identification of T cell epitopes by harnessing fusion features}},
\newblock \bibinfo{journal}{Frontiers in Genetics} \bibinfo{volume}{14} (\bibinfo{year}{2023}) \bibinfo{pages}{1141535}.
\bibitem[{Guarrasi and Soda(2023)}]{bib:33_guarrasi2023multi}
\bibinfo{author}{V.~Guarrasi}, \bibinfo{author}{P.~Soda},
\newblock \bibinfo{title}{{Multi-objective optimization determines when, which and how to fuse deep networks: An application to predict COVID-19 outcomes}},
\newblock \bibinfo{journal}{Computers in Biology and Medicine} \bibinfo{volume}{154} (\bibinfo{year}{2023}) \bibinfo{pages}{106625}.
\bibitem[{Perez and Ventura(2022)}]{bib:34_perez2022multi}
\bibinfo{author}{E.~Perez}, \bibinfo{author}{S.~Ventura},
\newblock \bibinfo{title}{{Multi-view Deep Neural Networks for multiclass skin lesion diagnosis}},
\newblock in: \bibinfo{booktitle}{2022 IEEE International Conference on Omni-layer Intelligent Systems (COINS)}, \bibinfo{organization}{IEEE}, \bibinfo{year}{2022}, pp. \bibinfo{pages}{1--6}.
\bibitem[{Steyaert et~al.(2023)Steyaert, Qiu, Zheng, Mukherjee, Vogel, and Gevaert}]{bib:35_steyaert2023multimodal}
\bibinfo{author}{S.~Steyaert}, \bibinfo{author}{Y.~L. Qiu}, \bibinfo{author}{Y.~Zheng}, \bibinfo{author}{P.~Mukherjee}, \bibinfo{author}{H.~Vogel}, \bibinfo{author}{O.~Gevaert},
\newblock \bibinfo{title}{{Multimodal deep learning to predict prognosis in adult and pediatric brain tumors}},
\newblock \bibinfo{journal}{Communications Medicine} \bibinfo{volume}{3} (\bibinfo{year}{2023}) \bibinfo{pages}{44}.
\bibitem[{Wu et~al.(2023)Wu, Daoudi, and Amad}]{bib:51_wu2023transformer}
\bibinfo{author}{Y.~Wu}, \bibinfo{author}{M.~Daoudi}, \bibinfo{author}{A.~Amad},
\newblock \bibinfo{title}{{Transformer-based self-supervised multimodal representation learning for wearable emotion recognition}},
\newblock \bibinfo{journal}{IEEE Transactions on Affective Computing}  (\bibinfo{year}{2023}).
\bibitem[{Han et~al.(2022)Han, Yang, Huang, Zhang, and Yao}]{bib:36_han2022multimodal}
\bibinfo{author}{Z.~Han}, \bibinfo{author}{F.~Yang}, \bibinfo{author}{J.~Huang}, \bibinfo{author}{C.~Zhang}, \bibinfo{author}{J.~Yao},
\newblock \bibinfo{title}{{Multimodal dynamics: Dynamical fusion for trustworthy multimodal classification}},
\newblock in: \bibinfo{booktitle}{Proceedings of the IEEE/CVF Conference on Computer Vision and Pattern Recognition}, \bibinfo{year}{2022}, pp. \bibinfo{pages}{20707--20717}.
\bibitem[{Rashid et~al.(2023)Rashid, Kallakuri, and Mohsenin}]{bib:49_rashid2023tinym2net}
\bibinfo{author}{H.-A. Rashid}, \bibinfo{author}{U.~Kallakuri}, \bibinfo{author}{T.~Mohsenin},
\newblock \bibinfo{title}{{TinyM}$^2${Net-V2: A Compact Low Power Software Hardware Architecture for Multimodal Deep Neural Networks}},
\newblock \bibinfo{journal}{ACM Transactions on Embedded Computing Systems}  (\bibinfo{year}{2023}).
\bibitem[{Jiao et~al.(2023)Jiao, Sun, Huang, Xia, Qiao, Ren, Wang, and Guo}]{bib:18_jiao2023gmrlnet}
\bibinfo{author}{J.~Jiao}, \bibinfo{author}{H.~Sun}, \bibinfo{author}{Y.~Huang}, \bibinfo{author}{M.~Xia}, \bibinfo{author}{M.~Qiao}, \bibinfo{author}{Y.~Ren}, \bibinfo{author}{Y.~Wang}, \bibinfo{author}{Y.~Guo},
\newblock \bibinfo{title}{{GMRLNet: A graph-based manifold regularization learning framework for placental insufficiency diagnosis on incomplete multimodal ultrasound data}},
\newblock \bibinfo{journal}{IEEE Transactions on Medical Imaging}  (\bibinfo{year}{2023}).
\bibitem[{Li et~al.(2022)Li, El~Habib~Daho, Conze, Al~Hajj, Bonnin, Ren, Manivannan, Magazzeni, Tadayoni, Cochener et~al.}]{bib:41_li2022multimodal}
\bibinfo{author}{Y.~Li}, \bibinfo{author}{M.~El~Habib~Daho}, \bibinfo{author}{P.-H. Conze}, \bibinfo{author}{H.~Al~Hajj}, \bibinfo{author}{S.~Bonnin}, \bibinfo{author}{H.~Ren}, \bibinfo{author}{N.~Manivannan}, \bibinfo{author}{S.~Magazzeni}, \bibinfo{author}{R.~Tadayoni}, \bibinfo{author}{B.~Cochener}, et~al.,
\newblock \bibinfo{title}{{Multimodal information fusion for glaucoma and diabetic retinopathy classification}},
\newblock in: \bibinfo{booktitle}{International Workshop on Ophthalmic Medical Image Analysis}, \bibinfo{organization}{Springer}, \bibinfo{year}{2022}, pp. \bibinfo{pages}{53--62}.
\bibitem[{Qiu et~al.(2023)Qiu, Lin, Chen, and Xu}]{bib:qiu2023pre}
\bibinfo{author}{Y.~Qiu}, \bibinfo{author}{F.~Lin}, \bibinfo{author}{W.~Chen}, \bibinfo{author}{M.~Xu},
\newblock \bibinfo{title}{{Pre-training in Medical Data: A Survey}},
\newblock \bibinfo{journal}{Machine Intelligence Research} \bibinfo{volume}{20} (\bibinfo{year}{2023}) \bibinfo{pages}{147--179}.
\bibitem[{Matsoukas et~al.(2022)Matsoukas, Haslum, Sorkhei, S{\"o}derberg, and Smith}]{bib:matsoukas2022makes}
\bibinfo{author}{C.~Matsoukas}, \bibinfo{author}{J.~F. Haslum}, \bibinfo{author}{M.~Sorkhei}, \bibinfo{author}{M.~S{\"o}derberg}, \bibinfo{author}{K.~Smith},
\newblock \bibinfo{title}{{What makes transfer learning work for medical images: Feature reuse \& other factors}},
\newblock in: \bibinfo{booktitle}{Proceedings of the IEEE/CVF Conference on Computer Vision and Pattern Recognition}, \bibinfo{year}{2022}, pp. \bibinfo{pages}{9225--9234}.
\bibitem[{Russakovsky et~al.(2015)Russakovsky, Deng, Su, Krause, Satheesh, Ma, Huang, Karpathy, Khosla, Bernstein et~al.}]{russakovsky2015imagenet}
\bibinfo{author}{O.~Russakovsky}, \bibinfo{author}{J.~Deng}, \bibinfo{author}{H.~Su}, \bibinfo{author}{J.~Krause}, \bibinfo{author}{S.~Satheesh}, \bibinfo{author}{S.~Ma}, \bibinfo{author}{Z.~Huang}, \bibinfo{author}{A.~Karpathy}, \bibinfo{author}{A.~Khosla}, \bibinfo{author}{M.~Bernstein}, et~al.,
\newblock \bibinfo{title}{{Imagenet large scale visual recognition challenge}},
\newblock \bibinfo{journal}{International journal of computer vision} \bibinfo{volume}{115} (\bibinfo{year}{2015}) \bibinfo{pages}{211--252}.
\bibitem[{Mumuni and Mumuni(2022)}]{mumuni2022data}
\bibinfo{author}{A.~Mumuni}, \bibinfo{author}{F.~Mumuni},
\newblock \bibinfo{title}{{Data augmentation: A comprehensive survey of modern approaches}},
\newblock \bibinfo{journal}{Array} \bibinfo{volume}{16} (\bibinfo{year}{2022}) \bibinfo{pages}{100258}.
\bibitem[{Gat et~al.(2020)Gat, Schwartz, Schwing, and Hazan}]{bib:gat2020removing}
\bibinfo{author}{I.~Gat}, \bibinfo{author}{I.~Schwartz}, \bibinfo{author}{A.~Schwing}, \bibinfo{author}{T.~Hazan},
\newblock \bibinfo{title}{{Removing bias in multi-modal classifiers: Regularization by maximizing functional entropies}},
\newblock \bibinfo{journal}{Advances in Neural Information Processing Systems} \bibinfo{volume}{33} (\bibinfo{year}{2020}) \bibinfo{pages}{3197--3208}.
\bibitem[{Samek and M{\"u}ller(2019)}]{bib:samek2019towards}
\bibinfo{author}{W.~Samek}, \bibinfo{author}{K.-R. M{\"u}ller},
\newblock \bibinfo{title}{{Towards explainable artificial intelligence}},
\newblock \bibinfo{journal}{Explainable AI: interpreting, explaining and visualizing deep learning}  (\bibinfo{year}{2019}) \bibinfo{pages}{5--22}.
\bibitem[{Joshi et~al.(2021)Joshi, Walambe, and Kotecha}]{bib:joshi2021review}
\bibinfo{author}{G.~Joshi}, \bibinfo{author}{R.~Walambe}, \bibinfo{author}{K.~Kotecha},
\newblock \bibinfo{title}{{A review on explainability in multimodal deep neural nets}},
\newblock \bibinfo{journal}{IEEE Access} \bibinfo{volume}{9} (\bibinfo{year}{2021}) \bibinfo{pages}{59800--59821}.
\bibitem[{Caruso et~al.(2022)Caruso, Guarrasi, Cordelli, Sicilia, Gentile, Messina, Fiore, Piccolo, Beomonte~Zobel, Iannello et~al.}]{bib:caruso2022multimodal}
\bibinfo{author}{C.~M. Caruso}, \bibinfo{author}{V.~Guarrasi}, \bibinfo{author}{E.~Cordelli}, \bibinfo{author}{R.~Sicilia}, \bibinfo{author}{S.~Gentile}, \bibinfo{author}{L.~Messina}, \bibinfo{author}{M.~Fiore}, \bibinfo{author}{C.~Piccolo}, \bibinfo{author}{B.~Beomonte~Zobel}, \bibinfo{author}{G.~Iannello}, et~al.,
\newblock \bibinfo{title}{{A multimodal ensemble driven by multiobjective optimisation to predict overall survival in non-small-cell lung cancer}},
\newblock \bibinfo{journal}{Journal of Imaging} \bibinfo{volume}{8} (\bibinfo{year}{2022}) \bibinfo{pages}{298}.
\bibitem[{Deng et~al.(2009)Deng, Dong, Socher, Li, Li, and Fei-Fei}]{bib:deng2009imagenet}
\bibinfo{author}{J.~Deng}, \bibinfo{author}{W.~Dong}, \bibinfo{author}{R.~Socher}, \bibinfo{author}{L.-J. Li}, \bibinfo{author}{K.~Li}, \bibinfo{author}{L.~Fei-Fei},
\newblock \bibinfo{title}{{Imagenet: A large-scale hierarchical image database}},
\newblock in: \bibinfo{booktitle}{2009 IEEE conference on computer vision and pattern recognition}, \bibinfo{organization}{Ieee}, \bibinfo{year}{2009}, pp. \bibinfo{pages}{248--255}.
\bibitem[{Guarrasi et~al.(2024)Guarrasi, Tronchin, Albano, Faiella, Fazzini, Santucci, and Soda}]{bib:guarrasi2024multimodal}
\bibinfo{author}{V.~Guarrasi}, \bibinfo{author}{L.~Tronchin}, \bibinfo{author}{D.~Albano}, \bibinfo{author}{E.~Faiella}, \bibinfo{author}{D.~Fazzini}, \bibinfo{author}{D.~Santucci}, \bibinfo{author}{P.~Soda},
\newblock \bibinfo{title}{{Multimodal explainability via latent shift applied to COVID-19 stratification}},
\newblock \bibinfo{journal}{Pattern Recognition}  (\bibinfo{year}{2024}) \bibinfo{pages}{110825}.
\bibitem[{Rofena et~al.(2024)Rofena, Guarrasi, Sarli, Piccolo, Sammarra, Zobel, and Soda}]{bib:rofena2024deep}
\bibinfo{author}{A.~Rofena}, \bibinfo{author}{V.~Guarrasi}, \bibinfo{author}{M.~Sarli}, \bibinfo{author}{C.~L. Piccolo}, \bibinfo{author}{M.~Sammarra}, \bibinfo{author}{B.~B. Zobel}, \bibinfo{author}{P.~Soda},
\newblock \bibinfo{title}{{A deep learning approach for virtual contrast enhancement in Contrast Enhanced Spectral Mammography}},
\newblock \bibinfo{journal}{Computerized Medical Imaging and Graphics} \bibinfo{volume}{116} (\bibinfo{year}{2024}) \bibinfo{pages}{102398}.
\bibitem[{Poon(2023)}]{poon2023multimodal}
\bibinfo{author}{H.~Poon},
\newblock \bibinfo{title}{{Multimodal Generative AI for Precision Health}},
\newblock \bibinfo{journal}{NEJM AI Sponsored}  (\bibinfo{year}{2023}).

\end{thebibliography}

\end{document}